\documentclass[journal,transmag]{IEEEtran}

\usepackage{mathrsfs,bm,enumerate}
\usepackage[bottom]{footmisc}
\usepackage{setspace}
\usepackage{multirow}
\usepackage{graphicx}
\usepackage{epstopdf}
\usepackage{braket}
\usepackage{tabularx}
\usepackage{multirow,booktabs}
\usepackage{subfigure}
\usepackage{rotating}
\usepackage{amsmath}
\usepackage{amsfonts,stmaryrd}
\usepackage{algorithm}
\usepackage{algorithmic}
\usepackage{array}

\ifCLASSOPTIONcompsoc
 \usepackage[caption=false,font=normalsize,labelfont=sf,textfont=sf]{subfig}
\else
 \usepackage[caption=false,font=footnotesize]{subfig}
\fi

\begin{document}
  \graphicspath{{./figures_dir/}}
\title{Low-Rank and Sparse Enhanced Tucker Decomposition for Tensor Completion}

% \author{\IEEEauthorblockN{Michael Shell\IEEEauthorrefmark{1},
% Homer Simpson\IEEEauthorrefmark{2},
% James Kirk\IEEEauthorrefmark{3},
% Montgomery Scott\IEEEauthorrefmark{3}, and
% Eldon Tyrell\IEEEauthorrefmark{4},~\IEEEmembership{Fellow,~IEEE}}
% \IEEEauthorblockA{\IEEEauthorrefmark{1}School of Electrical and Computer Engineering,
% Georgia Institute of Technology, Atlanta, GA 30332 USA}
% \IEEEauthorblockA{\IEEEauthorrefmark{2}Twentieth Century Fox, Springfield, USA}
% \IEEEauthorblockA{\IEEEauthorrefmark{3}Starfleet Academy, San Francisco, CA 96678 USA}
% \IEEEauthorblockA{\IEEEauthorrefmark{4}Tyrell Inc., 123 Replicant Street, Los Angeles, CA 90210 USA}% <-this % stops an unwanted space
% \thanks{Manuscript received December 1, 2012; revised August 26, 2015.
% Corresponding author: M. Shell (email: http://www.michaelshell.org/contact.html).}}

\author{\IEEEauthorblockN{Chenjian Pan\IEEEauthorrefmark{1},
Chen Ling\IEEEauthorrefmark{1}, Hongjin He\IEEEauthorrefmark{2}, Liqun Qi\IEEEauthorrefmark{3}, Yanwei Xu\IEEEauthorrefmark{4}
% James Kirk\IEEEauthorrefmark{3},
% Montgomery Scott\IEEEauthorrefmark{3}, and
% Eldon Tyrell\IEEEauthorrefmark{4},~\IEEEmembership{Fellow,~IEEE}
}
\IEEEauthorblockA{\IEEEauthorrefmark{1}Department of Mathematics, School of Science, Hangzhou Dianzi University,
Hangzhou, 310018, China.}
\IEEEauthorblockA{\IEEEauthorrefmark{2} School of Mathematics and Statistics, Ningbo University, Ningbo 315211, China.}
\IEEEauthorblockA{\IEEEauthorrefmark{3} Department of Applied Mathematics, The Hong Kong Polytechnic University, Hung Hom, Kowloon, Hong Kong, China.}
\IEEEauthorblockA{\IEEEauthorrefmark{4} Future Network Theory Lab, 2012 Labs Huawei Tech. Investment Co., Ltd, Shatin, New Territory, Hong Kong, China.}
\thanks{Manuscript received May XX, 2021; revised XX, 2021.}}

\markboth{Journal of \LaTeX\ Class Files,~Vol.~xx, No.~xx, May~2021}%
{Shell \MakeLowercase{\textit{et al.}}: Bare Demo of IEEEtran.cls for IEEE Transactions on Magnetics Journals}



% for Transactions on Magnetics papers, we must declare the abstract and
% index terms PRIOR to the title within the \IEEEtitleabstractindextext
% IEEEtran command as these need to go into the title area created by
\IEEEtitleabstractindextext{%
\begin{abstract}
Tensor completion refers to the task of estimating the missing data from an incomplete measurement or observation, which is a core problem frequently arising from the areas of big data analysis, computer vision, and network engineering. Due to the multidimensional nature of high-order tensors, the matrix approaches, e.g., matrix factorization and direct matricization of tensors, are often not ideal for tensor completion and recovery. In this paper, we introduce a unified low-rank and sparse enhanced Tucker decomposition model for tensor completion. Our model possesses a sparse regularization term to promote a sparse core tensor of the Tucker decomposition, which is beneficial for tensor data compression. Moreover, we enforce low-rank regularization terms on factor matrices of the Tucker decomposition for inducing the low-rankness of the tensor with a cheap computational cost. Numerically, we propose a customized ADMM with enough easy  subproblems to solve the underlying model. It is remarkable that our model is able to deal with different types of real-world data sets, since it exploits the potential periodicity and inherent correlation properties appeared in tensors. A series of computational experiments on real-world data sets, including internet traffic data sets, color images, and face recognition, demonstrate that our model performs better than many existing state-of-the-art matricization and tensorization approaches in terms of achieving higher recovery accuracy.
\end{abstract}

}

\maketitle

\IEEEdisplaynontitleabstractindextext

% \fi
\IEEEpeerreviewmaketitle

\section{Introduction}
\IEEEPARstart{I}{n} the era of big data and artificial intelligence, more and more information are collected for analysis and making decisions. One remarkable feature of these information is that they usually have complex structures (or multi-label) and higher dimensions. Naturally, these data would be stored as higher-order tensor (a.k.a., multi-dimensional array), which can better express the underlying complex essential structures of multi-dimensional data than vectors and matrices. In the past decades, we have witnessed the widespread applications of tensors in psychometrics, chemometrics, data mining, graph analysis, signal processing, and machine learning, e.g., see \cite{KB09,KS12,SF06,SDFHPF17}, to name just a few. However, it often takes unacceptable cost to acquire or collect complete data, or some information is possibly missed during transmission. Hence, we usually obtain incomplete tensors which cannot be further used directly. In this situation, a natural and fundamental problem is to estimate the missing entries from an incomplete tensor. Namely, such a problem is called tensor completion. In the literature, it has been well-documented that higher-order tensor completion has been widely used in image inpainting \cite{BSCB00,K06,TCWZR14,ZBN20}, face recognition \cite{HKBH13,SH94,VT02}, magnetic resonance imaging data recovery \cite{JHZJD18,VMG12}, internet traffic data recovery \cite{RZWQE12,XWWXWZCZ18,ZZXC15}, high-order web link analysis \cite{KBK05} and personalized web search \cite{SZLLC05}, and so on.

\subsection{Related Work}
Generally speaking, it is very difficult, even impossible, to accurately fill the missing entries for general incomplete tensors, provided that we have no prior information or property on these incomplete tensors. However, it is fortunate that many real-world big data sets often have a strongly or an approximately inherent correlation property, which can be regarded as the low-rank property \cite{UT19}. When considering the matrix completion \cite{CR09}, which is a special case of tensor completion, one natural low-rank minimization model has been well studied in the literature. Specifically, we can recover the incomplete data matrix by solving the optimization problem
\begin{align}\label{MC}
	\min_F \;\;\;& {\rm rank}(F) \nonumber \\
	{\rm s.t.}\;\;\;& {\mathscr P}_{\Omega}(F)={\mathscr P}_{\Omega}(M),
\end{align}
where ${\rm rank}(F)$ represents the rank of the underlying matrix $F$ and $M\in\mathbb{R}^{I_1\times I_2}$ is an observed incomplete matrix, and $\Omega$ is the set of locations corresponding to the observed entries (i.e., $(i,j)\in\Omega$ if $M_{ij}$ is observed). Notice that the linear operator ${\mathscr P}_{\Omega}(\cdot)$ used throughout this paper extracts known elements in the set $\Omega$ and fills the others that are not in $\Omega$ with zeros. Following the spirit of matrix completion model \eqref{MC}, the higher-order (i.e., $N$-order for $N\geq 3$) low-rank tensor completion problem can be mathematically characterized as
\begin{align}\label{LRTC-optim}
\min_{\mathcal{F}}\;\;\;&{\rm rank}(\mathcal{F}) \nonumber\\
\text{s.t.}\;\;\;&{\mathscr P}_{\Omega}(\mathcal{F})={\mathscr P}_{\Omega}(\mathcal{M}),
\end{align}
where ${\rm rank}(\mathcal{F})$ denotes the rank of tensor $\mathcal{F}\in\mathbb{R}^{I_1\times I_2\times\cdots\times I_N}$; $\mathcal{M}\in\mathbb{R}^{I_1\times I_2\times\cdots\times I_N}$ is an observed incomplete tensor and $\Omega$ is the index set corresponding to the observed entries of $\mathcal{M}$. Even though model \eqref{LRTC-optim} is a generalization of \eqref{MC}, both models have essential differences since there is no unique definition for tensor ranks, while the rank of a matrix is exactly equivalent to its number of nonzero singular values. In the tensor literature, there are two popular ways to characterize tensor ranks, i.e., CANDECOMP/PARAFAC (CP) rank and Tucker rank (also named $n$-rank), which are closely related to CP decomposition and Tucker decomposition (e.g., see \cite{KB09}), respectively. However, directly minimizing the CP-rank or Tucker rank in model \eqref{LRTC-optim} is NP-hard \cite{HL13,MLH17}. Actually, even though we consider the simplest low-rank matrix model \eqref{MC}, it is unfortunate that such a minimization problem is NP-hard \cite{CR09}. Therefore, inspired by the relationship between the matrix rank and singular values, an alternative convex relaxation model, which minimizes the sum of the singular values over the constraint set, is proposed as follows:
\begin{align}\label{CMC}
\min_F \;\;\;& \|F\|_*\nonumber \\
{\rm s.t.}\;\;\;& {\mathscr P}_{\Omega}(F)={\mathscr P}_{\Omega}(M),
\end{align}
where $\|\cdot\|_*$ is called the nuclear norm representing the sum of singular values of a matrix. As a consequence, the convex optimization model \eqref{CMC} can be tackled perfectly by many state-of-the-art first-order methods. Comparatively, it is not so lucky for model \eqref{LRTC-optim} since we cannot find a unified convex surrogate approximation for tensor ranks. However, it is well-known that any tensor can be unfolded (or matricized) in mode-$n$ (see Section \ref{NarPrel}). Consequently, we can gainfully employ the nuclear norms of the resulting unfolded matrices of $\mathcal{F}$ instead of the ${\rm rank}(\mathcal{F})$ in \eqref{LRTC-optim}. Such an idea has been considered in \cite{LMWY13,GRY11}, where the authors
considered the minimization of weighted sum of nuclear norms of all modes unfolding, which takes the form
\begin{align}\label{LNNTC-optim0}
\min_{\mathcal{F}}\;\;\;&\sum_{n=1}^N\alpha_n\|F_{(n)}\|_* \nonumber\\
\text{s.t.}\;\;\;&{\mathscr P}_{\Omega}(\mathcal{F})={\mathscr P}_{\Omega}(\mathcal{M}),
\end{align}
where $\alpha_n\geq 0$ for $n=1,2,\ldots,N$ are weighting parameters satisfying $\sum_{n=1}^N\alpha_n=1$, and $F_{(n)}$ is the mode-$n$ unfolding matrix of tensor $\mathcal{F}$ for every $n\in [N]:=\{1,2,\cdots,N\}$. Clearly, optimization model \eqref{LNNTC-optim0} fully exploits the unfolding structure of tensors and can be solved by some state-of-the-art first-order methods, e.g., block coordinate descent method, Douglas-Rachford splitting method and alternating direction method of multipliers (ADMM) (see \cite{LMWY13,GRY11} for more details) with a high completion accuracy. However, these unfolding matrices usually are large scale, thereby reducing the efficiency of solving model \eqref{LNNTC-optim0} due to the computational expensive singular value decomposition (SVD) on every unfolding matrix $F_{(n)}$ at each iteration. To tackle this issue, Xu et al. \cite{XHYS15} proposed a  matrix factorization based completion model, which employs a series of matrices factorization to replace the nuclear norms, i.e.,
\begin{align}\label{LNNTC-optim}
\min_{\mathcal{F}, X,Y}\;\;&\sum_{n=1}^N\frac{\alpha_n}{2}\|F_{(n)}-X_nY_n\|_F^2 \nonumber\\
\text{s.t.   }\;\;&{\mathscr P}_{\Omega}(\mathcal{F})={\mathscr P}_{\Omega}(\mathcal{M}),
\end{align}
where $\mathcal{F}\in\mathbb{R}^{I_1\times I_2\times\cdots\times I_N}$, $X=(X_1,X_2,\ldots,X_N)$ and $Y=(Y_1,Y_2,\ldots,Y_N)$. The remarkable property of model \eqref{LNNTC-optim} is that such an optimization problem can be solved in a parallel way, which can efficiently exploit the advantages of modern supercomputers. Some computational results on real world problems also verified that model \eqref{LNNTC-optim} equipped with a simultaneous algorithm performs better than the so-called HaLRTC method in \cite{LMWY13} in terms of taking less computing time.

\subsection{Motivation}
Revisiting both models \eqref{LNNTC-optim0} and \eqref{LNNTC-optim}, it is not difficult to observe that they are matricization methods. It has been documented that matricization methods are efficient in many image/video data sets. However, the matricization for higher-order tensor ignores the nature of tensor, thereby potentially destroying some inherent properties (e.g.,  periodicity and correlation) of the data. Here, we refer the reader to \cite{YZ16} for theoretical analysis. Actually, many real world data sets, e.g.,  the internet traffic data, images and surveillance videos, often appear (approximate) periodicity and smooth prior in the spatial and/or temporal domains, and faces in face recognition share the high self-similarity. Numerically, it has been documented in \cite{ZZXC15,XPWXWCZQ18} that the tensor approaches are more efficient than matrix-based methods \cite{RZWQE12} for internet traffic data, especially for the extreme case when the traffic data on several time intervals are all lost. Moreover, some novel models absorbing the low-rank regularization or total variation regularizer into image/video inpainting models further supported that tensorization methods outperform matricization approaches when the data has tensor structure, e.g., see \cite{JHZML16,LYX17,LZJZJH19}, just to name a few.

Observing that matrix factorization is an efficient way to improve the recovery accuracy of matrix completion \cite{XHYS15}, it also encourages researchers to employ tensor decomposition techniques to improve the performance on tensor recovery, while keeping the nature of tensors. Recently, one of the most popular tensor decomposition technique, i.e., CP decomposition, has been successfully applied to internet traffic data recovery and inference, e.g., \cite{ZZXC15,XPWXWCZQ18}. However, it seems that these tensor models do not fully exploit the traffic periodicity in the traffic data, thereby reducing the recovery accuracy when handling the cases with a low sample ratio.  It is known that the CP decomposition method, which has a wonderful mathematical form, decomposes a tensor as a sum of rank-one tensors. However, it may not be an ideal decomposition for many real world data sets without favorable structures, due to the special form of CP decomposition. Hence, the Tucker decomposition, which is a form of higher-order principal component analysis and decomposes a tensor into a core tensor and some factor matrices, becomes another popular technique in tensor completion. Indeed, the CP decomposition can be regarded as a special Tucker decomposition with setting its core tensor as a superdiagonal tensor. Comparatively, the Tucker decomposition is more attractive than the CP decomposition in the community of image processing, e.g., \cite{SLH19,ZLLZ18,LZJZJH19,LYX17,LMWY13,FJ15}.

\begin{figure}[!htbp]
	\centering
	\includegraphics[width=0.9\linewidth]{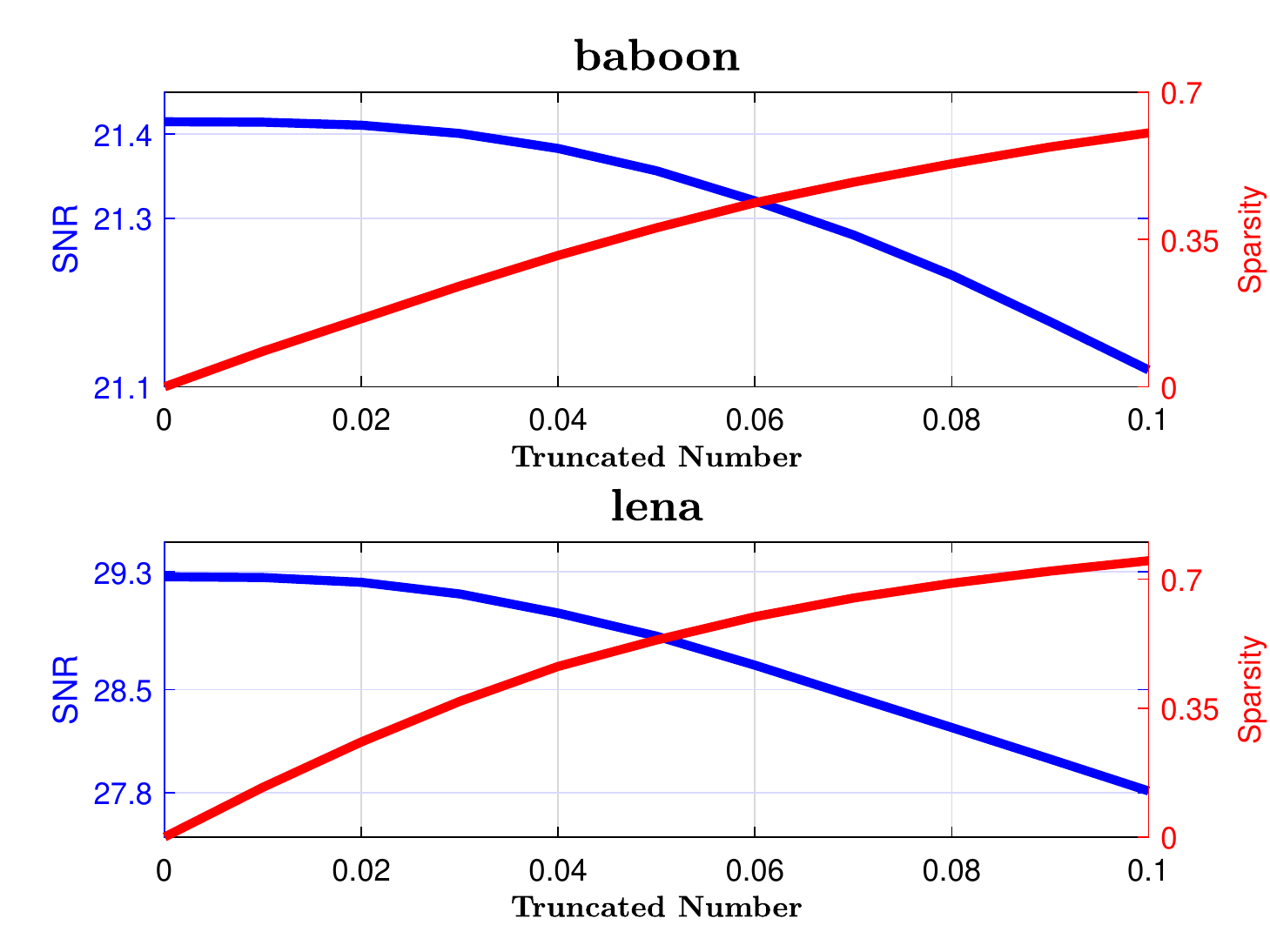}
	\caption{The level of sparsity of core tensors and SNR values of the images recovered by Tucker decomposition with truncated core tensors with respect to truncated numbers.}\label{fig-sparse}
\end{figure}

\begin{figure}[!htbp]
	\centering
	\includegraphics[width=.11\textwidth]{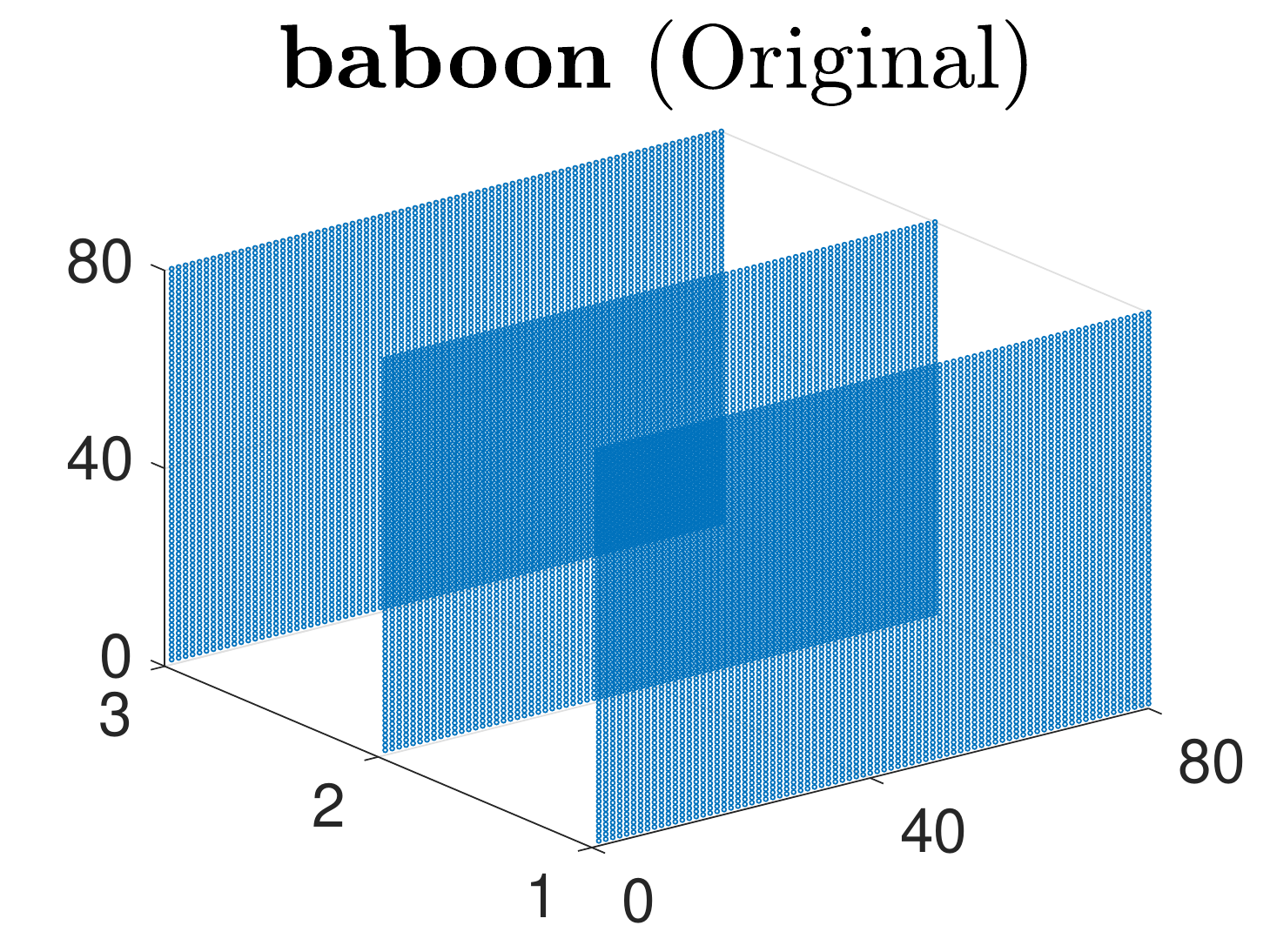}
	\includegraphics[width=.11\textwidth]{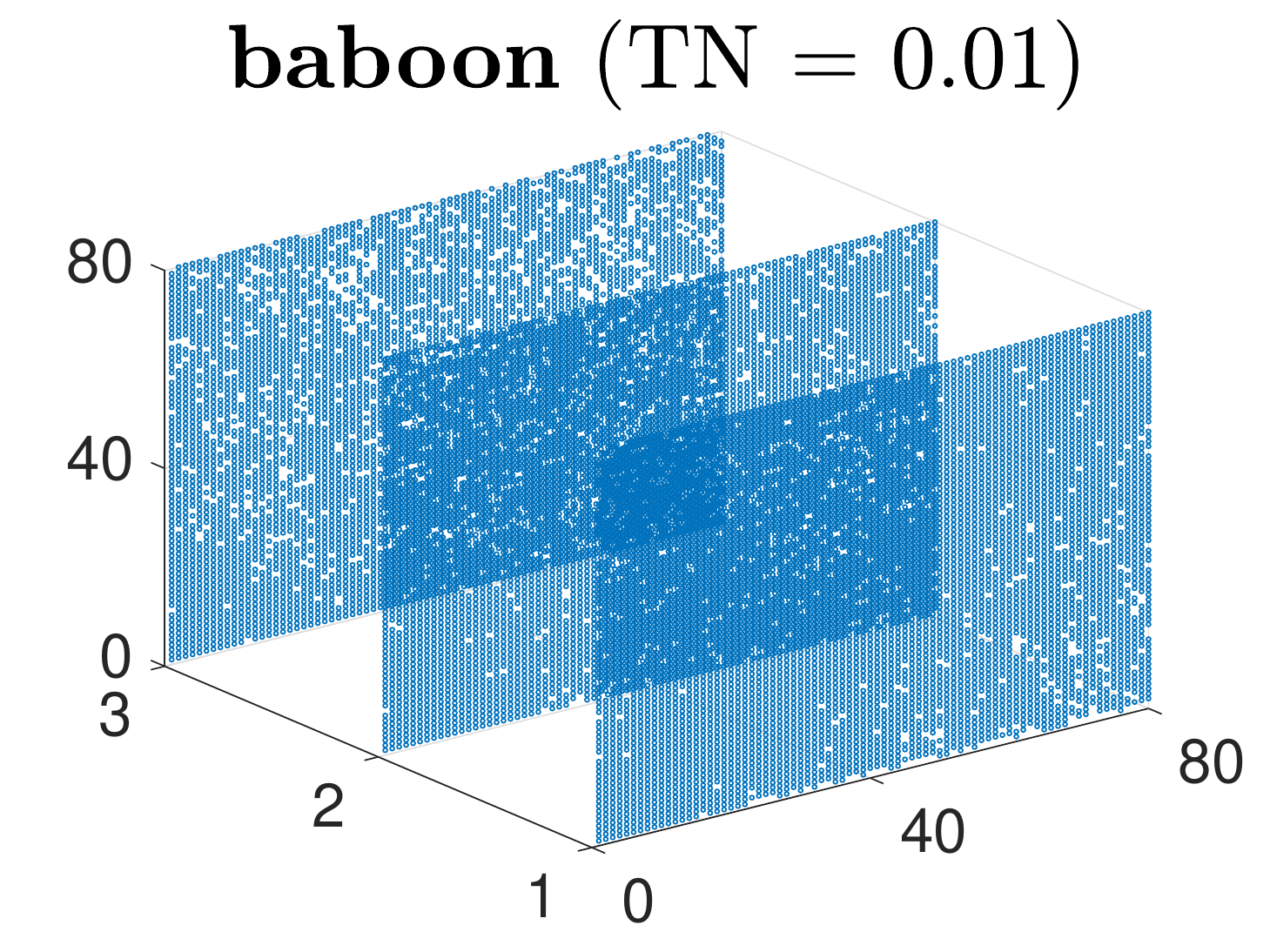}
	\includegraphics[width=.11\textwidth]{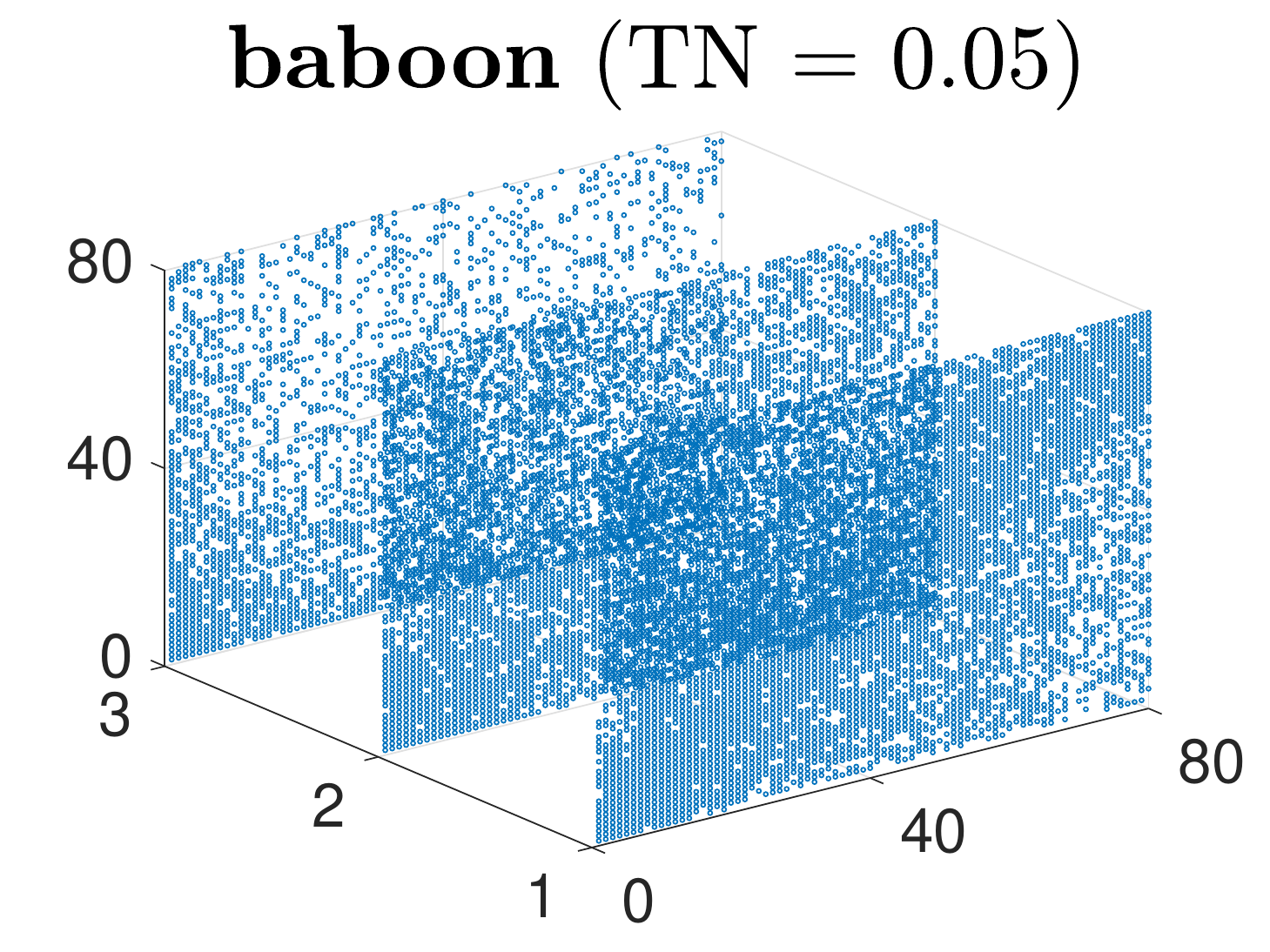}
	\includegraphics[width=.11\textwidth]{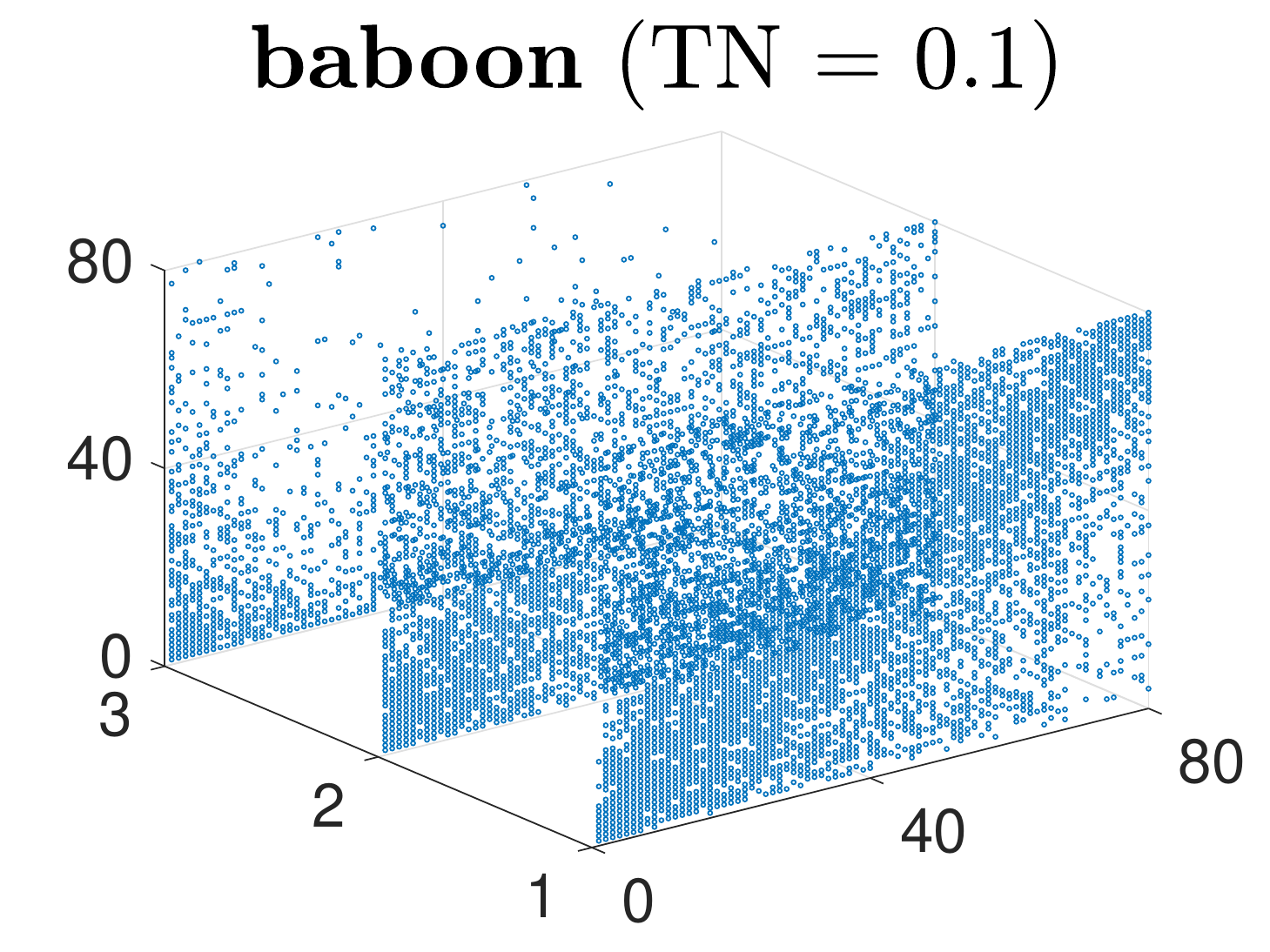}
	\includegraphics[width=.11\textwidth]{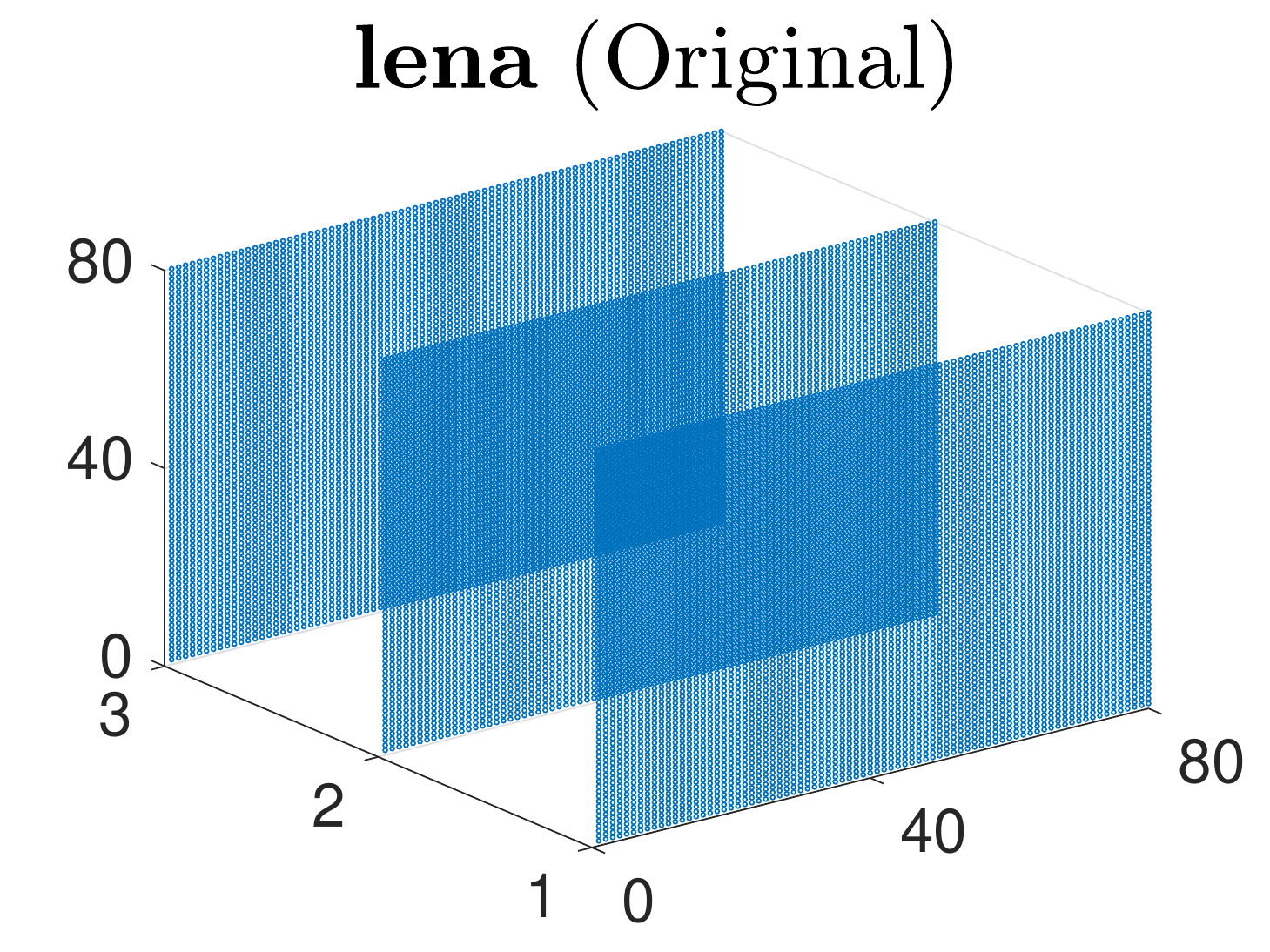}
	\includegraphics[width=.11\textwidth]{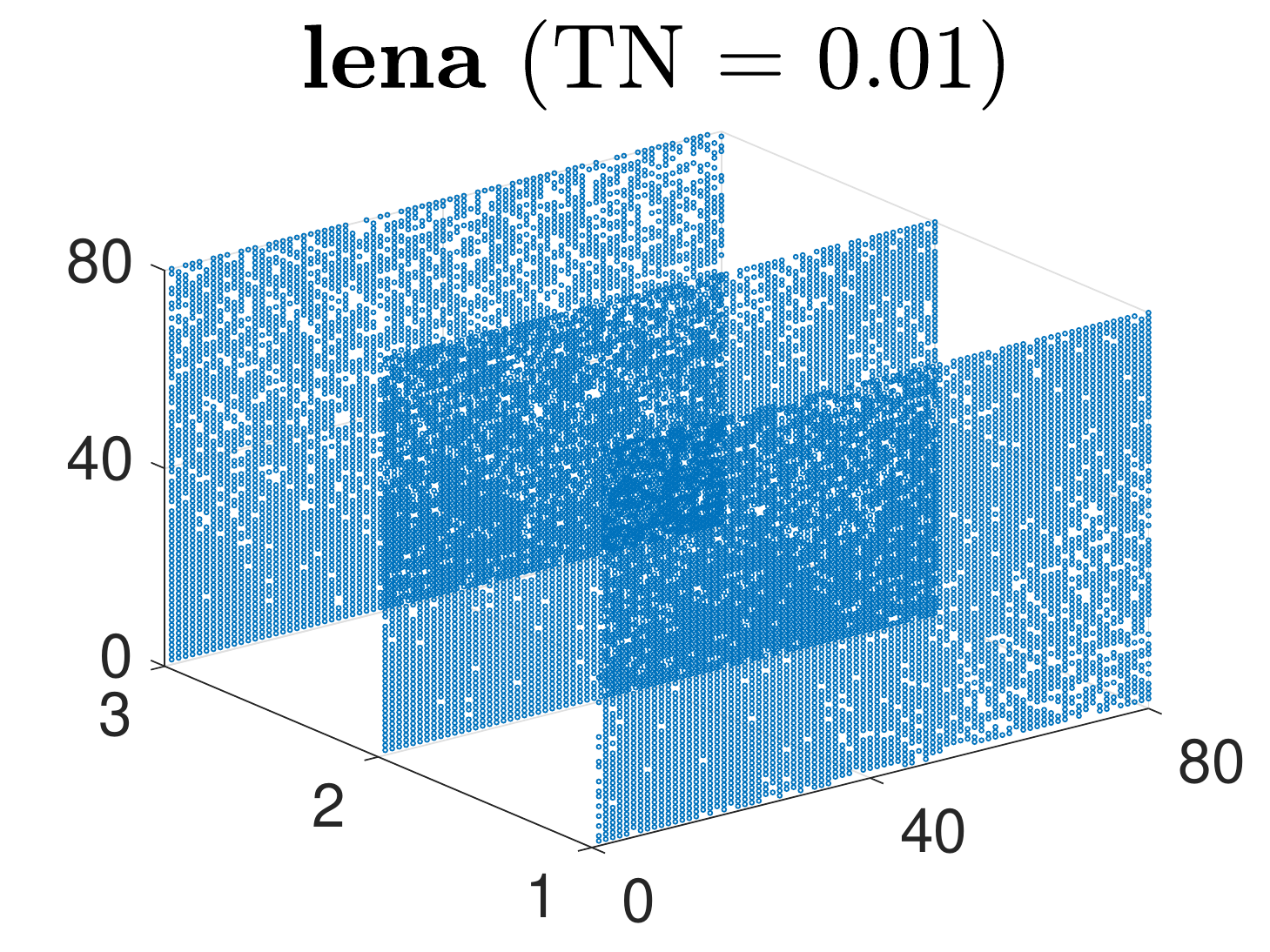}
	\includegraphics[width=.11\textwidth]{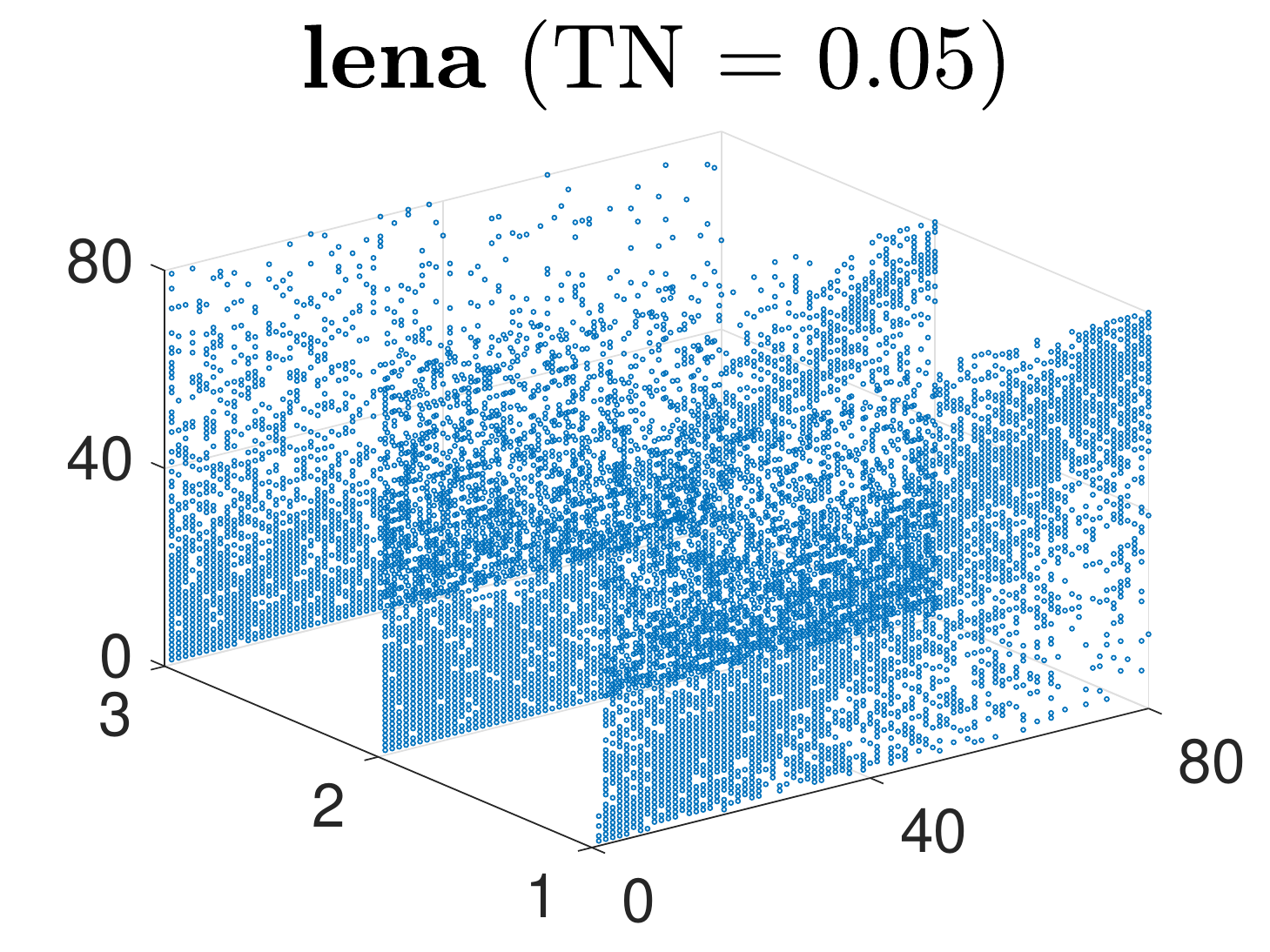}
	\includegraphics[width=.11\textwidth]{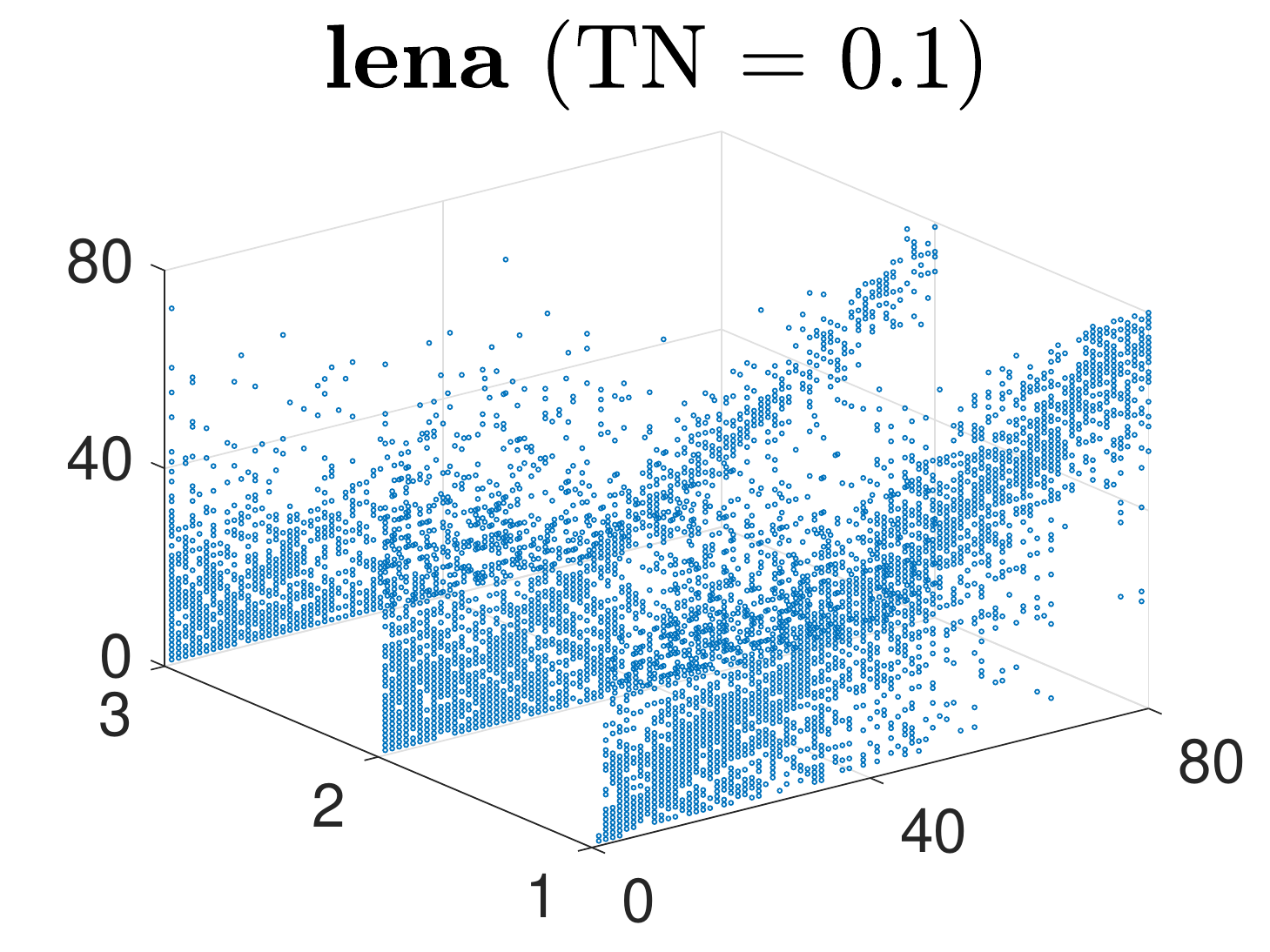}
	\caption{Illustration of the sparse structure of core tensors after a truncation process.}\label{fig-core}
\end{figure}

\begin{figure}[!htbp]
	\centering
	\includegraphics[width=\linewidth]{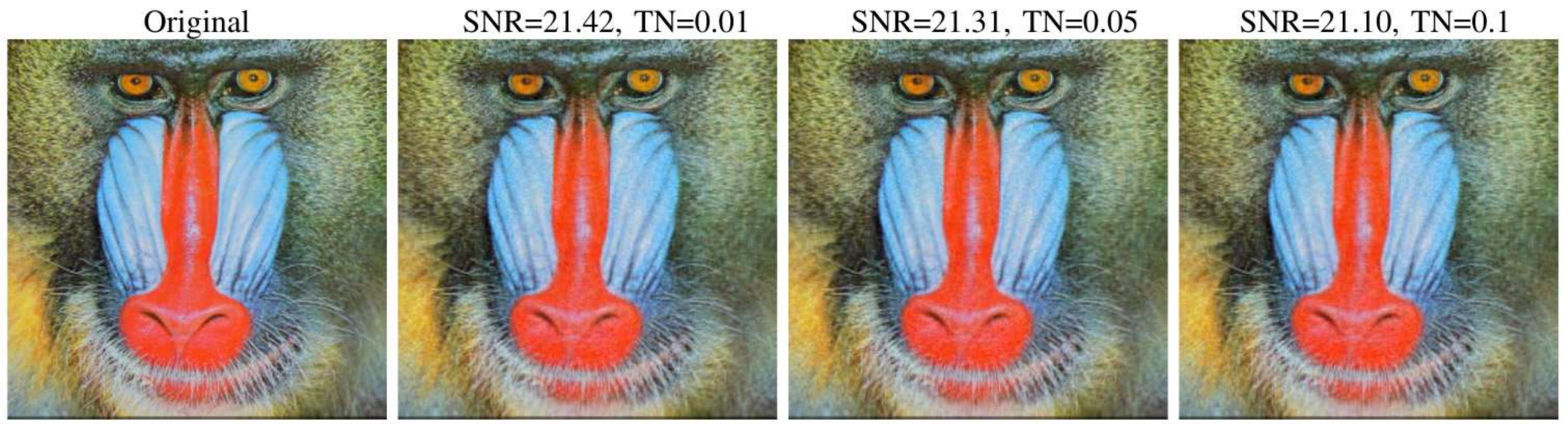}\\
	\includegraphics[width=\linewidth]{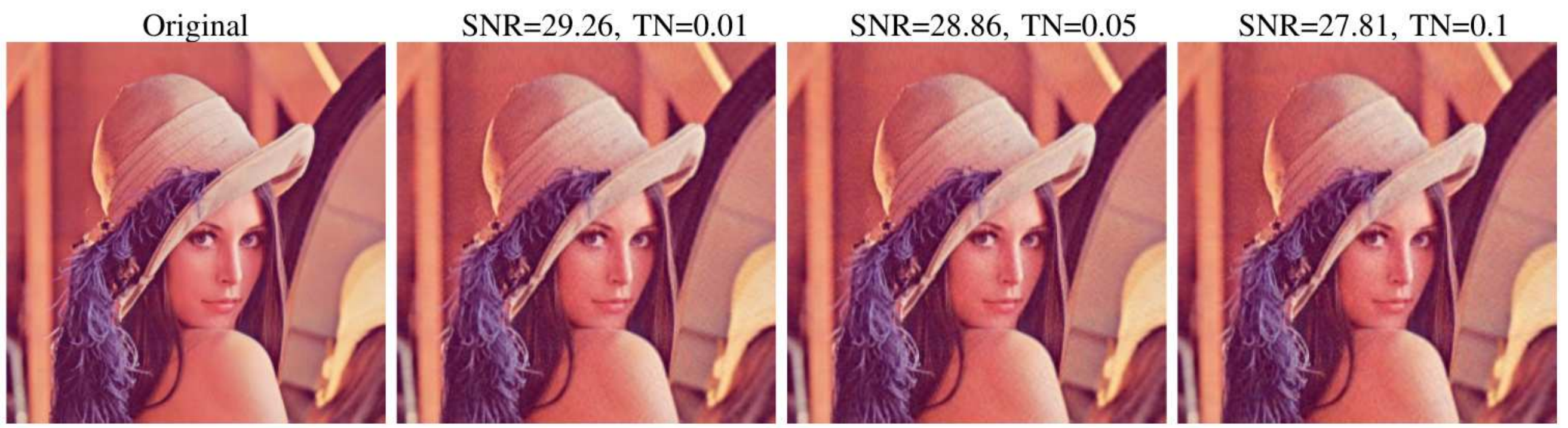}
	\caption{Comparisons of the original images with the ones reconstructed by Tucker decomposition with a truncated core tensor.}\label{fig-demoim}
\end{figure}

However, to the best of our knowledge, most papers did not fully consider the properties, e.g., low-rank and sparsity, of the decomposed core tensor and factor matrices. Considering the widely used two images, i.e., \textbf{baboon} and \textbf{lena} (see Fig. \ref{fig-demoim}), we employ the well-known HOSVD algorithm \cite{DDV00} to each image and obtain a Tucker decomposition. In Fig.  \ref{fig-sparse}, we plot the sparsity levels of the core tensors of the Tucker decomposition with respect to different truncated numbers. Clearly, a larger truncated number (TN) means that more nonzero numbers being smaller than the preset TN are dropped as zeros, thereby leading to a higher level of sparsity. Correspondingly, the SNR value (which is used to measure the quality of the image reconstructed by Tucker decomposition with truncated core tensors) will decrease as the TNs increase. In Fig. \ref{fig-core}, we can intuitively see the sparse structure of the core tensors (which are third-order tensors) for \textbf{baboon} and \textbf{lena} after Tucker decomposition with a truncation process. Especially, when the TN is set as $0.1$, we can see that the levels of sparsity of both core tensors for \textbf{baboon} and \textbf{lena} are higher than $60\%$. However, the SNR values are closed to the case without a truncation. Moreover, we can easily see that the reconstructed images in Fig. \ref{fig-demoim} are almost the same as the original images. Such an observation strongly encourages us to consider a sparse regularization on the core tensor of Tucker decomposition.

On the other hand, it is well-known that there are $N$ factor matrices for the Tucker decomposition of an $N$-order tensor. In the literature, the most popular Tucker rank of a higher-order tensor $\mathcal{F}$ is defined as a vector, whose $n$-th component corresponds to the rank of the mode-$n$ unfolding matrix $F_{(n)}$. However, it is apparent that each mode-$n$ unfolding matrix can be regarded as a large-scale matrix with a large number of rows or/and columns. In this case, those low-Tucker-rank based models, e.g., model \eqref{LNNTC-optim0}, usually take much time to compute the SVDs due to the appearance of nuclear norm inducing the low rankness of the tensor. So, we can naturally image that if the scale of the matrix in SVD could be reduced, we could accordingly save certain computing time for algorithmic acceleration. It is amazing from \cite{KB09} that the rank of the mode-$n$ unfolding matrix is always smaller than the rank of the $n$-th factor matrix of its Tucker decomposition, see also \eqref{rankAX}. Therefore, we are motivated to exploit the property of the factor matrices to promote the low rankness of the tensor. Specifically, we use nuclear norms of these factor matrices to force the low rankness of the unfolding matrices, thereby achieving the goal of inducing a low-rank tensor and saving computing time.

\subsection{Contribution}
In this paper, we make a further study on the Tucker decomposition for tensor completion. Our contributions are four-fold.
\begin{enumerate}
\item We propose a unified tensor completion model based on the Tucker decomposition. Concretely, we use a sparse regularization term to promote the sparsity of the core tensor of the Tucker decomposition. Additionally, we employ low-rank regularization terms to factor matrices of the Tucker decomposition for the purpose of promoting a low-rank tensor. Indeed, the combination of low-rank and sparse terms maximally reflects the inherent correlation property and self-similarity (e.g., image cartoon and texture) appeared in image data sets. In practice, the proposed model has the adaptive ability to deal with different types of real world data sets by tuning regularization parameters, which is verified by a series of numerical experiments.
\item Observing that many tensor data sets, e.g., videos and traffic data sets, have spatio-temporal stability, we incorporate a general Toeplitz matrix into the Tucker decomposition (see $\Psi(X,\mathcal{S})$ in \eqref{eq:optim}) to characterize the hidden structure of the tensor data. Such a model can greatly improve its ability to deal with internet traffic data sets, e.g., see Section \ref{Exp-traffic}.
\item The proposed unified model is a highly nonlinear and nonsmooth optimization problem with coupled variables. If we directly employed traditional optimization approaches, it would suffer from complicated subproblems, thereby reducing the algorithmic implementablity. Therefore, we skillfully introduce some auxiliary variables to decouple the objective function, which is of benefit to propose a customized ADMM to solve the resulting model. It is noteworthy that there possibly exist many ways to employ ADMM to the unified model. However, we believe that the proposed ADMM is enough implementable, since each subproblem is easy enough with a closed-form solution.
\item  To highlight the power of our approach, we investigate a series of computational experiments on internet traffic data, color image inpainting, and face recognition. Computational results demonstrate that our new completion approach has a strongly competitive advantage over many state-of-the-art matricization and tensorization methods, especially when completing the data with a quite low sample ratio or seriously structural missing.
\end{enumerate}

The structure of this paper is as follows: In Section \ref{NarPrel}, we will summarize some notations and basic properties of tensors. In Section \ref{ModAlg}, we first introduce the unified Tucker decomposition based model for tensor completion. Then, we reformulate the new model so that we can solve it by fully exploiting the closed-form expressions of the low-rank and sparse regularization subproblems. In Section \ref{ExpTests}, we apply the proposed model and algorithm to some real world problems including internet traffic recovery, color image inpainting, and face recognition.  Finally, we give some concluding remarks in Section \ref{ConRemark} to complete this paper.

\section{Notations and Preliminaries}\label{NarPrel}
In this section, we recall some notations, definitions and properties on tensors and Tucker decomposition that will be used in the paper.

A tensor is a multi-dimensional array, and the order of a tensor is the number of dimensions, which is also called way or mode. Particularly, vectors and matrices can be regarded as first and second order tensors, respectively. Mathematically, an $N$-order real tensor is denoted by $\mathcal{A}\in \mathbb{R}^{I_1\times I_2\times\cdots\times I_N}$, whose $(i_1, i_2, \ldots, i_N )$-th component is denoted as $a_{i_1i_2\ldots i_N}$. For the sake of notational convenience, in general, higher order (i.e., $N\geq 3$) tensors are denoted by calligraphic letters $\{\mathcal{A}, \mathcal{B},\ldots\}$, capital letters $\{A, B,\ldots\}$ represent matrices, bold-case lowercase letters $\{{\bm a}, {\bm b},\ldots\}$ correspond to vectors, and scalars are denoted by lowercase letters $\{a,b,\ldots\}$. Throughout this paper, given two $N$-order tensors $\mathcal{A},\mathcal{B}\in \mathbb{R}^{I_1\times I_2\times \cdots \times I_N}$, the inner product between $\mathcal{A}$ and $\mathcal{B}$ is defined by $$\langle\mathcal{A},\mathcal{B}\rangle:=\sum_{i_1,i_2,\ldots,i_N}a_{i_1i_2\ldots i_N} b_{i_1i_2\ldots i_N},$$ and the Frobenius norm associated with the above inner product is $\|\mathcal{A}\|_F=\sqrt{\langle\mathcal{A},\mathcal{A}\rangle}$. For every $n\in[N]:=\{1,2,\ldots,N\}$, we denote the mode-$n$ matricization (a.k.a., unfolding or flattening) of an $N$-order tensor $\mathcal{A}$ by $A_{(n)}$, whose $(i_n,j)$-th element in the lexicographical order is mapped from the $(i_1,i_2,\ldots,i_N)$-th entry of tensor $\mathcal{A}$, where
$$
j=1+\sum_{1\leq l\leq N,l\neq n} (i_l-1)J_l~~{\rm with}~~ J_l=\prod_{1\leq t\leq l-1,t\neq n}I_t.
$$
Particularly, for a given third-order tensor $\mathcal{A}\in\mathbb{R}^{I_1\times I_2\times I_3}$, we have its three $n$-mode matrices as follows:
\begin{equation*}
\left\{\begin{array}{l}
	A_{(1)}=[\mathcal{A}_{::1},\mathcal{A}_{::2},\cdots,\mathcal{A}_{::I_3}]; \\
	A_{(2)}=[\mathcal{A}_{::1}^\top,\mathcal{A}_{::2}^\top,\cdots,\mathcal{A}_{::I_3}^\top] ;\\
	A_{(3)}=[\mathcal{A}_{:1:}^\top,\mathcal{A}_{:2:}^\top,\cdots,\mathcal{A}_{:I_2:}^\top],
\end{array}\right.
\end{equation*}
where $\mathcal{A}_{i::}$ and $ \mathcal{A}_{:j:}$ denote the horizontal and lateral slices of tensor $\mathcal{A}$ respectively, and the superscript `$^\top$' represents the transpose of vectors or matrices. Generally, given a mode-$n$ unfolded matrix of tensor $\mathcal{A}$, we can employ the inverse operator named as ``${\rm fold}_n$" to express tensor $\mathcal{A}$, i.e., $\mathcal{A}={\rm fold}_n(A_{(n)})$.

\begin{figure}[!htbp]
	\includegraphics[width=0.48\textwidth]{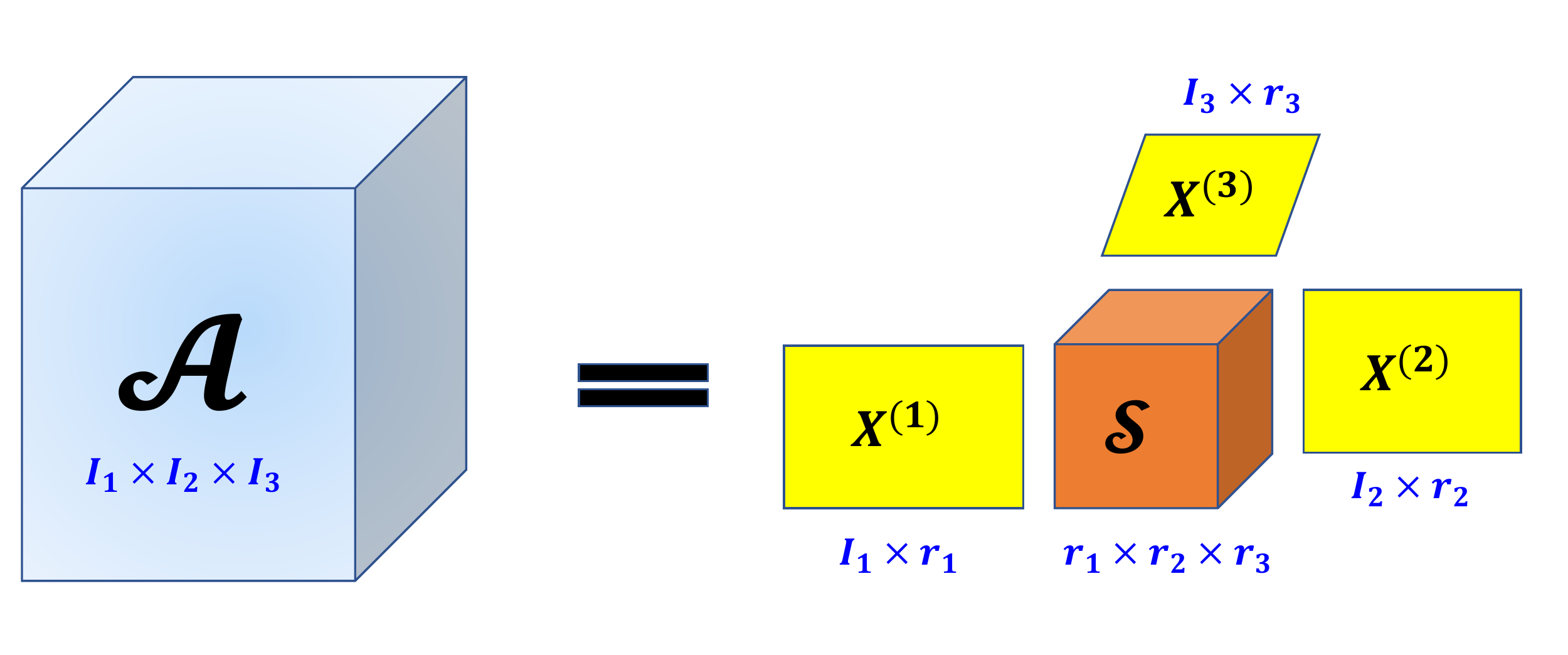}
	\caption{Tucker decomposition for third-order tensors.}\label{tucker}
\end{figure}

Hereafter, we recall the Tucker decomposition \cite{Tucker66}, which is a form of higher-order principal component analysis. Given a tensor $\mathcal{A}\in \mathbb{R}^{I_1\times I_2\times\cdots\times I_N}$, it can be decomposed into a core tensor multiplying a matrix along each mode, i.e.,
\begin{align}\label{Tuckeform}
\mathcal{A} &=\llbracket \mathcal{S};X^{(1)},X^{(2)},\cdots,X^{(N)}\rrbracket \nonumber\\
& := \mathcal{S}\times_1X^{(1)}\times_2X^{(2)}\times_3\cdots \times_NX^{(N)},
%&=\mathcal{S}\times_{i=1}^3X^{(i)}, \\
\end{align}
where $\mathcal{S}\in \mathbb{R}^{r_1\times r_2\times \cdots\times r_N}$ is the so-called core tensor, whose entries show the level of interaction between the different components, and $X^{(n)}\in \mathbb{R}^{I_n\times r_n}~(n\in [N])$ are the factor matrices, which can be viewed as the principal components in each mode.  In particular, if we consider a third-order tensor $\mathcal{A}\in\mathbb{R}^{I_1\times I_2\times I_3}$, the Tucker decomposition can be graphically shown in Fig. \ref{tucker}.
Clearly, if $r_1, r_2, \ldots, r_N$ are significantly smaller than $I_1, I_2,\ldots, I_N$, respectively, then the core tensor $\mathcal{S}$ can be regarded as a compressed version of $\mathcal{A}$, thereby greatly reducing the storage of tensor $\mathcal{A}$. In general, the Tucker decomposition of a tensor is not unique, and it is impossible to obtain a superdiagonal core tensor, even for symmetric tensors, e.g., see \cite{CGLM08}. However, it is possible to get a core tensor being sparse or approximately sparse in the sense that its most elements are small enough, which is helpful for eliminating
interactions between corresponding components and improving the uniqueness of Tucker decomposition.  Recalling the matrix Kronecker product, i.e., for $A=(a_{ij})_{m\times n}$ and $B=(b_{kl})_{p\times q}$, we have
$$
A\otimes B=\left[
\begin{array}{cccc}
a_{11}B&a_{12}B&\cdots&a_{1n}B\\
a_{21}B&a_{22}B&\cdots&a_{2n}B\\
\vdots&\vdots&\ddots&\vdots\\
a_{m1}B&a_{m2}B&\cdots&a_{mn}B\\
\end{array}
\right]_{mp\times nq}.
$$
Consequently, it is well-known from \cite{KB09} that the Tucker decomposition \eqref{Tuckeform} can also be represented by the matricized version as follows:
\begin{align*}
A_{(n)}=X^{(n)}S_{(n)}\big(X^{(N)}\otimes \cdots &\otimes X^{(n+1)} \\
&\otimes X^{(n-1)}\otimes\cdots \otimes X^{(1)}\big)^\top,
\end{align*}
for any $n\in [N]$. As a result, it is easy to see that
\begin{equation}\label{rankAX}
{\rm rank}(A_{(n)})\leq {\rm rank}(X^{(n)}), \quad \forall n\in [N].
\end{equation}

Finally, for a given tensor $\mathcal{A}\in\mathbb{R}^{I_1\times I_2\times\cdots\times I_N}$ and $\Omega\subset [I_1]\times [I_2]\times \cdots\times [I_N]$, we define ${\mathscr P}_{\Omega}(\mathcal{A})$ is a function that keeps the entries of $\mathcal{A}$ in $\Omega$ while making others be zeros, i.e,
$$
({\mathscr P}_{\Omega}(\mathcal{A}))_{i_1i_2\ldots i_N}:=\left\{
\begin{array}{ll}
a_{i_1i_2\ldots i_N},& {\rm if}~(i_1, i_2,\ldots, i_N)\in \Omega,\\
0,&{\rm otherwise}.
\end{array}
\right.
$$

\section{Model and Algorithm}\label{ModAlg}
In this section, we will first introduce the unified tensor completion model equipped with low-rank and sparse enhanced Tucker decomposition. Then, we reformulate the model and employ the state-of-the-art ADMM to solve the resulting model.

\subsection{Enhanced Tucker Decomposition Model}
First, let us introduce the following notation
$$\llbracket \mathcal{S};X^{(1)},\cdots,X^{(N)}\rrbracket \equiv \llbracket \mathcal{S};X^{(i)},X^{(-i)}\rrbracket, \quad \forall i\in[N]$$
for notational brevity. Let $\mathcal{M}\in\mathbb{R}^{I_1\times I_2\times\cdots\times I_N}$ be an observed incomplete tensor and $\Omega$ be the index set corresponding to the observed entries of $\mathcal{M}$.
We consider the following optimization problem
\begin{align}\label{eq:optim}
\min_{X,\mathcal{S}}\;\;&\Psi(X,\mathcal{S})+ \sum_{i=1}^N\alpha_i{\rm rank}(X^{(i)})+\sigma\|\mathcal{S}\|_0 \nonumber\\
\text{s.t.}\;\;&{\mathscr P}_{\Omega}\left( \llbracket \mathcal{S};X^{(i)},X^{(-i)}\rrbracket \right)={\mathscr P}_{\Omega}(\mathcal{M}),
\end{align}
where
\begin{align*}
\Psi(X,\mathcal{S}):=\sum_{i=1}^{N} \omega_i\left\| \llbracket \mathcal{S};A_iX^{(i)},X^{(-i)}\rrbracket\right\|^2_F,
\end{align*}
$X$ is a group of factor matrices, i.e., $X=(X^{(1)},\cdots,X^{(N)})$ with $X^{(i)}\in \mathbb{R}^{I_i\times r_i}$; $\mathcal{S}\in \mathbb{R}^{r_1\times r_2\times \cdots\times r_N}$ is a small-scale core tensor; $\alpha_i\geq 0$ ($i=1,2,\cdots,N$) are regularization parameters balancing each unfolded mode of the given tensor and satisfying $\sum_{i=1}^N\alpha_i\approx 1$; $\sigma>0$ is a tuning parameter for the sparsity of tensor $\mathcal{S}$; $\omega_i\geq 0$ ($i=1,2,\cdots,N$) also are weighted parameters; $A_i$ ($i=1,2,\cdots,N$) are constant matrices which serve as the role of characterizing the feature of the data in $\mathcal{M}$. Generally, we can specify them as identity matrices for simplicity. However, it plays an important role in some specific real world applications, e.g., internet traffic data recovery, thereby greatly improving the recovery accuracy. Throughout the numerical experiments of this paper, we specify each $A_i$ as a special Toeplitz matrix, i.e., $A_i=\text{Toeplitz}(0,1,-1)$, whose central diagonals are ones, the first upper diagonal components are $-1$'s, and the others are zeros. Recalling the relationship between the rank of $A_{(n)}$ and $X^{(n)}$ as shown in \eqref{rankAX}, enforcement of the low-rankness for $X^{(n)}$ will lead to the lower rank of $A_{(n)}$. More importantly, the scale of $X^{(n)}$ is less than $A_{(n)}$, which would greatly reduces the computational complexity of the subproblems.

Due to the appearances of low-rank function and $\ell_0$-norm, the optimization model \eqref{eq:optim} is a highly non-convex problem and NP-hard. Consequently, we are not be able to find efficient solution methods to solve it directly. However, we can fortunately employ the well-known nuclear norm and $\ell_1$ norm to approximate the low-rank function and $\ell_0$ norm (e.g., see \cite{CLMW11,WGRM09}),
respectively. Accordingly, model \eqref{eq:optim} can be relaxed into
\begin{align}\label{lrsetd}
\min_{X,\mathcal{S}}\;\;&\Psi(X,\mathcal{S})+ \sum_{i=1}^N\alpha_i \|X^{(i)}\|_*+\sigma\|\mathcal{S}\|_1 \nonumber\\
\text{s.t.}\;\;&{\mathscr P}_{\Omega}\left( \llbracket \mathcal{S};X^{(i)},X^{(-i)}\rrbracket \right)={\mathscr P}_{\Omega}(\mathcal{M}),
\end{align}
which is a comparatively tractable problem, since the first part $\Psi(X,\mathcal{S})$ of the objective function is differentiable with respect to each block, and the last two terms are convex with nice properties, e.g., their proximal operators have closed-form solutions (e.g., see \cite{ComP11}).

\subsection{Description of Algorithm}
In this subsection, we derive the algorithmic details for model \eqref{lrsetd} with setting $N=3$ for brevity, since our numerical experiments focus on third-order tensor completion problems. When considering higher order cases (i.e., $N\geq 4$), we can easily follow the way here to derive details.

Observing that the objective function is coupled by $\Psi(X,\mathcal{S})$, we accordingly introduce an auxiliary variable $\mathcal{Z}\in\mathbb{R}^{I_1\times I_2\times\cdots\times I_N}$ to separate the three parts of the objective. Concretely, by introducing $\mathcal{Z}=\llbracket \mathcal{S};X^{(i)},X^{(-i)}\rrbracket $, we can rewrite model \eqref{lrsetd} as
\begin{align}\label{optim-ReZ}
\min_{X,\mathcal{S},\mathcal{Z}}\;&\sum_{i=1}^N\omega_i\left\|A_iZ_{(i)}\right\|_F^2+\sum_{i=1}^N\alpha_i\big\|X^{(i)}\big\|_*+\sigma\|\mathcal{S}\|_1 \nonumber\\
\text{s.t.\;\;} \;&\mathcal{Z}=\llbracket \mathcal{S};X^{(i)},X^{(-i)}\rrbracket ,\\
\;&{\mathscr P}_{\Omega}(\mathcal Z)={\mathscr P}_{\Omega}(\mathcal{M}). \nonumber
\end{align}
where $Z_{(i)}$ is the mode-$i$ matricization (or unfolding) of tensor $\mathcal{Z}$ for every $i\in[N]$. It is clear that model \eqref{optim-ReZ} is an equality constrained optimization problem with separable objective function. Usually, we can follow the penalty method to reformulate it as
\begin{align}\label{optim-ReZp}
\min_{X,\mathcal{S},\mathcal{Z}}\;&\sum_{i=1}^N\omega_i\left\|A_iZ_{(i)}\right\|_F^2+\sum_{i=1}^N\alpha_i\big\|X^{(i)}\big\|_*+\sigma\|\mathcal{S}\|_1 \nonumber\\
&\;+\frac{\lambda}{2}\big\|\llbracket \mathcal{S};X^{(i)},X^{(-i)}\rrbracket-\mathcal{Z}\big\|_F^2 \\
\text{s.t.\;\;}\;&{\mathscr P}_{\Omega}(\mathcal Z)={\mathscr P}_{\Omega}(\mathcal{M}), \nonumber
\end{align}
where $\lambda>0$ can be regarded as a penalty parameter. In this situation, to exploit the nice properties of proximal operators for nuclear norm and $\ell_1$ norm, we further introduce auxiliary variables $\mathcal{W}_i\in \mathbb{R}^{I_1\times I_2\times \cdots\times I_N}$ and $Y^{(i)}\in\mathbb{R}^{I_i\times r_i}$ for $i=1,2,\cdots,N$. Consequently, we transform \eqref{optim-ReZp} into the following separable optimization model
\begin{align}\label{optim-ReZS}
\min_{X,Y,\mathcal{S},\mathcal{W},\mathcal{Z}}\;& \sum_{i=1}^N\omega_i\left\|A_iW_{i,(i)}\right\|_F^2+\sum_{i=1}^N\alpha_i\left\|Y^{(i)}\right\|_*+\sigma\|\mathcal{S}\|_1\nonumber\\
\;& +\frac{\lambda}{2}\big\|\llbracket \mathcal{S};X^{(i)},X^{(-i)}\rrbracket-\mathcal{Z}\big\|_F^2, \nonumber\\
\text{s.t. \;\;\;}\; &W_{i,(i)}=Z_{(i)},~i=1,2,\cdots,N,\\
\;&X^{(i)}=Y^{(i)},~i=1,2,\cdots,N, \nonumber\\
\;&{\mathscr P}_{\Omega}(\mathcal Z)={\mathscr P}_{\Omega}(\mathcal{M}). \nonumber
\end{align}
The benefit of reformulation \eqref{optim-ReZS} will be highlighted in the derivation of algorithmic details. Hereafter, for simplicity, we use $Y:=(Y^{(1)},\cdots,Y^{(N)})$ and $\mathcal{W}:=(\mathcal{W}_1,\cdots,\mathcal{W}_N)$ to denote grouped matrices and tensors, respectively.

Taking a close look at model  \eqref{optim-ReZS}, it is a linear equality constrained optimization problem. As we know, the benchmark solver is the augmented Lagrangian method (ALM). By introducing the Lagrangian multipliers $T^{(i)}$ and  $\mathcal{U}_i$ ($i=1,\cdots,N$) to linear constraints, we can get the augmented Lagrangian function of \eqref{optim-ReZS} as follows:
\begin{align}\label{Aug-lagrange}
&\mathscr{L}(X,Y,\mathcal{S},\mathcal{Z},\mathcal{W},\mathcal{U},T) \nonumber\\
&:= \sum_{i=1}^N\omega_i\big\|A_iW_{i,(i)}\big\|_F^2+ \sum_{i=1}^N\alpha_i\big\|Y^{(i)}\big\|_*+\sigma\|\mathcal{S}\|_1\nonumber\\
&~~~~+\frac{\lambda}{2}\big\|\llbracket \mathcal{S};X^{(i)},X^{(-i)}\rrbracket-\mathcal{Z}\big\|_F^2\nonumber\\
&~~~~+\sum_{i=1}^N\left(\big\langle U_{i,(i)},Z_{(i)}-W_{i,(i)}\big\rangle+\big\langle T^{(i)},X^{(i)}-Y^{(i)}\big\rangle\right)\nonumber\\
&~~~~+\frac{\beta}{2}\left\{\sum_{i=1}^N\left(\big\|Z_{(i)}-W_{i,(i)}\big\|_F^2+\big\|X^{(i)}-Y^{(i)}\big\|_F^2\right)\right\}\nonumber\\
&=\sum_{i=1}^N\omega_i\big\|A_iW_{i,(i)}\big\|_F^2+ \sum_{i=1}^N\alpha_i\big\|Y^{(i)}\big\|_*+\sigma\|\mathcal{S}\|_1\nonumber\\
&~~~~+\frac{\lambda}{2}\big\|\llbracket \mathcal {S};X^{(i)},X^{(-i)}\rrbracket-\mathcal{Z}\big\|_F^2\nonumber\\
&~~~~+\sum_{i=1}^N\left(\big\langle \mathcal{U}_{i},\mathcal{Z}-\mathcal{W}_{i}\big\rangle+\big\langle T^{(i)},X^{(i)}-Y^{(i)}\big\rangle\right)\nonumber\\
&~~~~+\frac{\beta}{2}\left\{\sum_{i=1}^N\left(\big\|\mathcal{Z}-\mathcal{W}_{i}\big\|_F^2+\big\|X^{(i)}-Y^{(i)}\big\|_F^2\right)\right\},
\end{align}
where $\beta>0$ is a penalty parameter and the last equality follows from the fact that
$$\langle \mathcal{A},\mathcal{B}\rangle=\langle A_{(n)},B_{(n)}\rangle~\text{and}~\|\mathcal{A}\|_F=\|A_{(n)}\|_F,\;\;\forall n\in [N],$$
for any $\mathcal{A},\mathcal{B}\in \mathbb{R}^{I_1\times I_2\times \cdots \times I_N}$; $T$ and $\mathcal{U}$ correspond to the grouped matrices and tensors, i.e., $T=(T^{(1)},\cdots,T^{(N)})$ and  $\mathcal{U}=(\mathcal{U}_1,\cdots,\mathcal{U}_N)$, respectively. With the help of \eqref{Aug-lagrange}, we can employ the ALM to find an approximate solution for model \eqref{optim-ReZS}. However, the direct application of ALM ignores the favorable separable structure of  \eqref{optim-ReZS}. Therefore, we employ the state-of-the-art ADMM to decouple the variables. Specifically, for the current point $ \left(X_k,Y_k,\mathcal{S}^{k},\mathcal{Z}^{k},\mathcal{W}^{k},\mathcal{U}^{k},T_k\right)$, we generate the next iterate in a sequential order, i.e., $X_{k+1}\to Y_{k+1}\to \mathcal{S}^{k+1}\to\mathcal{Z}^{k+1}\to\mathcal{W}^{k+1}\to\mathcal{U}^{k+1}\to T_{k+1}$, by alternatively minimizing the augmented Lagrangian function. It is noteworthy that here is a fully alternating (or Gauss-Seidel) order in the sense that for each block, e.g., $X=(X^{(1)},X^{(2)},X^{(3)})$, we update via $X^{(1)}_{k+1}\to X^{(2)}_{k+1}\to X^{(3)}_{k+1}$, which can ensure that the algorithm always absorbs the latest information into the update of the next variable. Here, we refer the reader to \cite{BST14,GZ17} for its detailed convergence analysis and skip proofs for conciseness.

Due to the high frequency of third-order tensors in internet traffic data recovery, color image inpainting, and face recognition, we below focus on the algorithmic details for the case where $N=3$ in \eqref{lrsetd}. By setting $N=3$ in model \eqref{lrsetd}, we can specify $\Psi(X,\mathcal{S})$ as follows:
\begin{align*}
\Psi(X,\mathcal{S}):=&\omega_1\| \llbracket \mathcal{S};A_1X^{(1)},X^{(2)},X^{(3)}\rrbracket \|_F^2 \\
&+\omega_2	\| \llbracket \mathcal{S};X^{(1)},A_2X^{(2)},X^{(3)}\rrbracket \|_F^2 \\
&+\omega_3 \| \llbracket \mathcal{S};X^{(1)},X^{(2)},A_3X^{(3)}\rrbracket \|_F^2.
\end{align*}
Under this setting, we shall derive the update schemes for all block variables one by one.

\begin{itemize}
	\item Update $X=(X^{(1)},X^{(2)},X^{(3)})$. As the aforementioned alternating idea, we compute $X_{k+1}$ via
	$$X_{k+1}\in\arg\min \mathscr{L}(X,Y_k,\mathcal{S}^{k},\mathcal{Z}^{k},\mathcal{W}^{k},\mathcal{U}^{k},T_k),$$
	which, by ignoring constant terms, can be simplified as finding an optimal point $X_{k+1}$ of the following optimization problem:
	\begin{align}\label{xsub}
	\min &\; \sum_{i=1}^{3}\left\{ \left\langle T^{(i)}_k, X^{(i)}-Y_k^{(i)}\right\rangle +\left\|X^{(i)}-Y_k^{(i)}\right\|^2_F \right\}\nonumber \\
	&\quad +\frac{\lambda}{2}\left\|\llbracket \mathcal {S}^{k};X^{(1)},X^{(2)},X^{(3)}\rrbracket-\mathcal{Z}^{(k)}\right\|_F^2.
	\end{align}
	Since $X^{(1)}$, $X^{(2)}$, and $X^{(3)}$ are coupled by the last term of \eqref{xsub}, we can further employ the alternating technique to find $X_{k+1}$ via $X^{(1)}_{k+1}\to X^{(2)}_{k+1} \to X^{(3)}_{k+1}$ by fixing the other two variables. Specifically, we first find $X^{(1)}_{k+1}$ via solving the following optimization problem:
	\begin{align}\label{X1-k}
	\min_{X^{(1)}%\in \mathbb{R}^{I_1\times r_1}
	}&\left\langle T_k^{(1)},X^{(1)}-Y_k^{(1)}\right\rangle+\displaystyle\frac{\beta}{2}\big\|X^{(1)}-Y_k^{(1)}\big\|_F^2 \nonumber\\
	&\quad+\frac{\lambda}{2}\left\|\llbracket \mathcal {S}^{k};X^{(1)},X_k^{(2)},X_k^{(3)}\rrbracket-\mathcal{Z}^{(k)}\right\|_F^2.
	\end{align}
	By a direct computation, we know that the solution of the model (\ref{X1-k}) can be expressed by
	\begin{align}\label{X1-k-Sol}
	X_{k+1}^{(1)}=&\left\{\lambda Z_{(1)}^{k}B_{1k}\big(S_{(1)}^{k}\big)^\top+\beta Y_k^{(1)}-T_k^{(1)}\right\} \nonumber\\
	&\left[\beta I+\lambda S_{(1)}^{k}B_{1k}^\top B_{1k} \big(S_{(1)}^{k}\big)^\top\right]^{-1}
	\end{align}
	with $B_{1k}:=X_k^{(3)}\otimes X_k^{(2)}.$ Then, by absorbing $X_{k+1}^{(1)}$ into the update of $X^{(2)}$, we obtain $X_{k+1}^{(2)}$ via solving
	\begin{align*}
	\min_{X^{(2)}}&\left\langle T_k^{(2)},X^{(2)}-Y_k^{(2)}\right\rangle+\frac{\beta}{2}\big\|X^{(2)}-Y_k^{(2)}\big\|_F^2 \nonumber\\
	&\;\;+\frac{\lambda}{2}\left\|\llbracket \mathcal {S}^{k};X^{(1)}_{k+1},X^{(2)},X_k^{(3)}\rrbracket-\mathcal{Z}^{(k)}\right\|_F^2,
	\end{align*}
	which implies that we can update $X_{k+1}^{(2)}$ via
	\begin{align}\label{X2-k-Sol}
	X_{k+1}^{(2)}=&\left\{\lambda Z_{(2)}^{k}B_{2k}\big(S_{(2)}^{k}\big)^\top+\beta Y_k^{(2)}-T_k^{(2)}\right\} \nonumber\\
	&\left[\beta I+\lambda S_{(2)}^{k}B_{2k}^\top B_{2k} \big(S_{(2)}^{k}\big)^\top\right]^{-1}
	\end{align}
	with $B_{2k}:=X_k^{(3)}\otimes X_{k+1}^{(1)}$. Finally, we find an optimal solution $X_{k+1}^{(3)}$ to
	\begin{align*}
	\min_{X^{(3)}}&\left\langle T_k^{(3)},X^{(3)}-Y_k^{(3)}\right\rangle+ \frac{\beta}{2}\big\|X^{(3)}-Y_k^{(3)}\big\|_F^2 \nonumber\\
	&\;\;+\frac{\lambda}{2}\left\|\llbracket \mathcal {S}^{k};X^{(1)}_{k+1},X^{(2)}_{k+1},X^{(3)}\rrbracket-\mathcal{Z}^{k}\right\|_F^2,
	\end{align*}
	which has a unique solution given by
	\begin{align}\label{X3-k-Sol}
	X_{k+1}^{(3)}=& \left\{\lambda Z_{(3)}^{k}B_{3k}\big(S_{(3)}^{k}\big)^\top+\beta Y_k^{(3)}-T_k^{(3)}\right\}\nonumber\\
	&\left[\beta I+\lambda S_{(3)}^{k}B_{3k}^\top B_{3k} \big(S_{(3)}^{k}\big)^\top\right]^{-1}
	\end{align}
	with $B_{3k}:=X_{k+1}^{(2)}\otimes X_{k+1}^{(1)}$, respectively.
	
	\item Update $Y=(Y^{1},Y^{2},Y^{3})$. After the computation of $X_{k+1}$, we immediately absorb it into the update of $Y_{k+1}$, i.e., finding $Y_{k+1}$ such that
	$$Y_{k+1}\in\arg\min \mathscr{L}(X_{k+1},Y,\mathcal{S}^{k},\mathcal{Z}^{k},\mathcal{W}^{k},\mathcal{U}^{k},T_k).$$
	By exploiting the separable structure of the underlying $Y$-subproblem, we can find $Y_{k+1}=(Y^{(1)}_{k+1},Y^{(2)}_{k+1},Y^{(3)}_{k+1})$ simultaneously via solving
	\begin{align*}
	\min_{Y^{(i)}}&\;\;\alpha_i\big\|Y^{(i)}\big\|_*+\big\langle T_k^{(i)},X_{k+1}^{(i)}-Y^{(i)}\big\rangle \nonumber\\
	&\;\;+\frac{\beta}{2}\big\|X_{k+1}^{(i)}-Y^{(i)}\big\|_F^2, \quad (i=1,2,3),
	\end{align*}
	which can be rewritten as
	\begin{equation}\label{Si-k}
	\min_{Y^{(i)}}\left\{\alpha_i\big\|Y^{(i)}\big\|_*+\frac{\beta}{2}\left\|Y^{(i)}-X_{k+1}^{(i)}-\frac{1}{\beta}T_k^{(i)}\right\|_F^2\right\}.
	\end{equation}
	By invoking \cite[Theorem 2.1]{CCS10}, the subproblem \eqref{Si-k} has a closed-form solution given by
	\begin{equation}\label{Si-k-Sol}
	Y^{(i)}_{k+1}=\mathscr{D}_{\frac{\alpha_i}{\beta}}\left(X^{(i)}_{k+1}+\frac{1}{\beta}T_k^{(i)}\right), \;\; (i=1,2,3),
	\end{equation}
	where $\mathscr{D}_\tau(\cdot)$ is the well-known singular value shrinkage operator $\mathscr{D}_\tau(M)=U\mathscr{D}_\tau(\Sigma)V^\top$ with $M=U\Sigma V^\top $ being the SVD of a matrix $M\in \mathbb{R}^{m\times n}$ of rank $r$ in the reduced form, i.e., $\Sigma = {\rm diag}\left(\{\sigma_i\}_{1\leq i \leq r}\right)$
	and $U$ and $V$ are, respectively, $m\times r$ and $n\times r$  matrices with orthonormal columns, and $\mathscr{D}_\tau(\Sigma) = {\rm diag}\left(\max\{\sigma_i -\tau,0  \} \right)$ for $\tau\geq 0$.
	
	\item Update the variable $\mathcal{S}$. We compute $\mathcal{S}^{k+1}$ such that
	$$\mathcal{S}^{k+1}\in\arg\min\mathscr{L}(X_{k+1},Y_{k+1},\mathcal{S},\mathcal{Z}^{k},\mathcal{W}^{k},\mathcal{U}^{k},T_k).$$
	By ignoring constant terms, we immediately can simplify the $\mathcal{S}$-subproblem as
	\begin{equation*}
	\min_{\mathcal{S}
	}~\sigma\|\mathcal{S}\|_1+\frac{\lambda}{2}\left\|\llbracket \mathcal {S};X^{(1)}_{k+1},X^{(2)}_{k+1},X_{k+1}^{(3)}\rrbracket-\mathcal{Z}^{k}\right\|_F^2,
	\end{equation*}
	which can be expressed as
	\begin{equation}\label{EEiY-k}
	\min_{S_{(1)}}\frac{\sigma}{\lambda}\big\|S_{(1)}\big\|_1+\frac{1}{2}\left\|X_{k+1}^{(1)}S_{(1)}B_{k}-Z_{(1)}^{k}\right\|_F^2,
	\end{equation}
	where $S_{(1)}\in \mathbb{R}^{r_1\times r_2r_3}$ and $B_k=(X_{k+1}^{(3)}\otimes X_{k+1}^{(2)})^\top$. Due to the presence of $X_{k+1}^{(1)}$ and $B_k$, we cannot obtain the closed-form solution of \eqref{EEiY-k} directly. In this situation, letting
	$$\phi_k(S_{(1)})=\frac{1}{2}\big\|X_{k+1}^{(1)}S_{(1)}B_k-Z_{(1)}^{k}\big\|_F^2,$$
	we can exploit the smoothness of $\phi_k(S_{(1)})$ to linearize it, thereby gain fully deriving an explicit form (e.g., see \cite{HYZ08}) to update $S^{k+1}_{(1)}$ via
	\begin{align}\label{pgm}
	S_{(1)}^{k+1}& = \text{shrink}\left(S^{k}_{(1)}-\frac{1}{\zeta_k}\nabla \phi_k(S^{k}_{(1)}), \frac{\sigma}{\lambda\zeta_k}\right),
	\end{align}
	where $\nabla \phi_k(S_{(1)})$ is the gradient given by
	$$\nabla \phi_k(S_{(1)})=\widehat{X}^{(1)}_{k+1}S_{(1)}\widehat{B}_k-\big(X_{k+1}^{(1)}\big)^\top Z_{(1)}^{k}B_k^\top$$
	with  $\widehat{X}^{(1)}_{k+1}=\big(X_{k+1}^{(1)}\big)^\top X_{k+1}^{(1)}$ and $\widehat{B}_k=B_kB_k^\top$; $\zeta_k=\big\|\widehat{X}^{(1)}_{k+1}\big\|_2\big\|\widehat{B}_k\big\|_2$ is the Lipschitz constant of the gradient $\nabla \phi_k(S_{(1)})$ and $\|\cdot\|_2$ denotes the standard spectral norm for matrices;  the `${\rm shrink}(\cdot,\cdot)$' is the shrinkage operator in component-wise, i.e.,
	$$
	\left(\text{shrink}(G,\tau)\right)_{ij}=\text{sign}(g_{ij})\cdot\left(\max\left\{|g_{ij}|-\tau,0\right\}\right)
	$$
	for a given matrix $G=(g_{ij})\in\mathbb{R}^{m\times n}$. By invoking the `${\rm fold}_n$' operator, we can update $\mathcal{S}^{k+1}$ via
	\begin{equation}\label{maclSk}
	\mathcal{S}^{k+1}={\rm fold}_1\left(S_{(1)}^{k+1}\right).
	\end{equation}
	
	\item Update the variable $\mathcal{Z}$. After obtaining $X_{k+1}$, $Y_{k+1}$ and $\mathcal{S}^{k+1}$, we find a new $\mathcal{Z}^{k+1}$ satisfying
	$$ \mathcal{Z}^{k+1}\in\arg\min\mathscr{L}(X_{k+1},Y_{k+1},\mathcal{S}^{k+1},\mathcal{Z},\mathcal{W}^{k},\mathcal{U}^{k},T_k),$$
	which amounts to finding $\mathcal{Z}^{k+1}$ to minimize
	\begin{align*}
	\Phi_k(\mathcal{Z})=\frac{\lambda}{2}\left\| \widehat{\mathcal{Z}}^{k+1} - \mathcal{Z}\right\|_F^2+\frac{\beta}{2}\sum_{i=1}^{3}\left\|\mathcal{Z}-\mathcal{W}_i^k+\frac{1}{\beta}\mathcal{U}_i^k\right\|_F^2 ,
	\end{align*}
	where
	\begin{equation*}
	\widehat{\mathcal{Z}}^{k+1}=\llbracket \mathcal {S}_{k+1};X^{(1)}_{k+1},X^{(2)}_{k+1},X_{k+1}^{(3)}\rrbracket.
	\end{equation*}
	Consequently, we have
	\begin{equation}\label{zupd}
	\mathcal{Z}^{k+1}=\displaystyle\frac{1}{\lambda+3\beta}\left\{\sum_{i=1}^3\left(\beta \mathcal{W}_i^{k}-\mathcal{U}_i^{k}\right)+\lambda \widehat{\mathcal{Z}}^{k+1}\right\}.
	\end{equation}
	Combining with the constraint ${\mathscr P}_{\Omega}(\mathcal{Z})={\mathscr P}_{\Omega}(\mathcal{M})$, the final update of $\mathcal{Z}^{k+1}$ reads as
	\begin{equation}\label{Zk+1}
	{\mathscr P}_{\Omega}(\mathcal{Z}^{k+1})={\mathscr P}_{\Omega}(\mathcal{M}).
	\end{equation}
	
	\item Update variables $\mathcal{W}_i$'s. To update $\mathcal{W}^{k+1}$, it amounts to solving
	$$ \min\mathscr{L}(X_{k+1},Y_{k+1},\mathcal{S}^{k+1},\mathcal{Z}^{k+1},\mathcal{W},\mathcal{U}^{k},T_k).$$
	By the fully separable nature of $\mathcal{W}_i$'s, we can also update $\mathcal{W}_i^{k+1}$ simultaneously, i.e., solving
	\begin{equation}\label{Wi-k}
	\min_{\mathcal{W}_i}\omega_i\left\|A_iW_{i,(i)}\right\|_F^2+
	\frac{\beta}{2}\left\|\mathcal{W}_{i}-\mathcal{Z}^{k+1}-\frac{1}{\beta}\mathcal{U}_i^k\right\|_F^2.
	\end{equation}
	Exploiting the smoothness of \eqref{Wi-k} immediately yields
	\begin{equation*}
	W^{k+1}_{i,(i)}=\left[\beta I+2\omega_iA_i^\top A_i\right]^{-1}\left[\beta Z^{k+1}_{(i)}+U^{k}_{i,(i)}\right].
	\end{equation*}
	By invoking the `fold' operator, we can obtain
	\begin{equation}\label{Wik-ksol}
	\mathcal{W}_i^{k+1}={\rm fold}_i\left(W^{k+1}_{i,(i)}\right), \quad i=1,2,3.
	\end{equation}
	
	\item Update variables $\mathcal{U}_i$'s. Since $\mathcal{U}_i$'s are Lagrangian multipliers associated to the linear constraints $W_{i,(i)}=Z_{(i)}$, we can update $\mathcal{U}_i^{k+1}$ via
	\begin{equation}\label{Ui-k}
	\mathcal{U}_i^{k+1}=\mathcal{U}_i^{k}+\beta\left(\mathcal Z^{k+1}-\mathcal{W}_i^{k+1}\right),\;\;i=1,2,3.
	\end{equation}
	
	\item Update variables $T^{(i)}$'s. Note that $T^{(i)}$'s are also dual variables corresponding to the linear constraints $X^{(i)}=Y^{(i)}$ for $i=1,2,3$. We accordingly update $T^{(i)}_{k+1}$ via
	\begin{equation}\label{Ti-k}
	T^{(i)}_{k+1}=T^{(i)}_{k}+\beta\left(X^{(i)}_{k+1}-Y^{(i)}_{k+1}\right), \;\; i=1,2,3.
	\end{equation}
\end{itemize}
With the above preparation, we can formally summarize the updating schemes for model \eqref{optim-ReZS} in Algorithm \ref{alg_ADMM}.

\begin{algorithm}[!htbp]
	\caption{ADMM for Tensor Completion Model \eqref{optim-ReZS}.}\label{alg_ADMM}
	\begin{algorithmic}[1]
		\renewcommand{\algorithmicrequire}{\textbf{Input:}}
		\renewcommand{\algorithmicensure}{\textbf{Output:}}
		\REQUIRE Starting points $X_0$, $Y_0$, $\mathcal{S}^0$, $\mathcal{Z}^0$, $\mathcal{W}^0$, $\mathcal{U}^0$, $T_0$.
		\STATE  Update $X_{k+1}^{(i)}$ sequentially via \eqref{X1-k-Sol}, \eqref{X2-k-Sol}, and \eqref{X3-k-Sol};
		\STATE  Update $Y_{k+1}^{(i)}$ simultaneously via \eqref{Si-k-Sol} for $i=1,2,3$;
		\STATE  Update $\mathcal{S}^{k+1}$ via \eqref{pgm} and \eqref{maclSk};		
		\STATE  Update $\mathcal{Z}^{k+1}$ via \eqref{zupd} and \eqref{Zk+1};
		\STATE  Update $\mathcal{W}^{k+1}_i$ simultaneously via \eqref{Wik-ksol} for $i=1,2,3$;
		\STATE  Update $\mathcal{U}^{k+1}_i$ simultaneously via \eqref{Ui-k} for $i=1,2,3$;
		\STATE  Update $T_{k+1}^{(i)}$ simultaneously via \eqref{Ti-k} for $i=1,2,3$.
		\ENSURE  Recovered tensor $\mathcal{Z}^*$.
	\end{algorithmic}
\end{algorithm}
It is clear from Algorithm \ref{alg_ADMM} that we can easily extend such an algorithm to deal with higher order cases where $N> 3$. Moreover, it is an implementable algorithm in the sense that we can update each variable via an explicit iterative scheme.

\section{Experiments Results}\label{ExpTests}
In this section, we will investigate the performance of the proposed model and algorithm on three real world tensor recovery problems including traffic data recovery, color image inpainting and face recognition. By the core ideas of our model and algorithm, we denote our approach to tensor completion by LR-SETD for brevity throughout our experiments.

\subsection{Traffic Data Recovery}\label{Exp-traffic}
In this subsection, we apply our LR-SETD approach on traffic flow data recovery. To fully exploit the traffic features of periodicity pattern in traffic data, we usually organize the traffic data into a third order tensor $\mathcal{Z}\in\mathbb{R}^{O\times T\times D}$ in our model, where $O$ corresponds to $n^2$ origin and destination (OD) pairs, and there are $D$ days to consider with each day having $T$ time intervals. In this situation, we can specify $I_1=O$, $I_2=T$ and $I_3=D$ in model \eqref{optim-ReZ}.

As used in the literature, we also use the normalized mean absolute error (NMAE) in the missing values to measure the quality of the recovered data by models and algorithms. Specifically, the NMAE is defined as follows
$$
{\rm NMAE}:= \frac{\sum_{(i,j,l)\notin \Omega} |(\mathcal{Z}_{\rm true})_{ijl}-\mathcal{Z}_{ijl} |}{\sum_{(i,j,l)\notin \Omega}|(\mathcal{Z}_{\rm true})_{ijl}|},
$$
where $\mathcal{Z}_{\rm true}$ and $\mathcal{Z}$ represent original data and recovered tensor, respectively. Clearly, lower NMAE value means better quality of the recovered data. As a simulation purpose, we evaluate the recovery performance of our approach LR-SETD for two scenarios on the missing data: (i) randomly missing; (ii) structurally missing. Concretely, the missing data in the traffic matrix (TM) of the former scenario is randomly and uniformly distributed, while the latter has all the data missing in a certain time interval but the data observed in other time intervals are randomly and evenly missing. To verify the performance of our approach LR-SETD, we compare it with some state-of-the-art completion methods proposed in the literature. The first one is matrix completion by the nuclear norm minimization (IST\_MC for short, see \cite{CR09,Maj20}), which does not consider the spatio-temporal structure of the traffic matrix. The second one is sparsity regularized matrix factorization (SRMF) method \cite{RZWQE12}, which is a low-rank matrix completion approach with a spatio-temporal regularization. The third one is the low-rank tensor factorization method without spatio-temporal constraints (TCTF\footnote{https://panzhous.github.io/}, see \cite{ZLLZ18}). The fourth method is spatio-temporal tensor completion method (STTC) proposed in \cite{ZZXC15}. The fifth approach is a scalable tensor factorization method (CPWOPT\footnote{https://gitlab.com/tensors/tensor\_toolbox}) introduced in \cite{ADKM11}.

We adopt the relative change of the two successive recovered tensors, i.e.,
\begin{equation}\label{StopCrit}
\frac{\|\mathcal{Z}^{k+1}-\mathcal{Z}^{k}\|_F}{\|\mathcal{Z}_{\rm true}\|_F} \leq {\rm Tol},
\end{equation}
as the stopping criterion for all methods, where $\mathcal{Z}_{\rm true}$ is the original tensor. For the parameters emerged in our model \eqref{optim-ReZ}, we set  $\sigma = 1$, $\lambda= 10^{-2}$, $\alpha_i =1/3, i=1,2,3$, and $(\omega_1,\omega_2,\omega_3) = (0,1,2\times10^{-3})$ for random missing scenario, $(\omega_1,\omega_2,\omega_3)= (0,1,1)$ for whole day missing scenario(missing data Id $\in \{11-15\}$).  For the other algorithms, we keep the default parameters used in their codes. Throughout this section, we take ${\rm Tol}=10^{-5}$ and set the maximum number of iterations as ${\rm MaxIter}= 250$ for all algorithms.

We first consider the problem of internet data recovery and perform the aforementioned methods on two widely used data sets including Abilene dataset\footnote{Abilene: http://abilene.internet2.edu/observatory/data-collections.html} and G\'{E}ANT dataset \cite{UQLB06}. In Abilene dataset, there are $N=11$ routers. For each OD pair, a count of network traffic flow is recorded for every $5$ minutes in a week from Dec 8, 2003 to Dec 14, 2003. Thus, there are $24\times(60/5)=288$ traffic values for each OD pair in one day. When organizing the data as a tensor, we can consider two ways: (i) OOT tensor of size ${11 \times 11 \times 2016}$, where the first mode stands for $11$ source routers, the second mode means $11$ destination routers, and the third mode represents $2016$ time intervals; (ii) OTD tensor of size ${121\times 288\times 7}$, where the first mode stands for $121$ OD pairs, the sencond mode means $288$ time interval in a day, and the final mode corresponds to $7$ days. However, when applying matrix based models to solve traffic recovery problem, since the total number of observations is $288\times 7=2016$ for each OD pair, the corresponding traffic data can be modeled as a matrix with size ${121\times 2016}$. In  G\'{E}ANT  dataset \cite{UQLB06},  there are $23$ routers and $529$ OD pairs. For each OD pair, a count of network traffic flow is recorded for every $15$ minutes in a day. Like Abilene data, we have also two ways to organize the data as a tensor: (i) OOT tensor of size ${23\times23\times672}$, where the three modes correspond to source mode, destination mode and time mode, respectively. (ii) OTD tensor of size ${529\times96\times7}$ is same as the setting used in Abilene data. When applying to matrix based models, we can set a traffic data matrix with size $529\times 672$. Here, it is noteworthy that our approach LR-SETD can handle the two types of tensors, i.e., OOT and OTD. So, we report the results of LR-SETD by LR-SETD(OOT) and LR-SETD(OTD), respectively.

\begin{figure}
	\centering
	\includegraphics[width=0.24\textwidth]{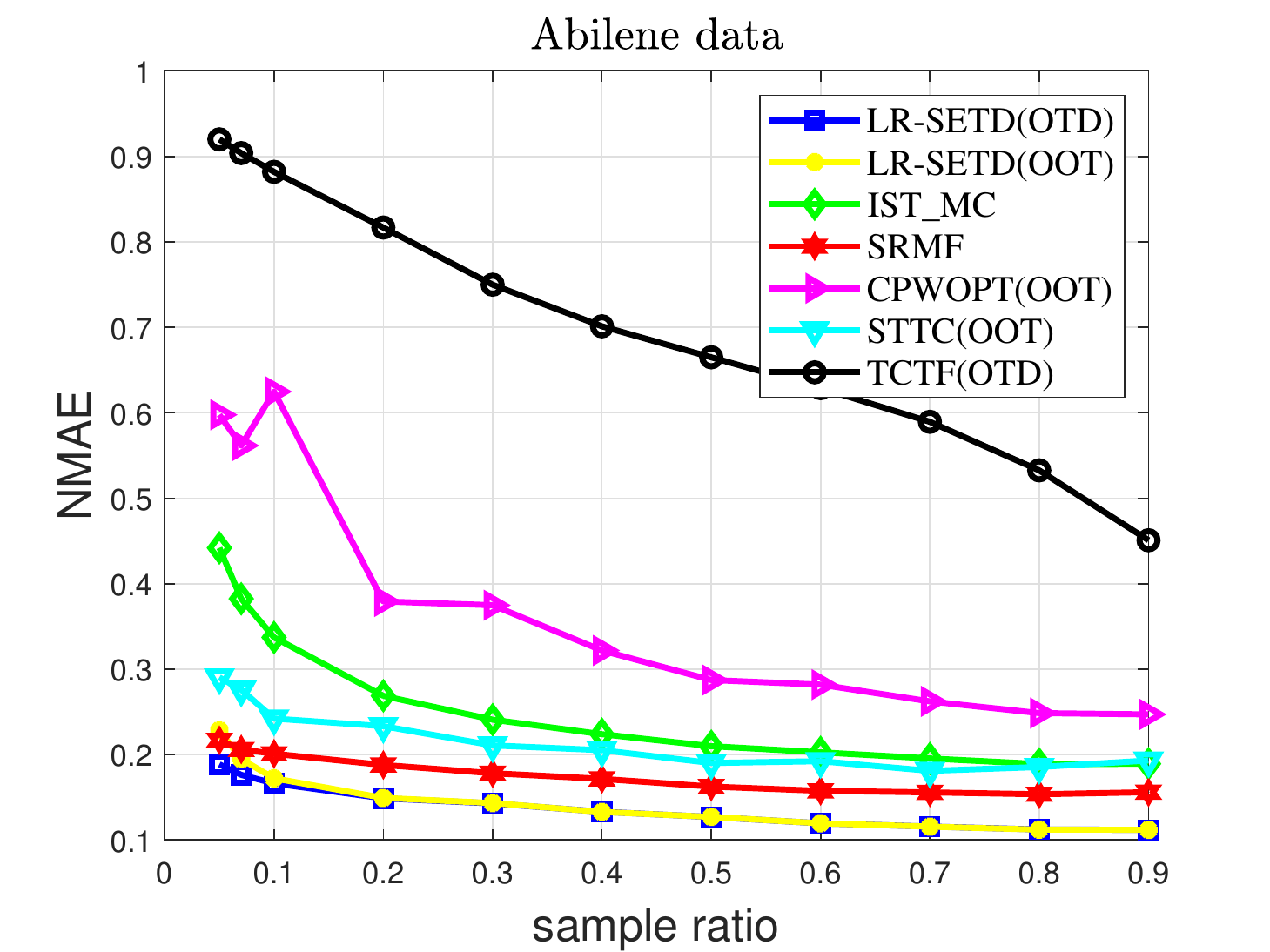}
		\includegraphics[width=0.24\textwidth]{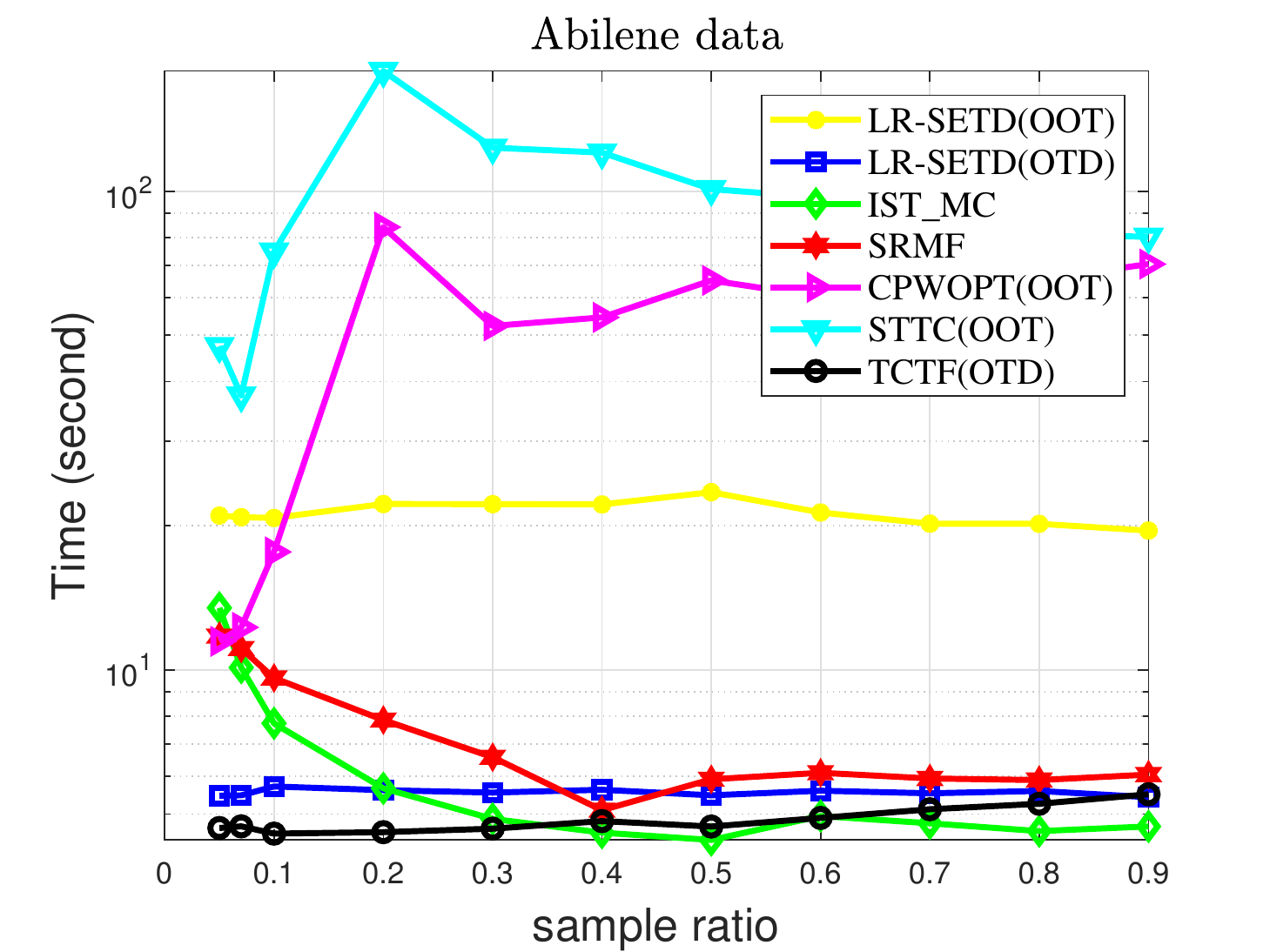}
	\includegraphics[width=0.24\textwidth]{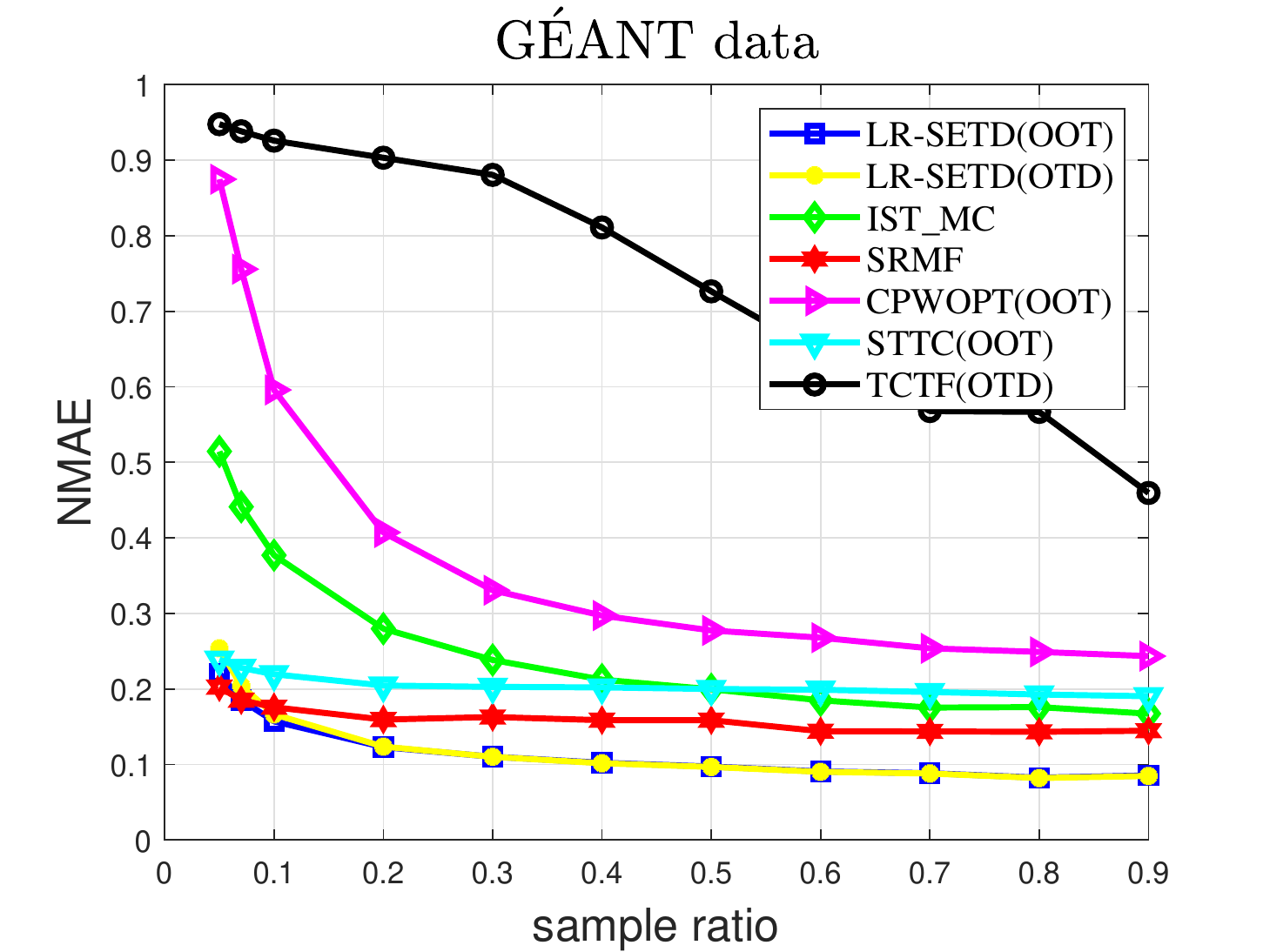}
		\includegraphics[width=0.24\textwidth]{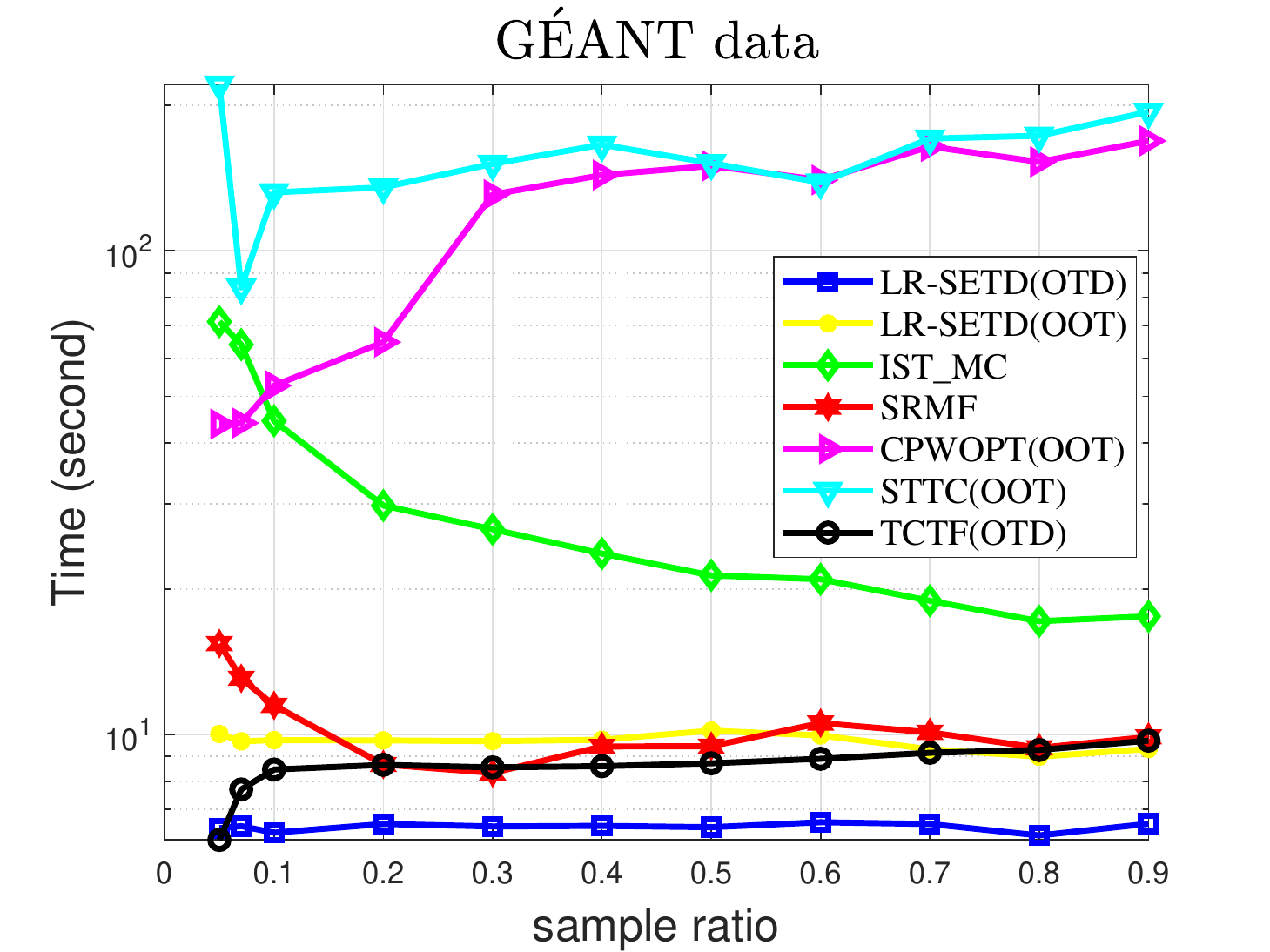}
	\caption{NMAE and computing time comparison under random missing for Abilene Data and G\'{E}ANT Data.}
	\label{AbeAndGeant}
\end{figure}

\begin{figure}[!htbp]
	\centering
	\includegraphics[width=.24\textwidth]{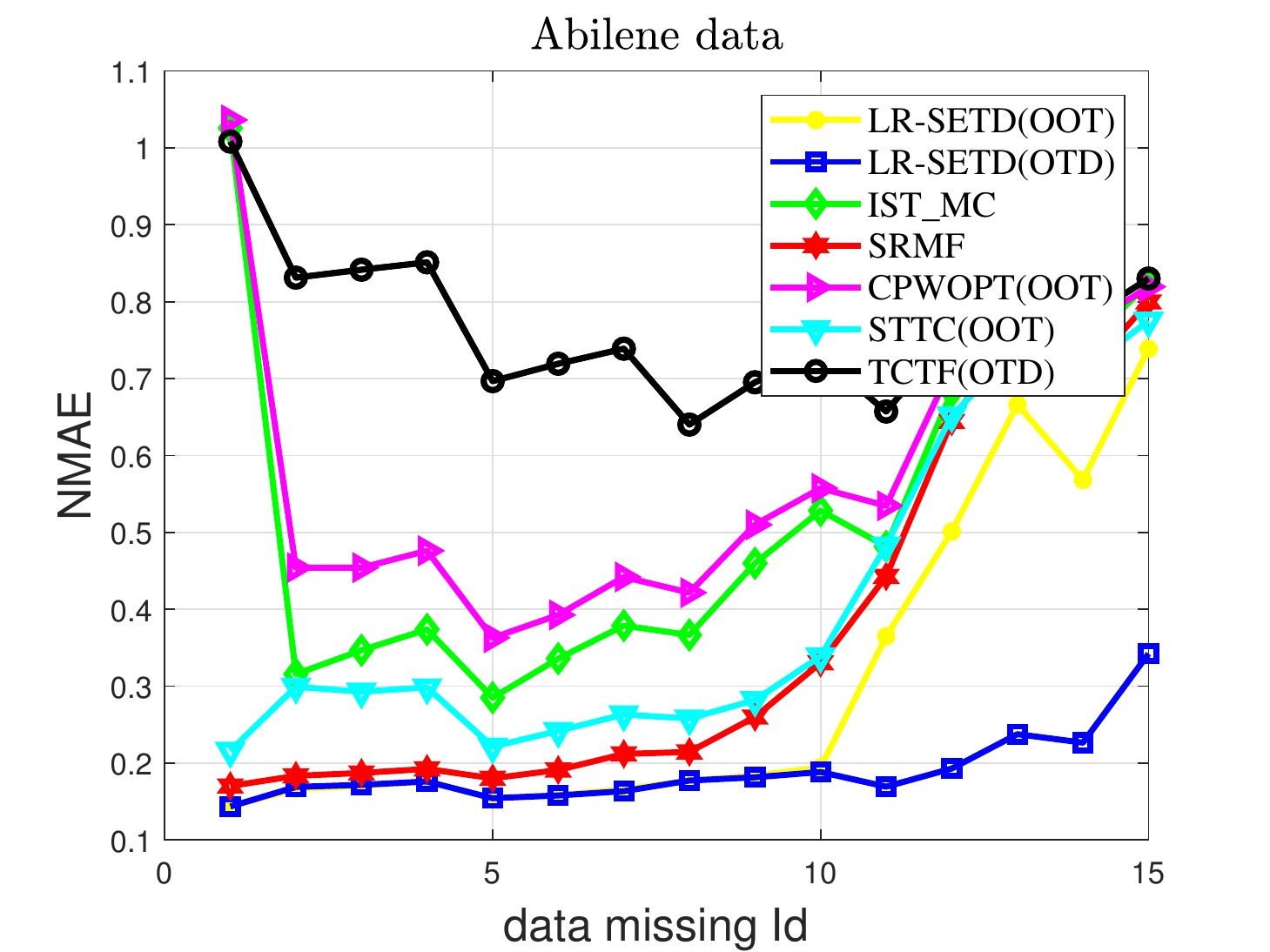}
	\centering
	\includegraphics[width=.24\textwidth]{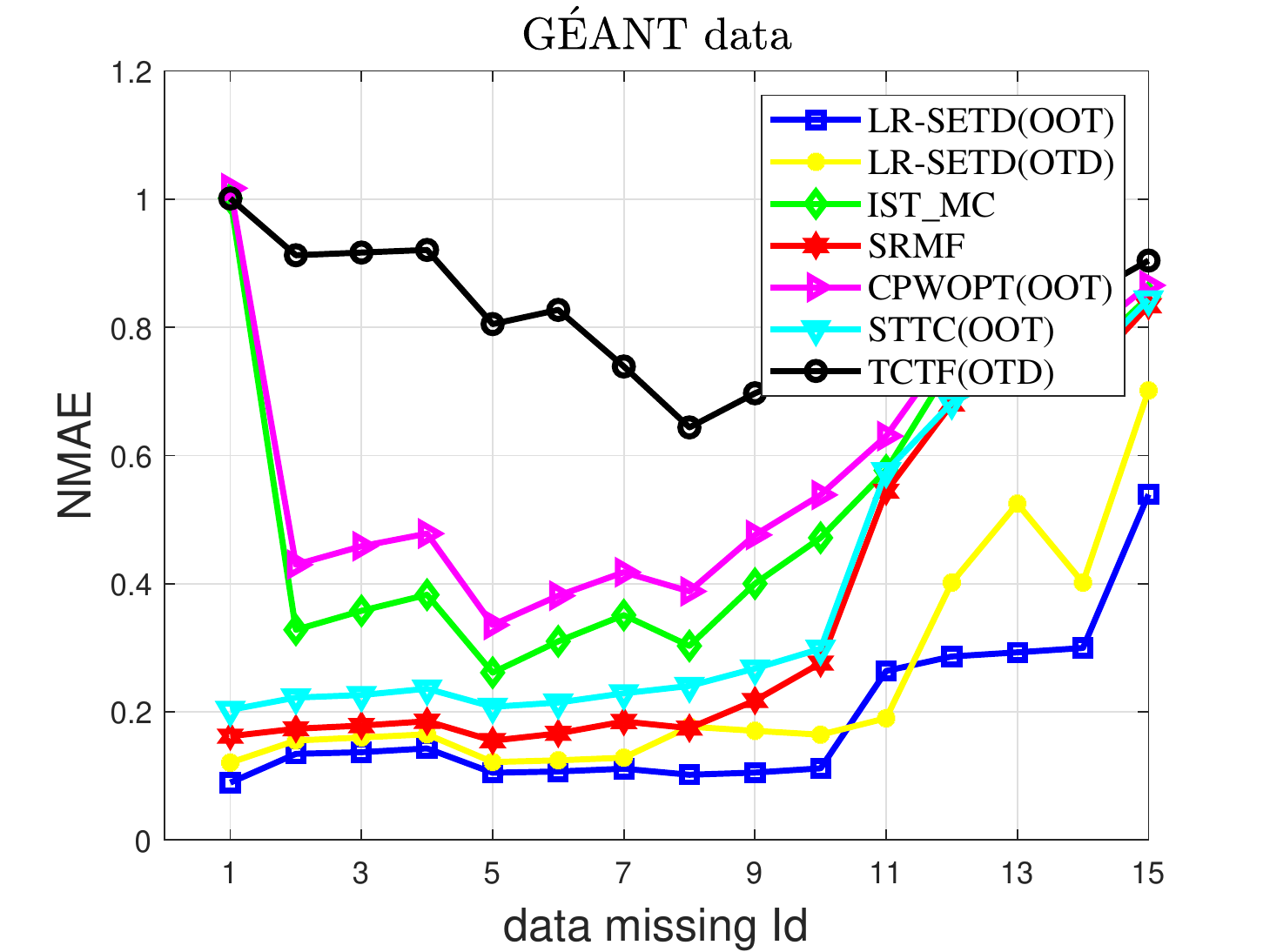}
	\caption{NMAE comparison under structurally missing for Abilene Data and G\'{E}ANT Data.}\label{Abeline_struc}
\end{figure}

\begin{figure}[!htbp]
	\includegraphics[width=.48\textwidth]{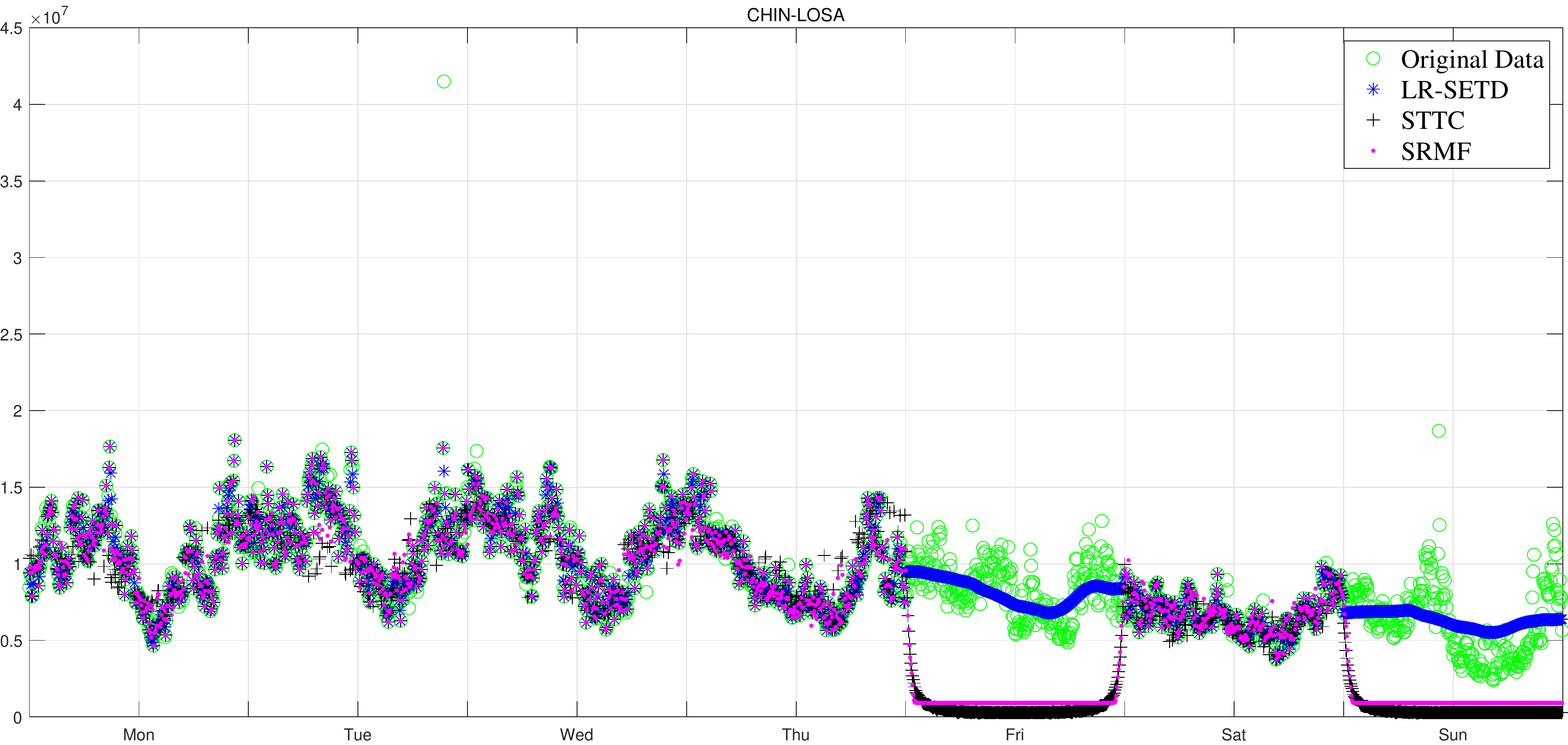}\\
	\includegraphics[width=.48\textwidth]{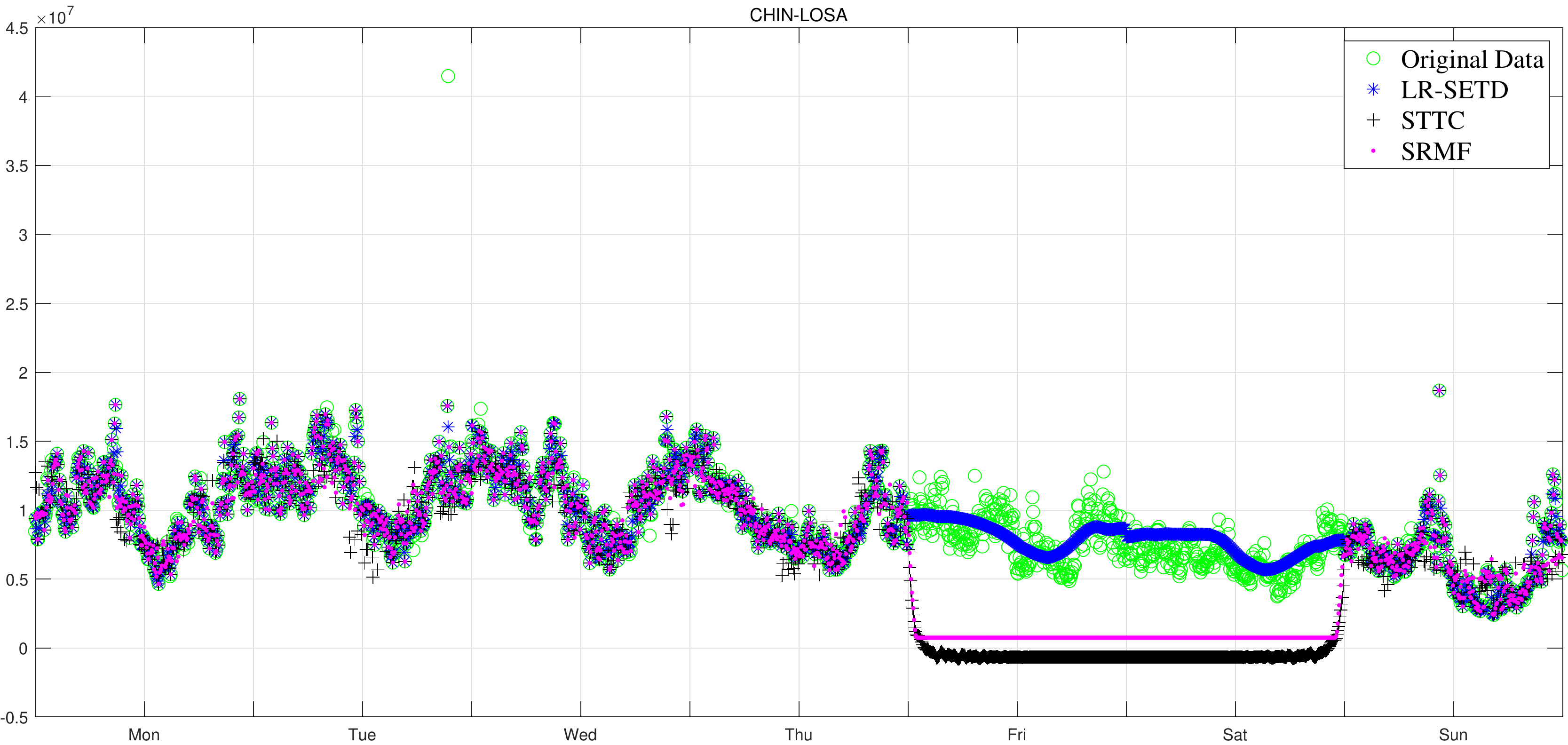}\\
	\caption{Visualization of the recovered data by LR-SETD, SRMF and STTC for the cases missing Id $\in\{12,14\}$.}
	\label{internet_struc2_visualization}
\end{figure}

Hereafter, we consider the randomly missing scenario, where the sample ratio is changed from $0.90$ to $0.05$. We plot NMAE values and computing time in seconds (denoted by Time (second)) in Fig. \ref{AbeAndGeant}. It can be easily seen from the left two plots in Fig. \ref{AbeAndGeant} that our approach LR-SETD has lower NMAE values in most cases, especially for the cases where sample ratios are greater than $0.10$. Moreover, from the right two plots in Fig. \ref{AbeAndGeant}, our approach is competitive to the others, since LR-SETD performs comparatively stable on both internet data sets. In some cases, the missing data may not be missed in a uniformly distributed way. So, we further consider the  structurally missing scenario. Here, we consider fifteen loss ways to degrade both Abilene and G\'{E}ANT internet traffic data and denote each case by missing Id  from $1$ to $15$. Specifically, we drop one slice data for every three slices with respect to the second mode (i.e., dropping $\mathcal{Z}(:,1,:)$, $\mathcal{Z}(:,4,:)$, $\mathcal{Z}(:,7,:)$, $\cdots$) for missing Id $= 1$. Then, we randomly remove $80\%$, $50\%$, $20\%$ data and, for missing Id $=\{2,5,8\}$ remove data from 11:00 to 12:00 every day for a week; for missing Id $=\{3,6,9\}$, remove data from 10:00 to 12:00 every day for a week; for missing Id $=\{4,7,10\}$, remove data from 10:00 to 13:00 every day for a week. Finally, we randomly generate 20\% loss and drop the whole (Friday), (Friday and Sunday), (Wednesday, Friday and Sunday), (Friday and Saturday) and (Friday, Saturday and Sunday) for missing Id $=\{11,\cdots,15\}$, respectively. We only show by Fig. \ref{Abeline_struc} the comparison on NMAE values. These curves in Fig. \ref{Abeline_struc} efficiently show that our approach LR-SETD outperforms the other completion methods in terms of achieving lower NMAE values, which means that LR-SETD has better performance on missing slices recovery cases than the others. We can also see the superiority of our LR-SETD from Fig. \ref{internet_struc2_visualization}. In our experiments, we observed that the type of tensors would affect the performance of tensorization methods. For example, both CPWOPT and STTC perform worse on OTD tensors than on OOT tensors, and TCTF works better on OTD tensors than on OOT tensors. Therefore, the results demonstrate that our approach LR-SETD works more ideally and reliably than the others.

\begin{figure}[!htbp]
	\centering	
	\includegraphics[width=0.24\textwidth]{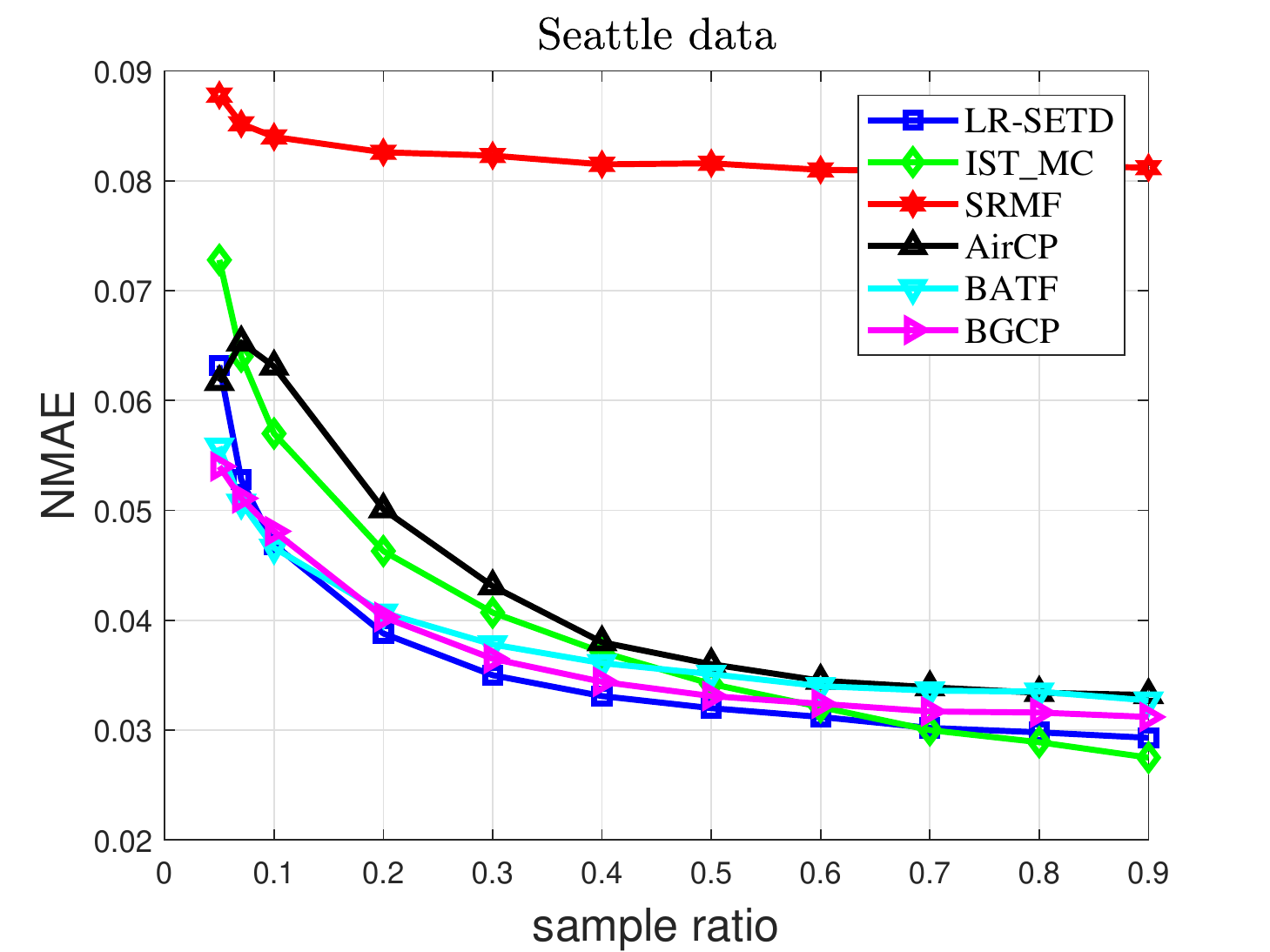}
		\includegraphics[width=0.24\textwidth]{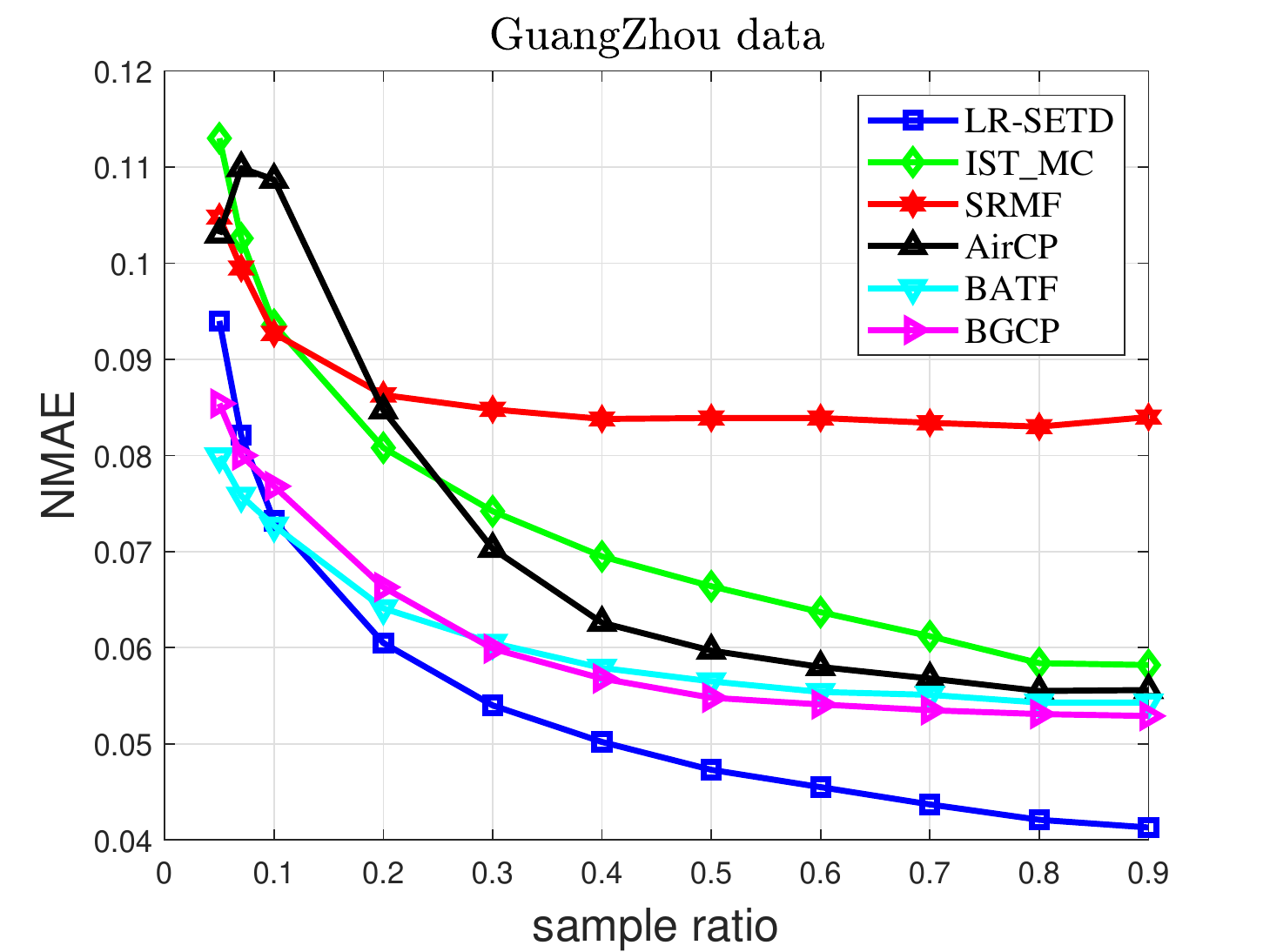}
		\caption{NMAE values with respect to sample ratio for both Seattle and Guangzhou data sets.}\label{GZ-N}
\end{figure}

\begin{figure}[!htbp]
	\includegraphics[width=.48\textwidth]{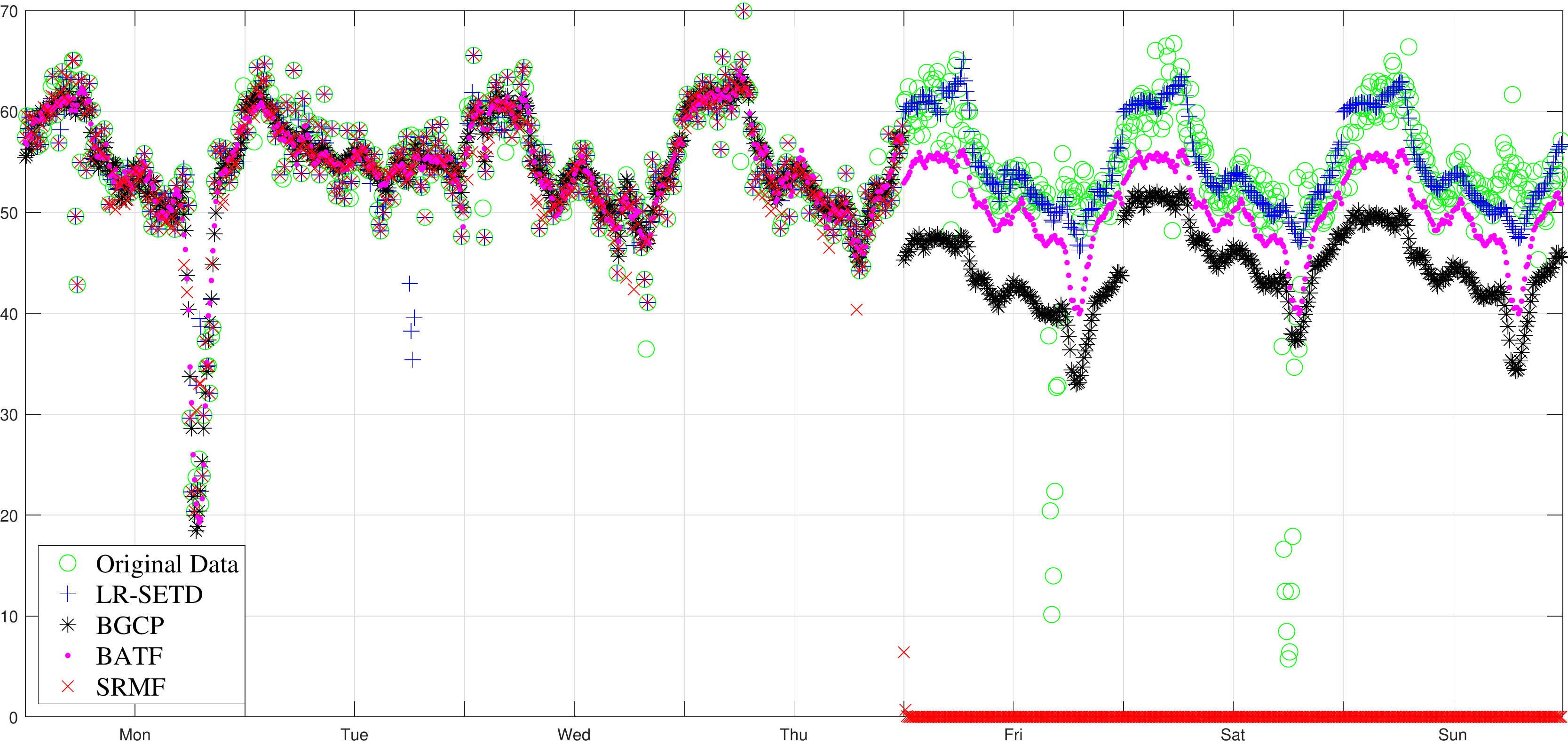}
	\includegraphics[width=.48\textwidth]{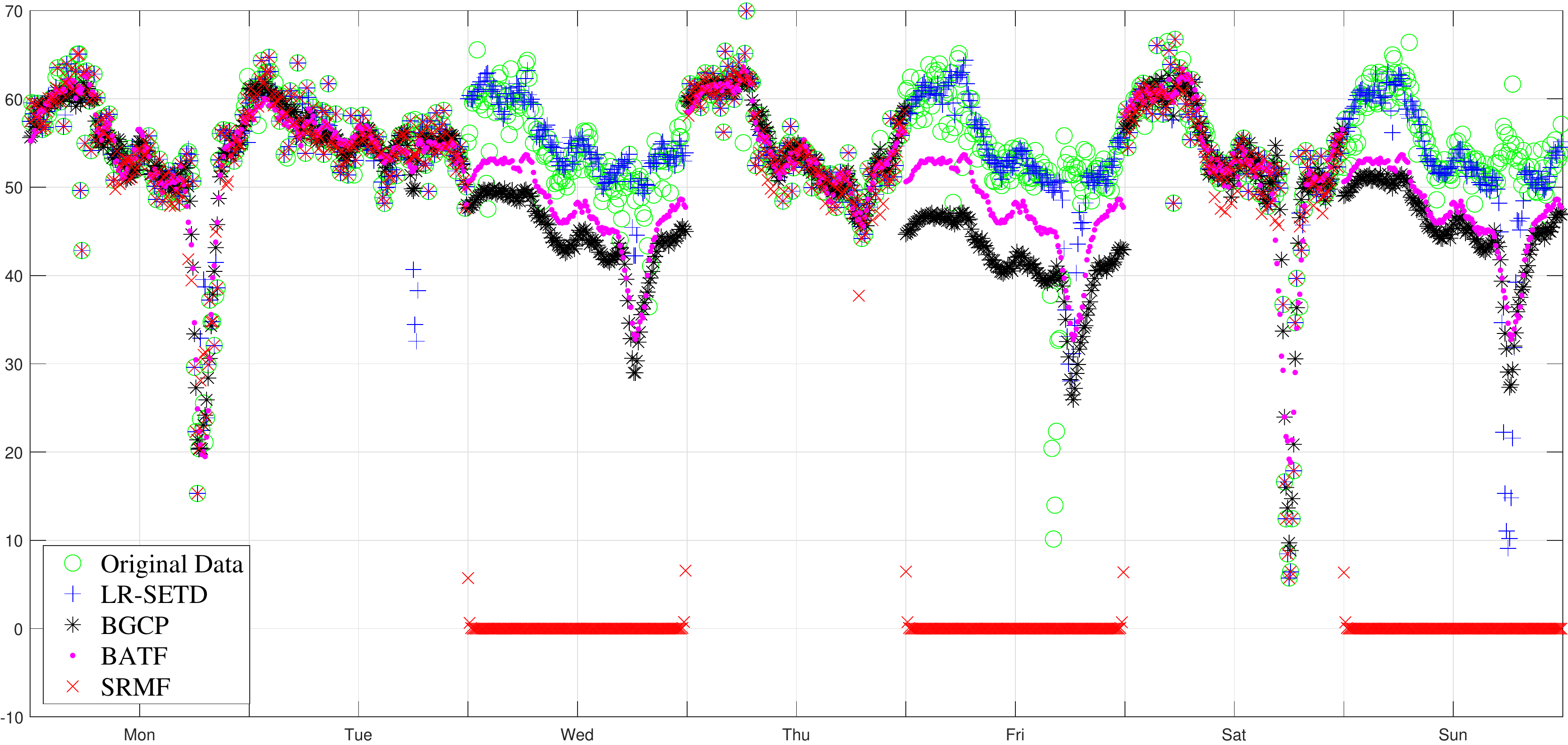}
	\caption{Recovered data by LR-SETD, BGCP, BATF, SRMF for the cases where Guangzhou data has missing slices data (which correspond to continuous or intermittent three days' data) and $80\%$ sample ratio for the observed part. The top plot corresponds to the case where all data in the last three days is not observed. The bottom one means that the data in Wednesday, Friday, and Sunday is missed. }\label{GZ-Rec}
\end{figure}

As another important issue, we further consider two real world public traffic flow data sets, i.e., Seattle freeway traffic flow\footnote{https://github.com/zhiyongc/Seattle-Loop-Data} and Guangzhou urban traffic speed data sets\footnote{https://doi.org/10.5281/zenodo.1205229}. For Seattle freeway traffic dataset, which is collected freeway traffic speed data from $323$ loop detectors with a $5$-minute resolution over the whole year of 2015 in Seattle, USA. In our experiments, we choose the data from January 1 to January 7 (i.e., one week) and organize the resulting data as a tensor with size ${323\times 144\times 7}$ and a matrix with size ${323\times 1008}$ for tensor- and matrix-based models, respectively. The Guangzhou urban traffic speed data is collected from $214$ road segments over two months (i.e., $61$ days from August 1 to September 30, 2016) with a $10$-minute resolution ($144$ time intervals per day) in Guangzhou, China. Correspondingly, we only consider the first seven days' data and organize the raw data set into a tensor of size $214\times 144\times 7$ and a matrix of size $214\times 1008$ for tensor- and matrix-based models, respectively. Noting that traffic data is different with the aforementioned internet data, we accordingly compare our approach with both SRMF and IST\_MC and the other three state-of-the-art algorithms designed for traffic data recovery. Specifically, we choose AirCP\footnote{https://github.com/hanchengge/AirCP} \cite{GCZS16}, BATF\footnote{https://github.com/sysuits/BATF} \cite{CHCLW19} and BGCP\footnote{https://github.com/lijunsun/bgcp\_imputation} \cite{CHS19}, which are tensorization methods. As tested for Abilene data, we consider the randomly missing scenario, where the sample ratio is changed from $0.90$ to $0.05$. Then we also plot the NMAE values with respect to sample ratios in Fig. \ref{GZ-N} and show the recovered data in Fig. \ref{GZ-Rec} for the case where the Guangzhou data has missing slices data (which correspond to three days' data) and $80\%$ sample ratio for the observed part. It can be seen from Figs. \ref{GZ-N} and \ref{GZ-Rec} that our approach is still competitive in traffic data recovery problems, especially for the cases where the sample ratio is greater than $10\%$.

\subsection{Image Inpainting}
In this subsection, we apply our approach LR-SETD to image restoration. As we know, color images are natural third order tensors. Therefore, we will first focus on color image inpainting and compare LR-SETD with seven efficient benchmark solvers introduced in the literature,  including the high accuracy low rank tensor completion model \cite{LMWY13} (HaLRTC) , TCTF \cite{ZLLZ18}, tensor nuclear norm regularized model \cite{LFLY18} (TNN\footnote{https://github.com/canyilu/tensor-completion-tensor-recovery}), Bayesian CP factorization model \cite{ZZC15} (BCPF\footnote{https://github.com/qbzhao/BCPF}), the low Tucker rank tensor recovery modle by iterative $p$-shrinkage thresholding algorithm \cite{SLH19} (IpST), truncated nuclear norm regularized model \cite{HZYL13} (TNNR(apgl)\footnote{https://github.com/xueshengke/TNNR}) and the low-$n$-rank tensor recovery  model by ADM \cite{GRY11} (ADM-TR).

Throughout this section, we still employ \eqref{StopCrit} as the termination criterion for all algorithms, where ${\rm Tol}=10^{-5}$ and the maximum iteration is $250$. Moreover, all model parameters of \eqref{optim-ReZ} keeps the same settings as used in internet traffic data recovery, i.e., $\sigma = 1$, $\lambda= 10^{-2}$, $\alpha_i=1/3$, $i=1,2,3$, and $(\omega_1,\omega_2,\omega_3) = (1,1,0)$. For the other compared algorithms, we follow the default setting used in their codes and papers.  The quality of the restored image is measured by the well known Peak Signal-to-Noise Ratio (PSNR) and the relative squared error (RSE), which are defined by
$$ {\rm PSNR} = 10\cdot{\rm log}_{10}\ \frac{(\mathcal{Z}_{\max})^2}{\|\mathcal{Z}-\mathcal{Z}_{\rm true}\|_F^2/|\Omega^C|} $$
and
$${\rm RSE}= \frac{\|\mathcal{Z}-\mathcal{Z}_{\rm true}\|_F}{\|\mathcal{Z}_{\rm true}\|_F},$$
where $\mathcal{Z}_{\rm true}$, ${\mathcal{Z}}_{\max}$, $\mathcal{Z}$ are the original tensor, the maximum pixel value of  $\mathcal{Z}_{\rm true}$, and the recovered tensor, respectively; Additionally, $|\Omega^C|$ represents the number of elements of the complement set $\Omega^C$ of $\Omega$.

\begin{figure}[!htbp]
	\centering
		\includegraphics[width=0.09\textwidth]{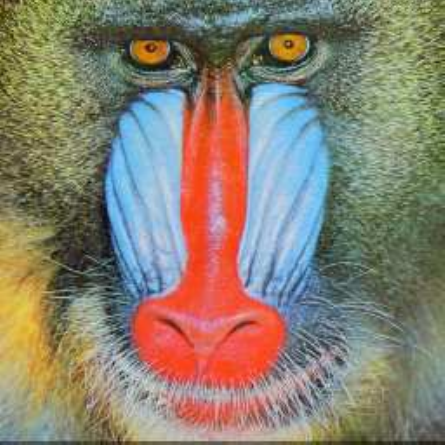}
		\includegraphics[width=0.09\textwidth]{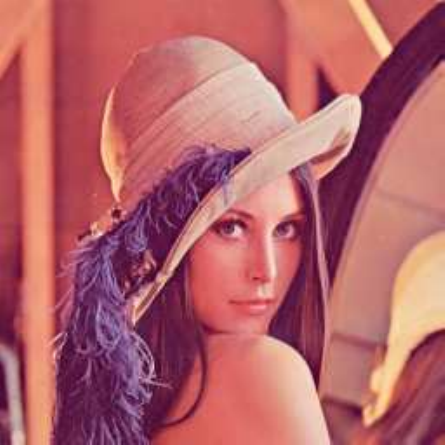}
		\includegraphics[width=0.09\textwidth]{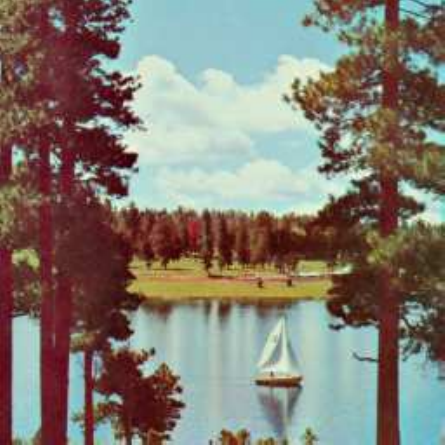}
		\includegraphics[width=0.09\textwidth]{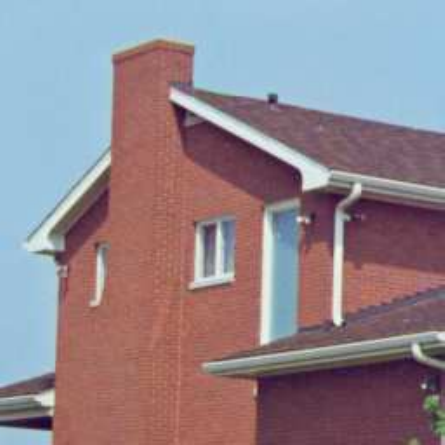}
		\includegraphics[width=0.09\textwidth]{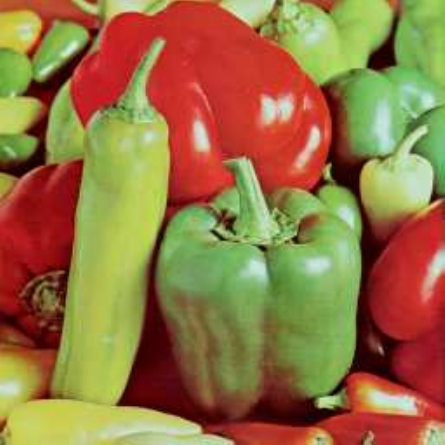}
	\caption{Five color images for experiments. From left to right: baboon, lena, sailboat, house, peppers.}\label{ColorTestImage}
\end{figure}

\begin{figure}[!htbp]
	\includegraphics[width=.49\textwidth]{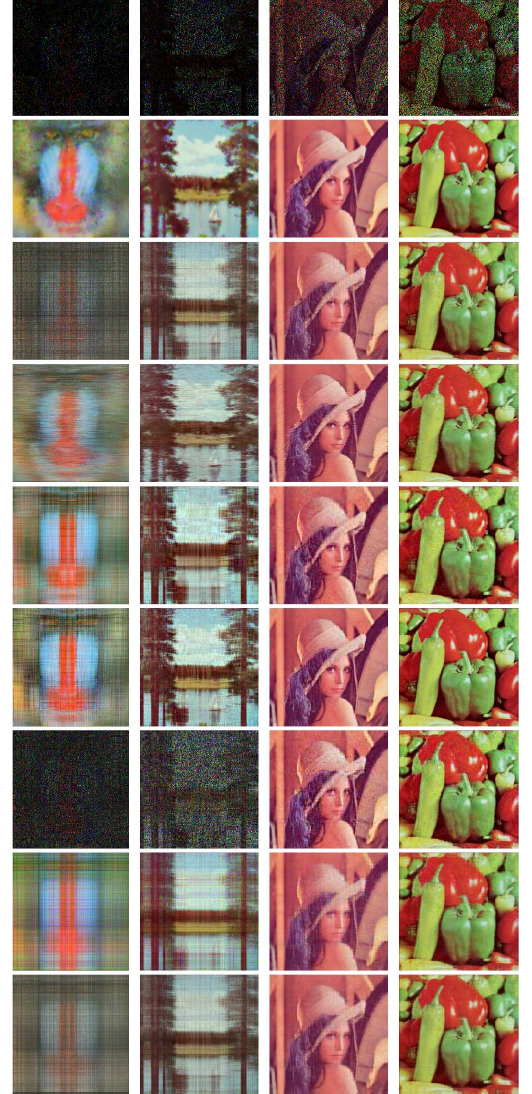}
	\centering
	\caption{Degraded images and recovered images by the algorithms. The baboon, sailboat, lena, and peppers have $5\%$, $10\%$, $30\%$ and $40\%$ observed information, respectively. From the second row to bottom: LR-SETD, HaLRTC, TNN, BCPF, IpST, TCTF, TNNR(apgl) and ADM(TR).}\label{sample_ration_002}
\end{figure}

\begin{figure}[!htbp]
	\includegraphics[width=.24\textwidth]{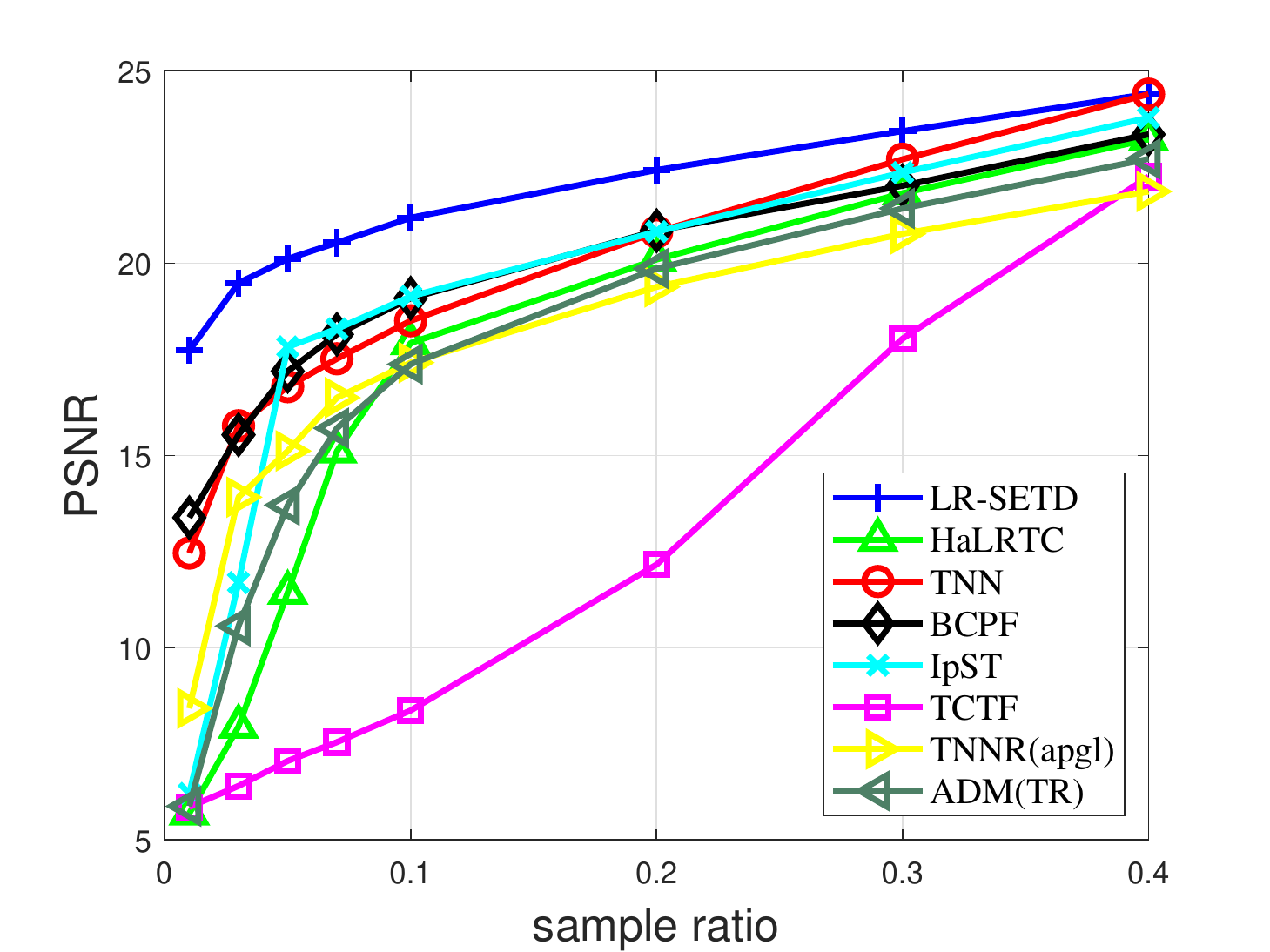}
	\includegraphics[width=.24\textwidth]{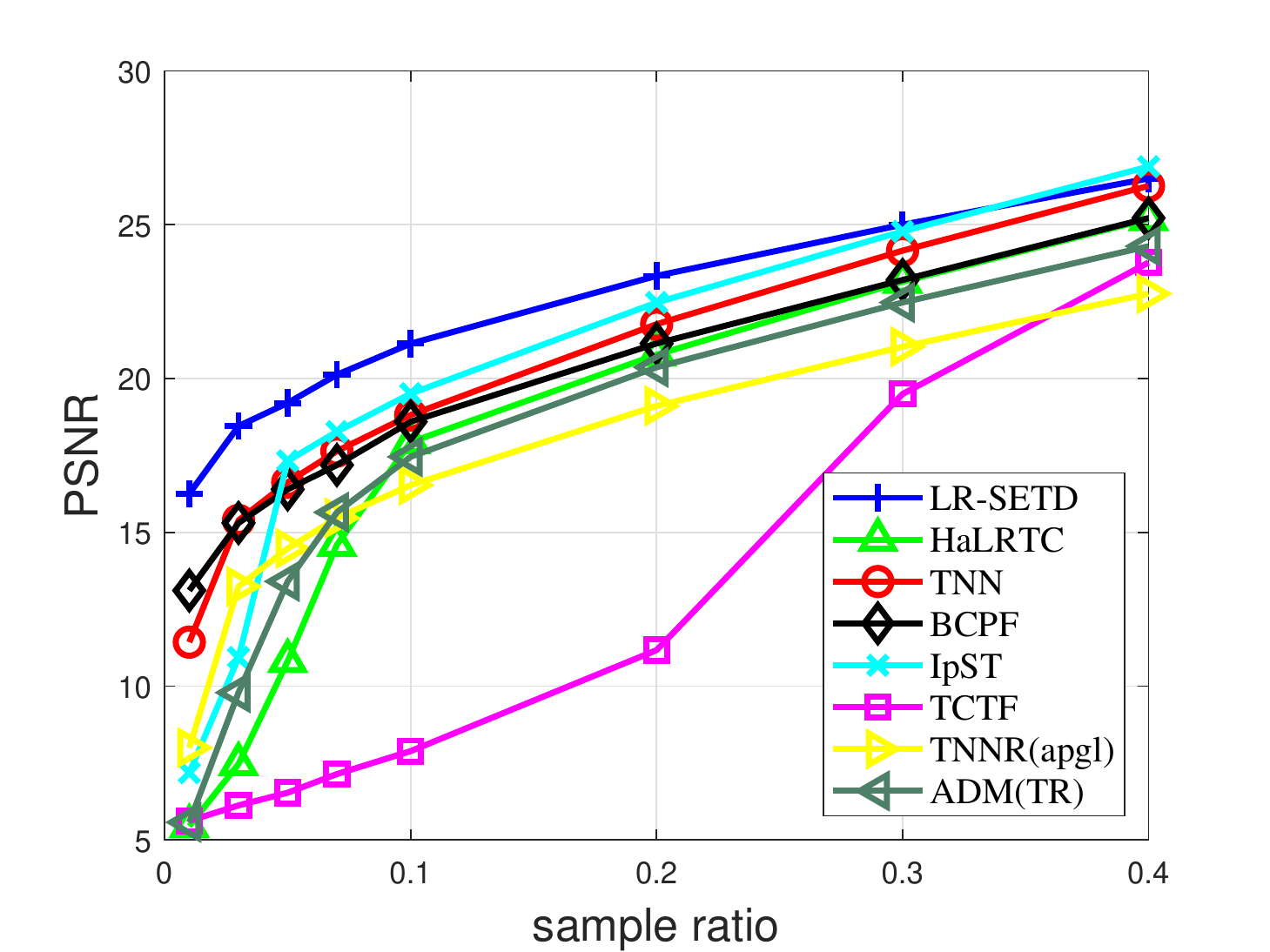}\\
	\includegraphics[width=.24\textwidth]{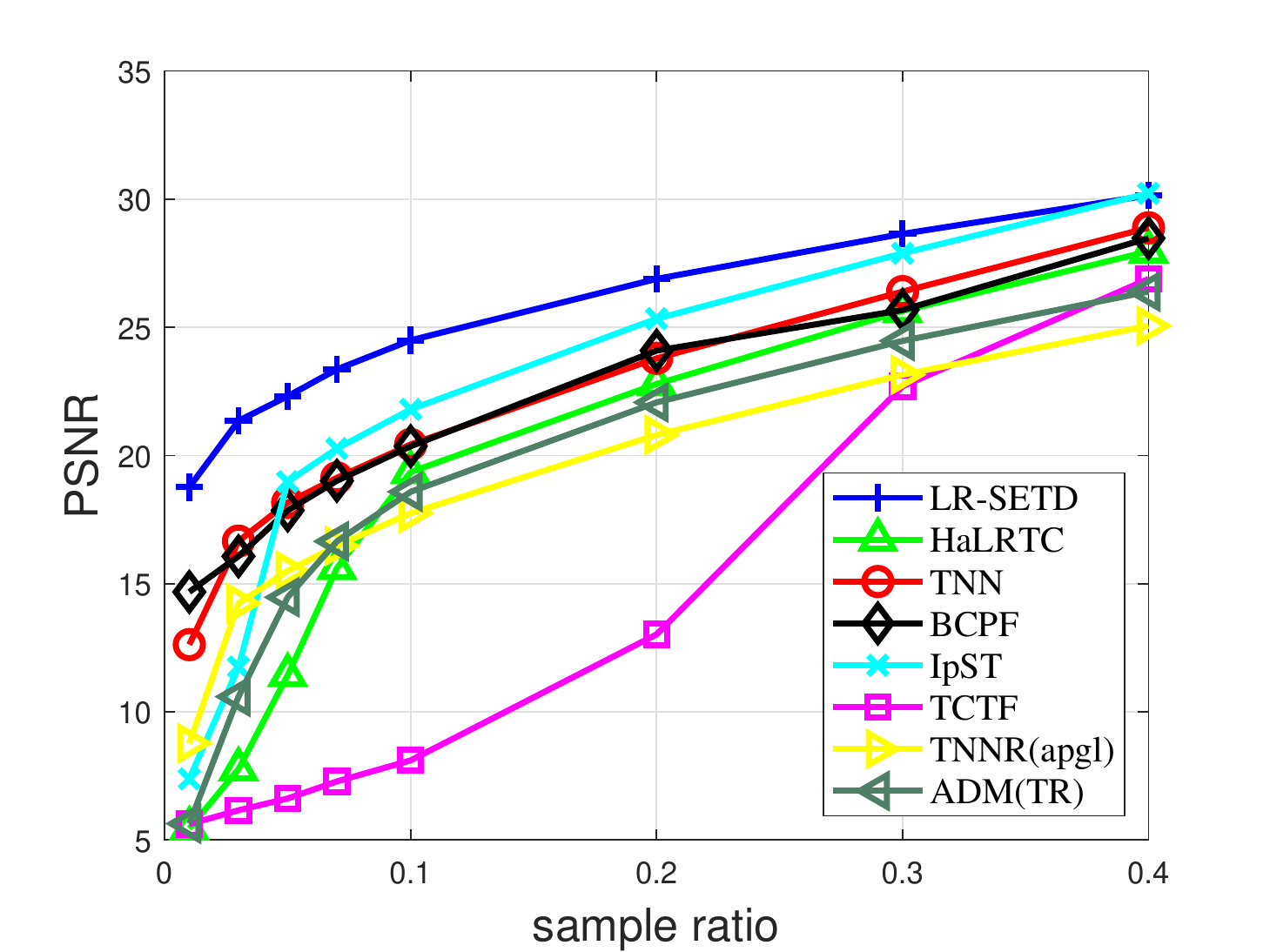}
	\includegraphics[width=.24\textwidth]{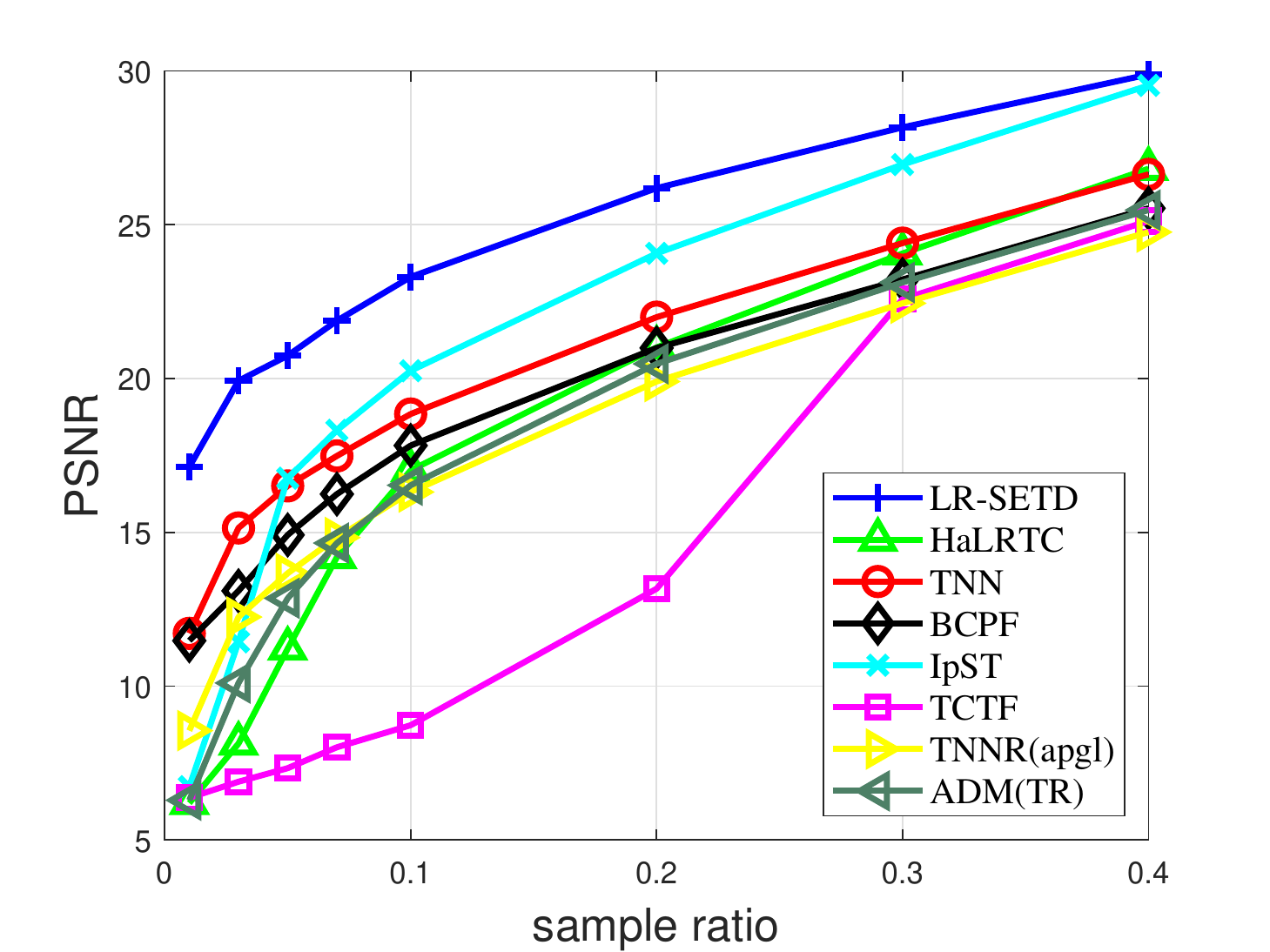}
	\caption{Comparison results of PSNR values with respect to sample ratios. From top to bottom, left to right: baboon, sailboat, lena and peppers.}
	\label{imageforRandMissing}
\end{figure}

We now consider five popular color images that are widely used in the image literature. All images are of size $256\times256\times3$ and are collected in Fig. \ref{ColorTestImage}. As the data loss ways used in internet traffic data recovery, we also consider two scenarios to degrade the original images. The first one is some data being dropped in a uniformly distributed way. In Fig. \ref{sample_ration_002}, we investigate four cases, i.e.,  the baboon, sailboat, lena, and peppers have $5\%$, $10\%$, $30\%$ and $40\%$ observed information, respectively. It can be easily seen from Fig. \ref{sample_ration_002} that our approach LR-SETD can recover better images than the other seven methods, especially for the baboon and sailboat images. To further show the superiority of our LR-SETD, we investigate another eight sample ratios, i.e., $\{0.01$, $0.03$, $0.05$, $0.07$, $0.10$, $0.20$, $0.30$, $0.40\}$, for the four images tested in Fig. \ref{sample_ration_002}. We only show the comparison by PSNR values with respect to sample ratios in Fig. \ref{imageforRandMissing}. The results in Fig. \ref{imageforRandMissing} show that our LR-SETD outperforms the other algorithms in terms of achieving higher PSNR values.

\begin{figure}[!htbp]
	\includegraphics[width=.48\textwidth]{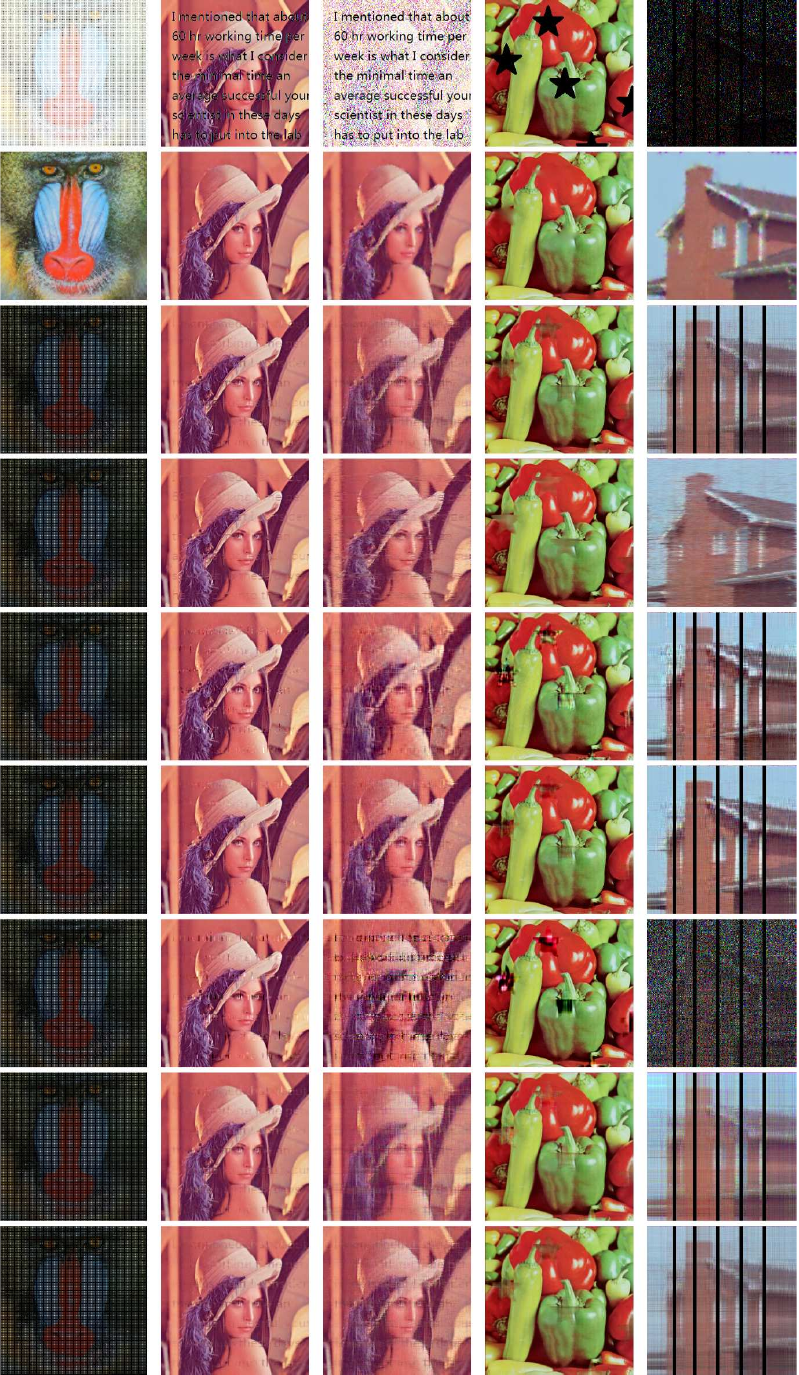}
	\centering
	\caption{Degraded images and recovered images by the algorithms. The baboon is corrupted by dropping one slice for every two slices with respect to mode one; the second column, i.e., image lena, loses its pixels with respect to these English sentences; the second lena image, i.e., the third column, is first corrupted by dropping $70\%$ pixels in a random way and then added a text mask; the peppers image is degraded by dropping pixels with star shape; the house image is corrupted by dropping $90\%$ information with slices missed data. From the second row to bottom: LR-SETD, HaLRTC, TNN, BCPF, IpST, TCTF, TNNR(apgl) and ADM(TR).}\label{sample_ration_003}
\end{figure}

Below, we consider structurally missing cases, that is, the observed images have missing slices data and/or shape missing data as shown by the first row in Fig. \ref{sample_ration_003}. The recovered images listed in Fig. \ref{sample_ration_003} demonstrate that our approach LR-SETD performs much better than the other seven methods, which also support that LR-SETD works more stable on different types of missing data. To see the details of these approaches on image inpainting, we report the RSE values, PSNR values, iterations and computing time in seconds (denoted by Iter (Time)) in Table \ref{tab1}, where the results clearly show that LR-SETD can recover better images in terms of achieving higher PSNR values and lower RSE values than the other solvers.

\begin{sidewaystable}[!htbp]
	\caption{Numerical results of images with structurally missing pixels as tested in Fig. \ref{sample_ration_003}.}\label{tab1}
	\resizebox{\textwidth}{25mm}{
		\begin{tabular}{lllllllllllllllllllll}\toprule
			\multicolumn{1}{l}{\multirow{2}{*}{}}&\multicolumn{1}{l}{\multirow{2}{*}{Method}} &\multicolumn{3}{c}{baboon} &&\multicolumn{3}{c}{lena-1}&&\multicolumn{3}{c}{lena-2}&&\multicolumn{3}{c}{peppers}&&\multicolumn{3}{c}{house}\\
			\cline{3-5} \cline{7-9}\cline{11-13}\cline{15-17}\cline{19-21}
			&& RSE & PSNR &  Iter (Time) && RSE & PSNR& Iter (Time) && RSE & PSNR & Iter (Time)&& RSE & PSNR & Iter (Time)&& RSE & PSNR & Iter (Time)\\\midrule		
			& LR-SETD  & \textbf{0.127} & \textbf{23.3} & {87 (4.6)} && \textbf{0.036} & \textbf{33.8} & {84 (4.1)} && \textbf{0.076} & \textbf{27.3} & {84 (4.5)}&& \textbf{0.059} & \textbf{30.5} & {81 (4.3)} && \textbf{0.111} &\textbf{23.7} & {91 (4.3)} \\
			& HaLRTC  & 0.863 & 6.6 & 41 (1.7) && 0.055 & 30.2 & 55 (2.5) && 0.117 & 23.7 & 61 (2.8)&& 0.086 & 27.3 & 60 (2.9) && 0.363 & 13.4 & 91 (4.3)\\
			& TNN  & 0.865& 6.6 & 66 (1.7) && 0.054  & 30.3 & 177 (3.5) && 0.111 & 24.1 & 176 (3.4)& & 0.084 & 27.5 & 143 (3.1) && 0.144 & 20.7 & 201 (3.9)\\
			& BCPF  & 0.866 & 6.6 & 20 (207.1) && 0.070 & 29.2 & 20 (24.7) && 0.116 & 24.3 & 20 (91.6)&& 0.129 & 23.5 & 50 (28.7) && 0.355 & 13.3 & 20 (71.4)\\
			& IpST  & 0.865 & 6.6 &  2 (0.2) && 0.049 & 31.3 & 35 (3.1) && 0.093 & 25.7 & 61 (5.1)& & 0.084 & 27.5 & 30 (2.6) && 0.348 & 13.7 & 75 (6.5)\\
			& TCTF  & 0.865 & 6.6 & 250 (2.1) && 0.084 & 26.6 & 250 (2.6) && 0.234 & 17.7 & 250 (2.8)& & 0.160 & 21.8 & 250 (2.7) && 0.755 & 7.1 & 250 (2.8) \\
			& TNNR(apgl)  & 0.866 & 6.6 & 87 (1.5) && 0.068 & 28.4 & 515 (9.2) && 0145 & 21.8 & 601 (10.9)& & 0.095 & 26.3 & 603 (10.7) && 0.371 & 13.2 & 602 (10.1) \\
			& ADM(TR) & 0.866 & 6.6 & 8 (28.8) && 0.066 & 28.7 & 116 (84.7) && 0.119 & 23.5 & 176 (124.6)&& 0.092 & 26.6 & 242 (169.9) && 0.370 & 13.2 & 221 (250.2)\\ \bottomrule		
	\end{tabular}}
\end{sidewaystable}

%\begin{figure}[!htbp]
%	\includegraphics[width=.49\textwidth]{reg_para_1.pdf}
%	\includegraphics[width=.49\textwidth]{reg_para_3.pdf}
%	\caption{RSE and Iter times under different $\lambda$ in sr = 10\%, 30\% from top to bottom.}\label{para_sensi}
%\end{figure}

\begin{figure}[!htbp]
	\includegraphics[width=.48\textwidth]{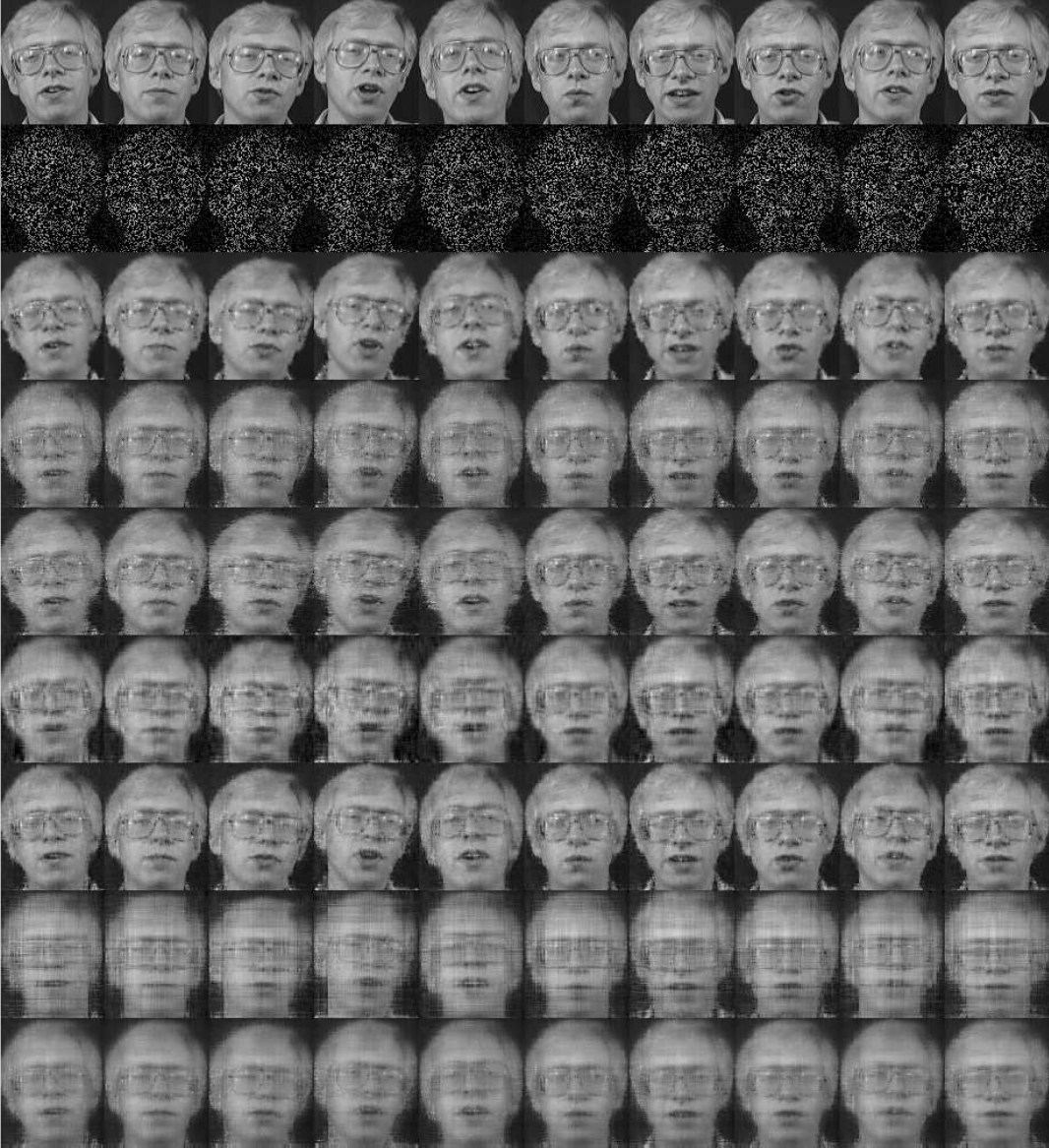}
	\caption{Comparison for the recovered results by the algorithms. From top to bottom: the original face data, observed face data with $70\%$ missing information, LR-SETD, HaLRTC, TNN, BCPF, IpSTA, TNNR(apgl) and ADM(TR).}
	\label{Xss2_visualization}
\end{figure}

\begin{figure}[!htbp]
	\includegraphics[width=0.24\textwidth]{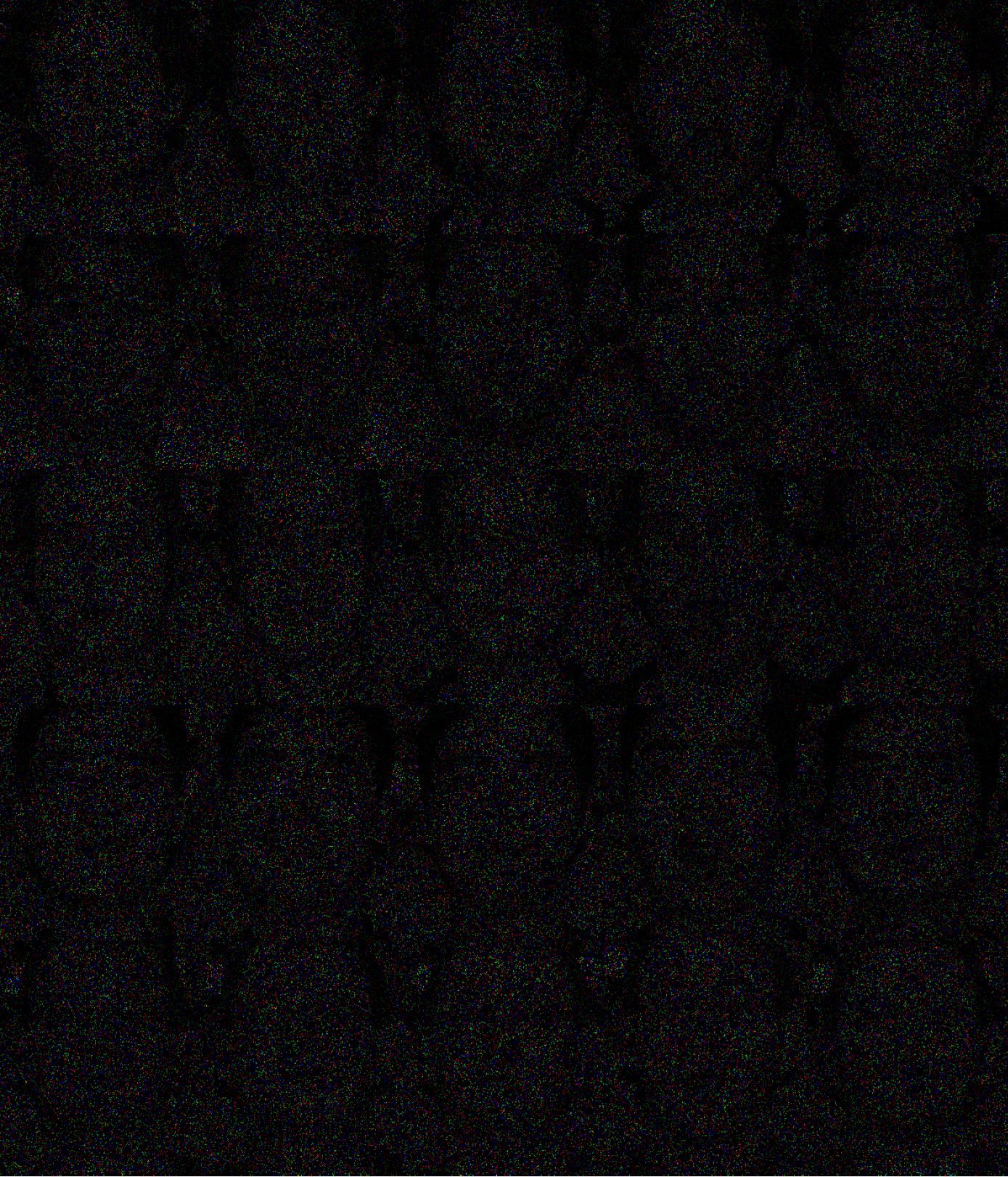}
	\includegraphics[width=0.24\textwidth]{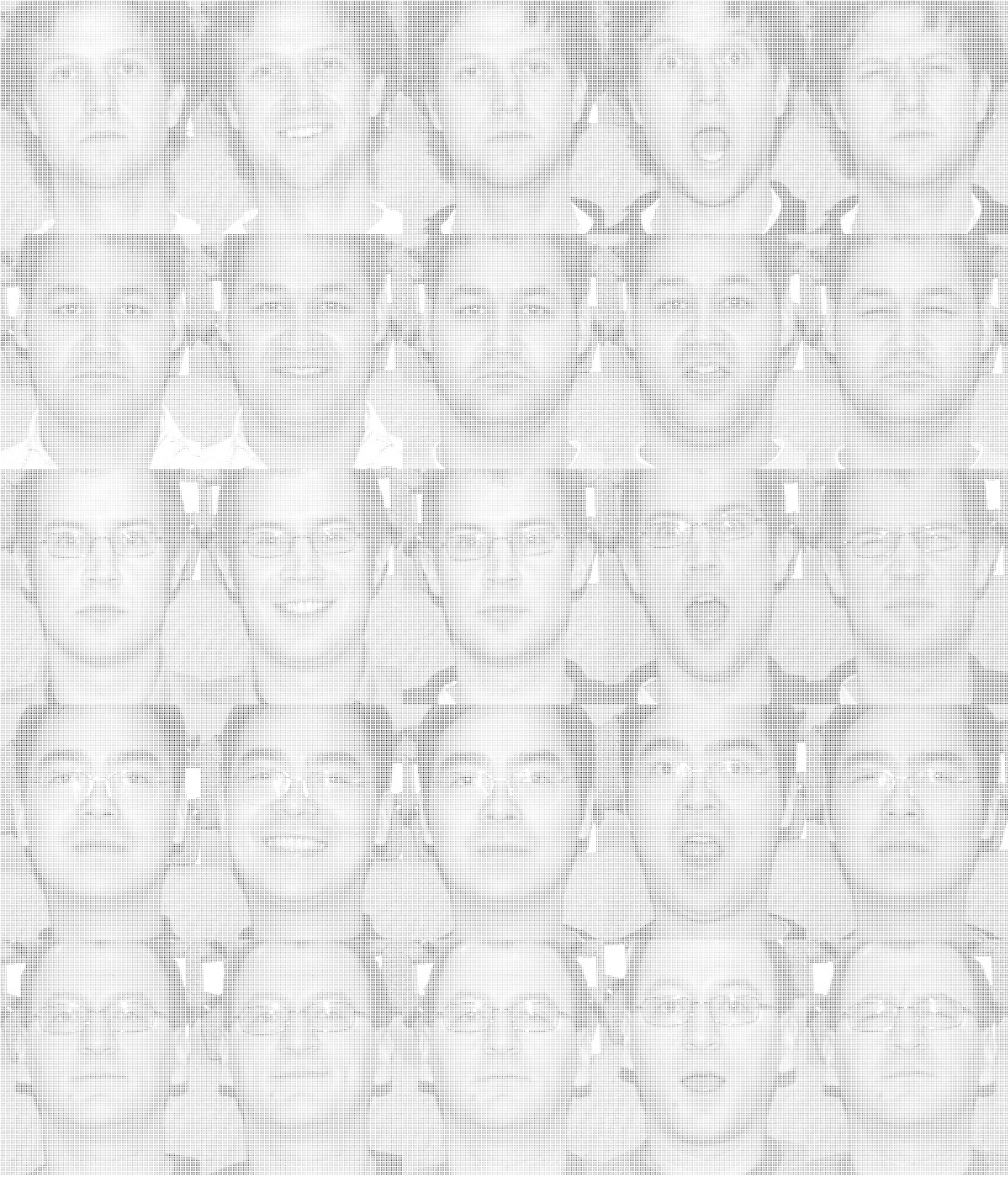}\\
	\includegraphics[width=0.24\textwidth]{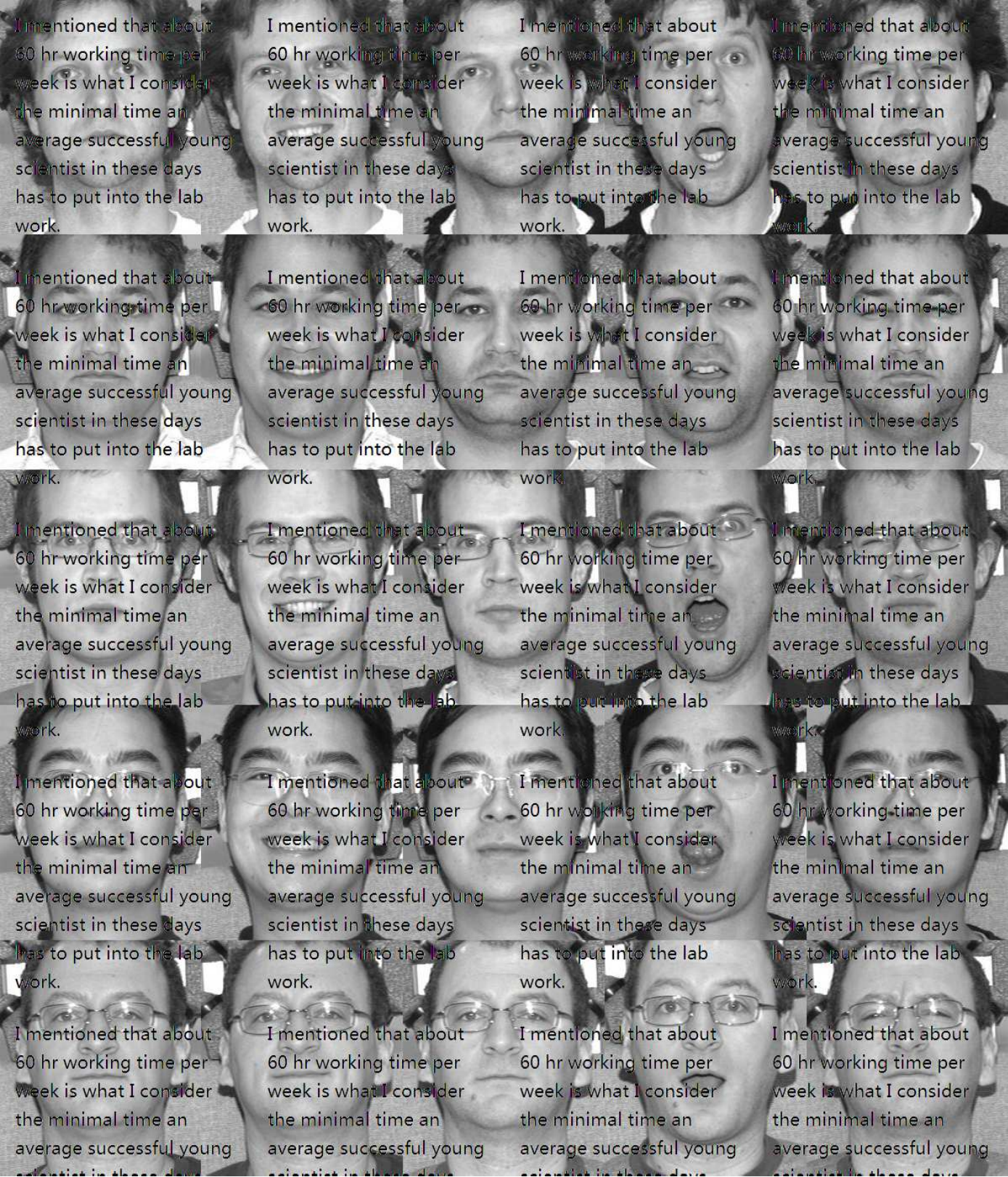}
	\includegraphics[width=0.24\textwidth]{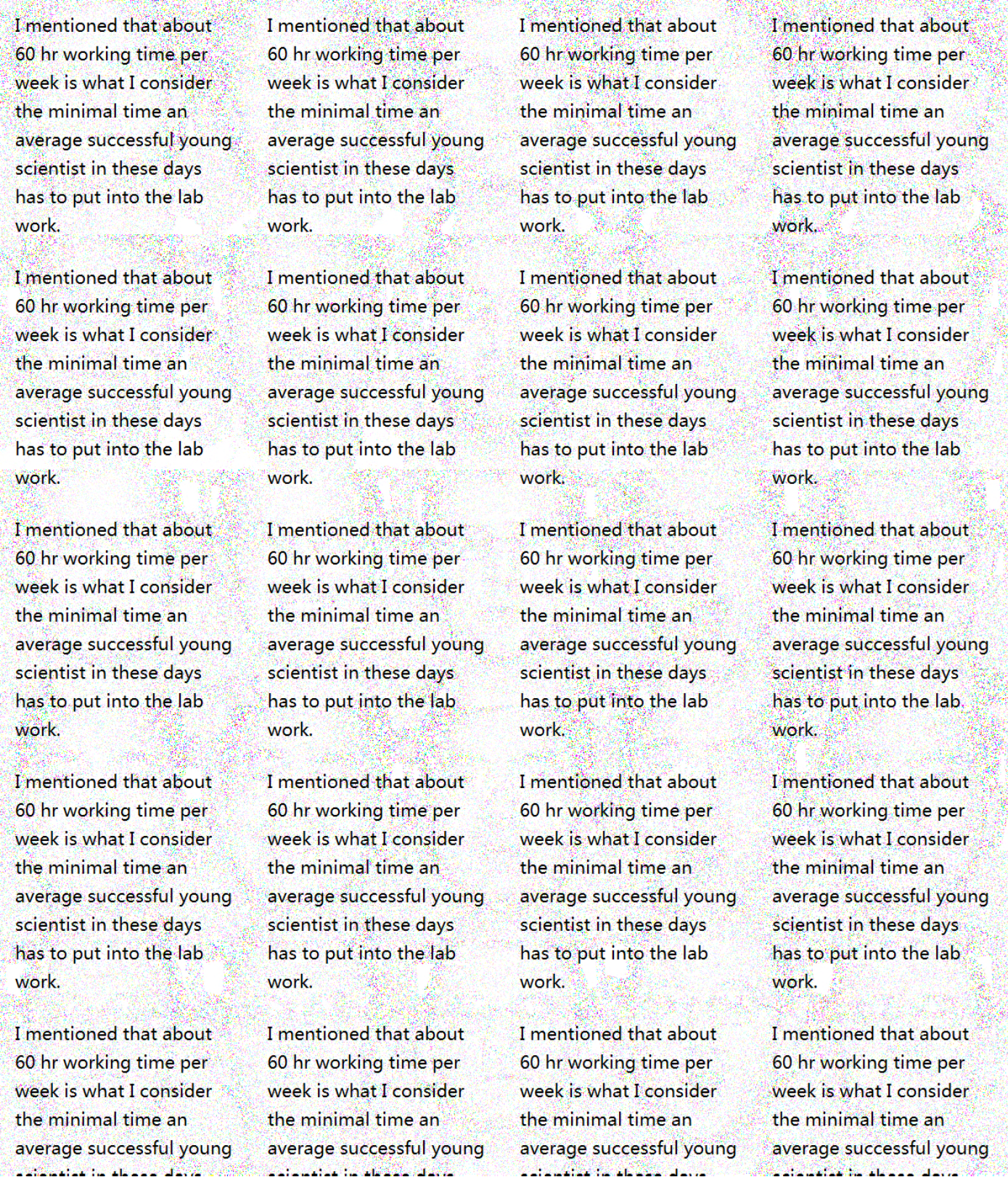}\\
	\includegraphics[width=0.24\textwidth]{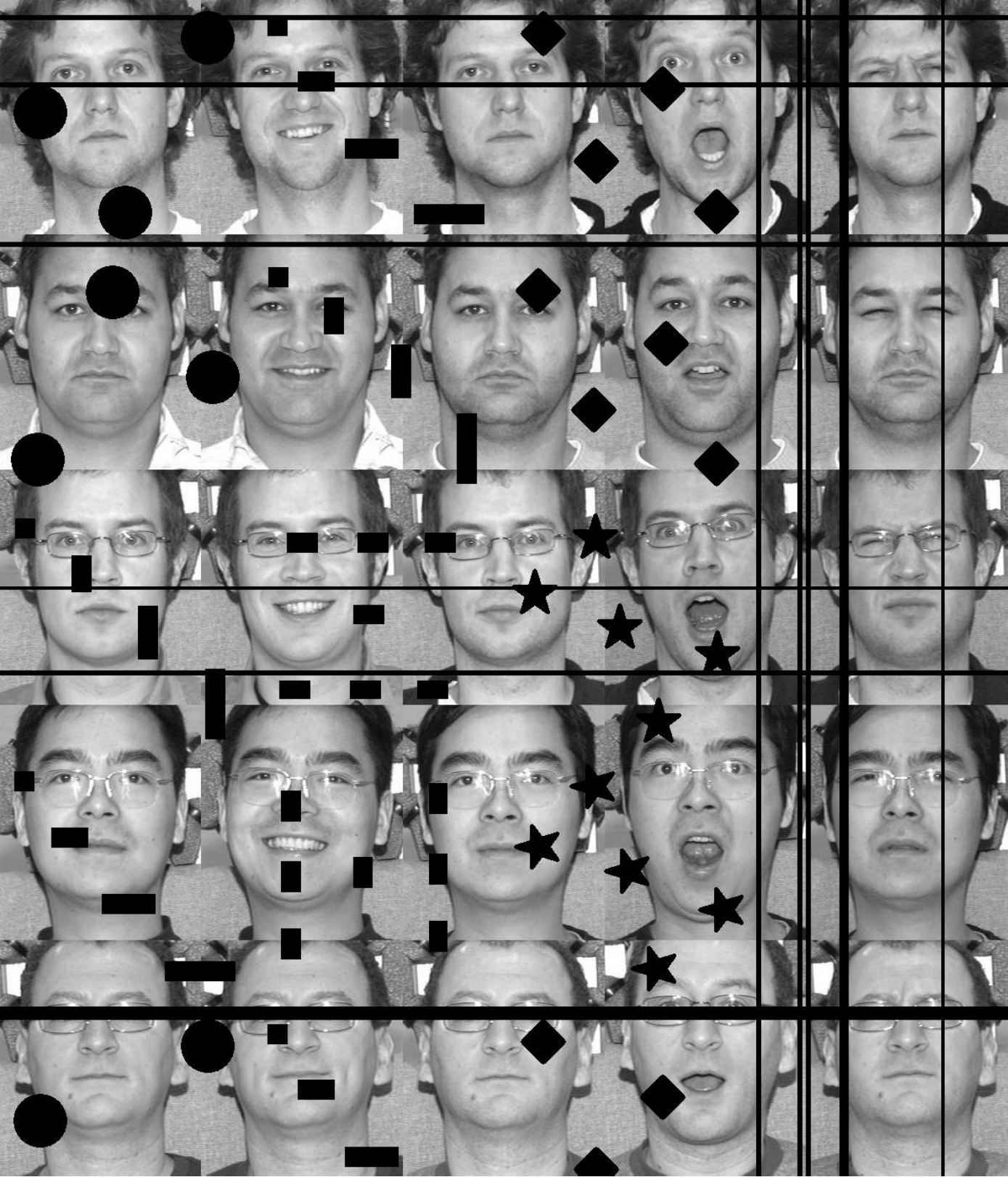}
	\includegraphics[width=0.24\textwidth]{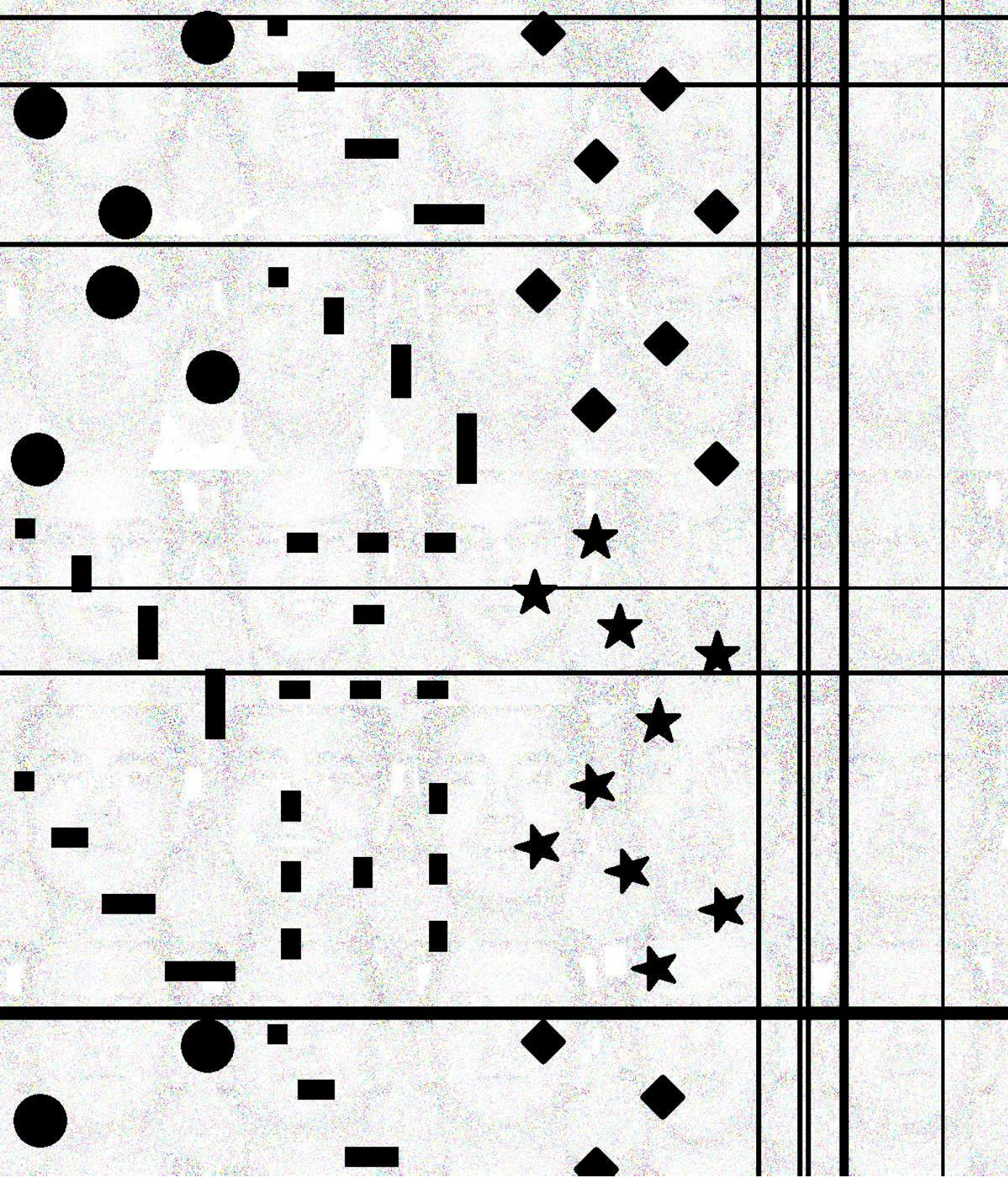}
	\caption{Six types of missing data in face images data sets. From left to right, top to bottom, the data sets are denoted by data missing Id $1,2,3,4,5,6$, respectively.}
	\label{six_type_of_missing_data}
\end{figure}

\begin{figure}[!htbp]
	\centering
	\includegraphics[width=.4\textwidth]{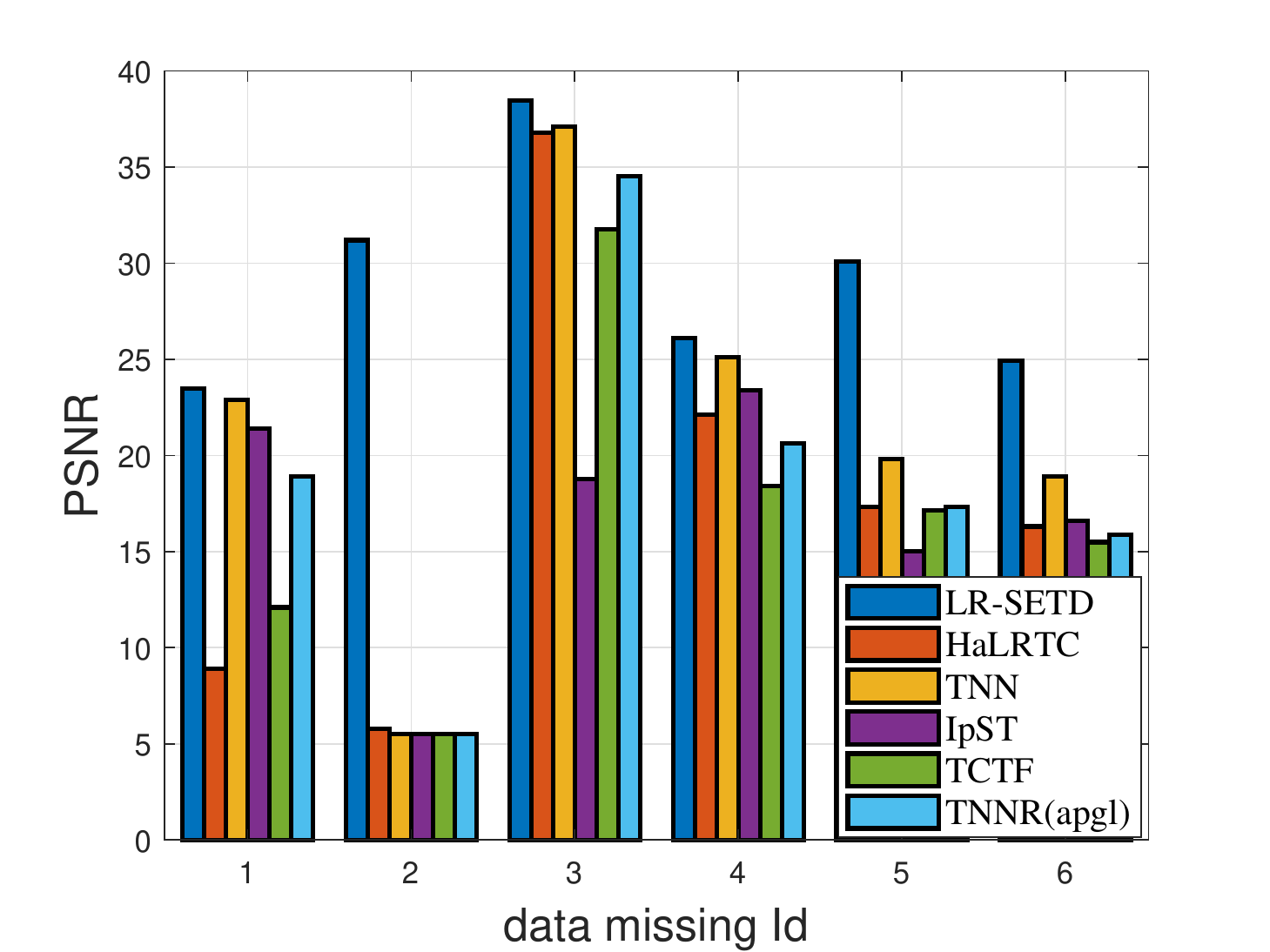}
	\includegraphics[width=.4\textwidth]{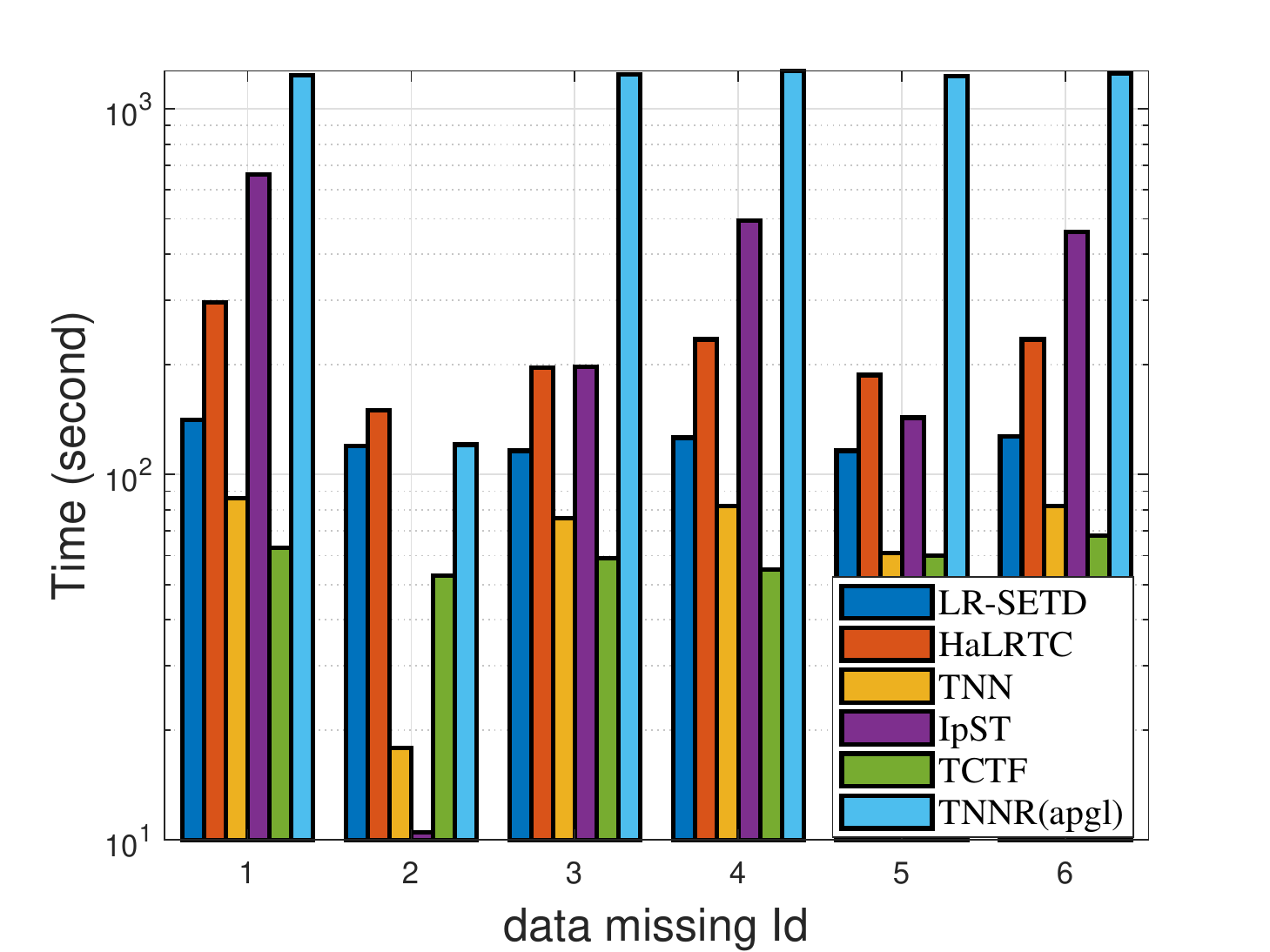}
	\caption{Numerical results obtained by the algorithms, where missing Id corresponds to the six types of missing data in Fig. \ref{six_type_of_missing_data}, respectively.}\label{Multipie_test_result}
\end{figure}

Finally, we consider the face recognition images data. The first face images dataset is the ORL face image database \cite{SH94} in AT/T Laboratories Cambridge, which contains images of $40$ persons, and each person has $10$ grayscale face images with different lighting, angles and facial expressions to form a group of photos. Moreover, each grayscale image has $112 \times92$ pixels, and one group of $10$ images forms a tensor with size $112\times 92\times 10$. The second image dataset is MultiPIE faces dataset\footnote{MultiPIE faces data sets: http://www.flintbox.com/public/project/4742/},  where each person has $5$ RGB face images, and each image has $280 \times 240 \times 3$ pixels. Consequently, one group of $5$ images forms a tensor with size $1400\times 1200\times 3$. In our experiments, we select two groups of photos in ORL dataset and five images of five persons in MultiPIE dataset, which can be seen in Fig. \ref{Xss2_visualization} and Fig. \ref{six_type_of_missing_data}, respectively, to validate the effectiveness of the proposed method. For the first dataset, we consider the case where $30\%$ information is observed, which is shown at the second row of Fig. \ref{Xss2_visualization}. It can be seen from the results in Fig. \ref{Xss2_visualization} that our approach LR-SETD is at least competitive with the other algorithms. For the corrupted data shown in Fig. \ref{six_type_of_missing_data}, we further show PSNR and computing time bar plots in Fig. \ref{Multipie_test_result}, which demonstrates that our approach LR-SETD can achieve higher PSNR values than the others, even it does not always perform the fastest one.

\begin{figure}
	\centering
	\includegraphics[width=0.24\textwidth]{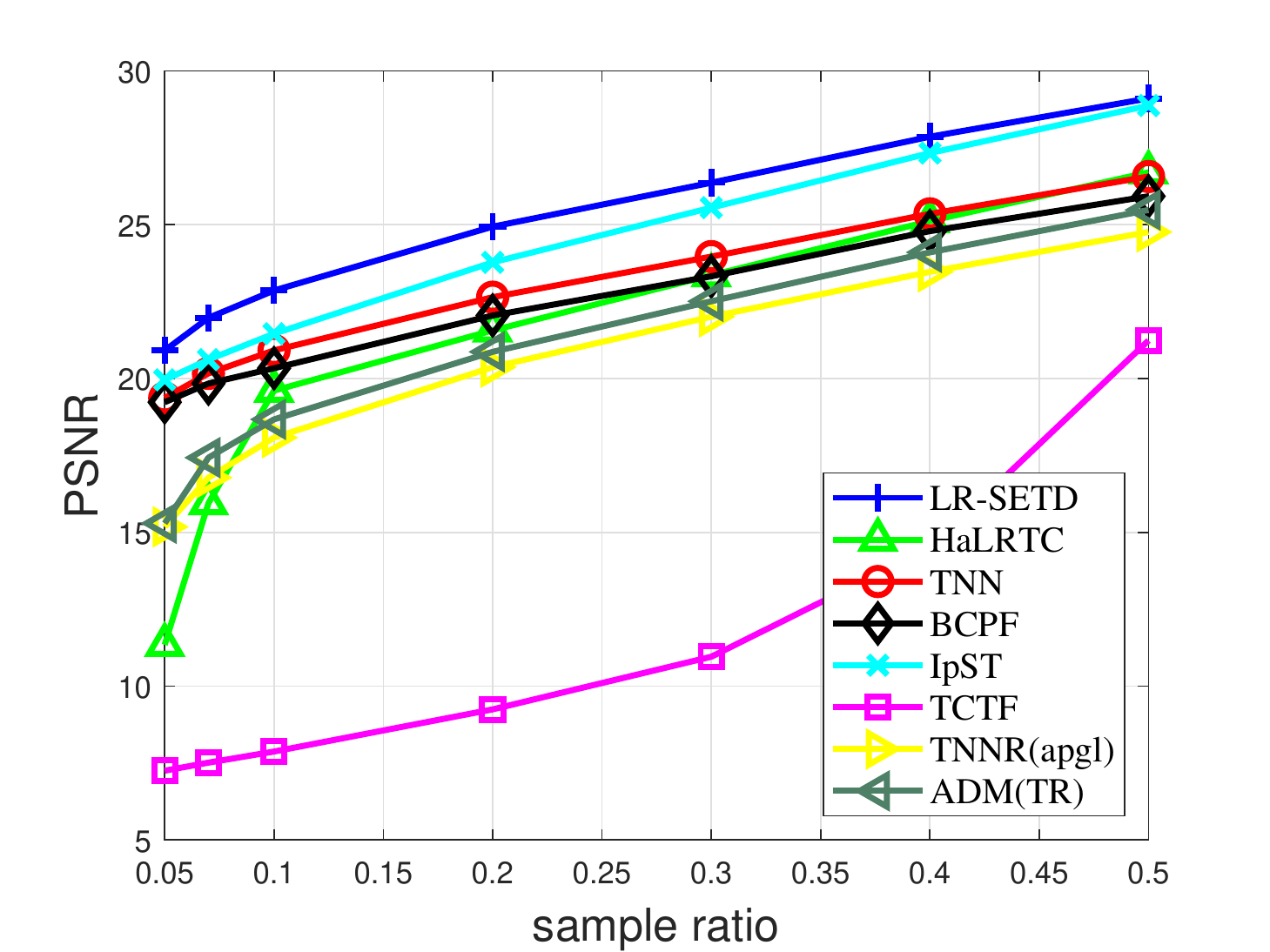}
	\includegraphics[width=0.24\textwidth]{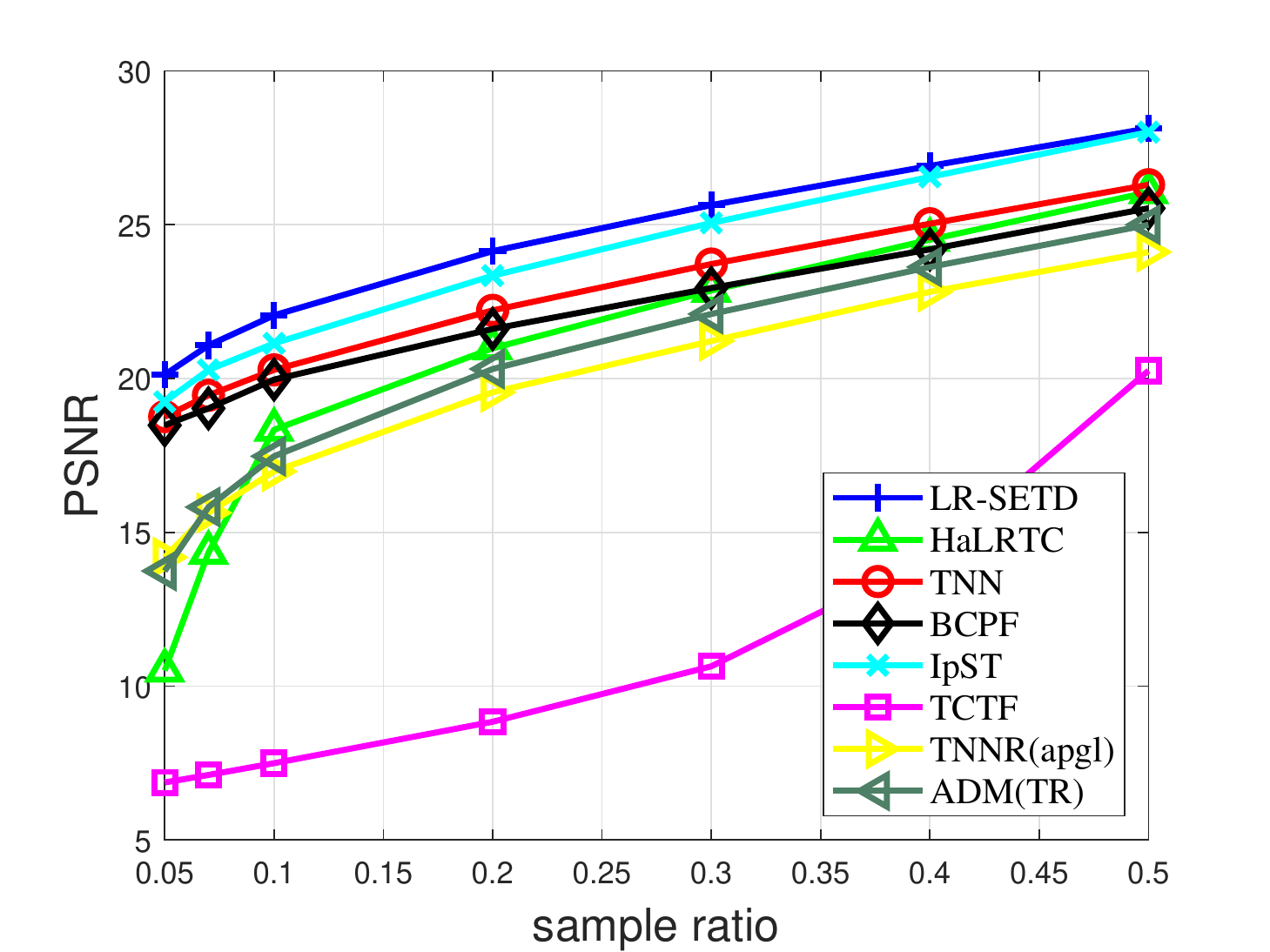}\\
	\includegraphics[width=0.24\textwidth]{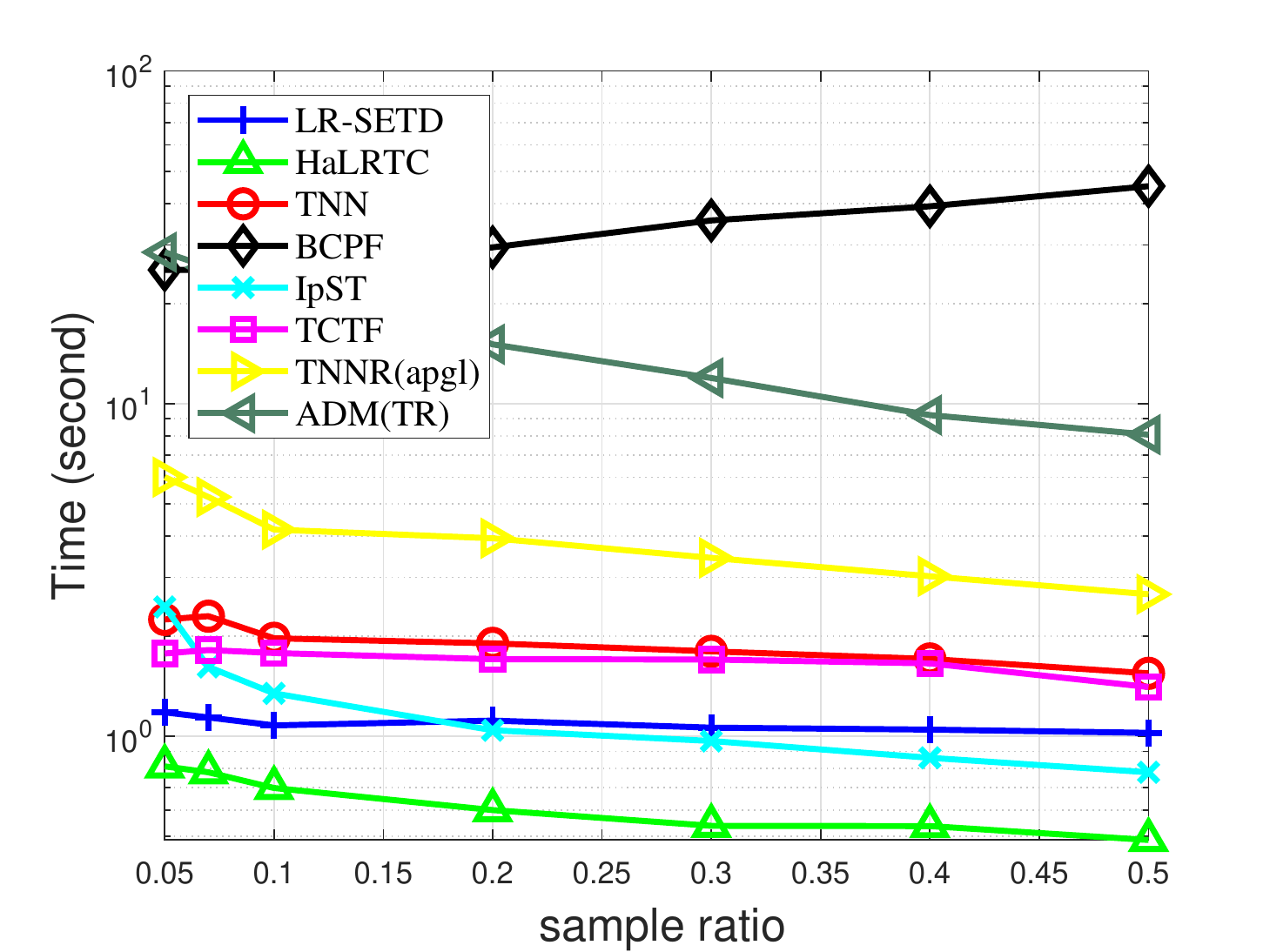}
	\includegraphics[width=0.24\textwidth]{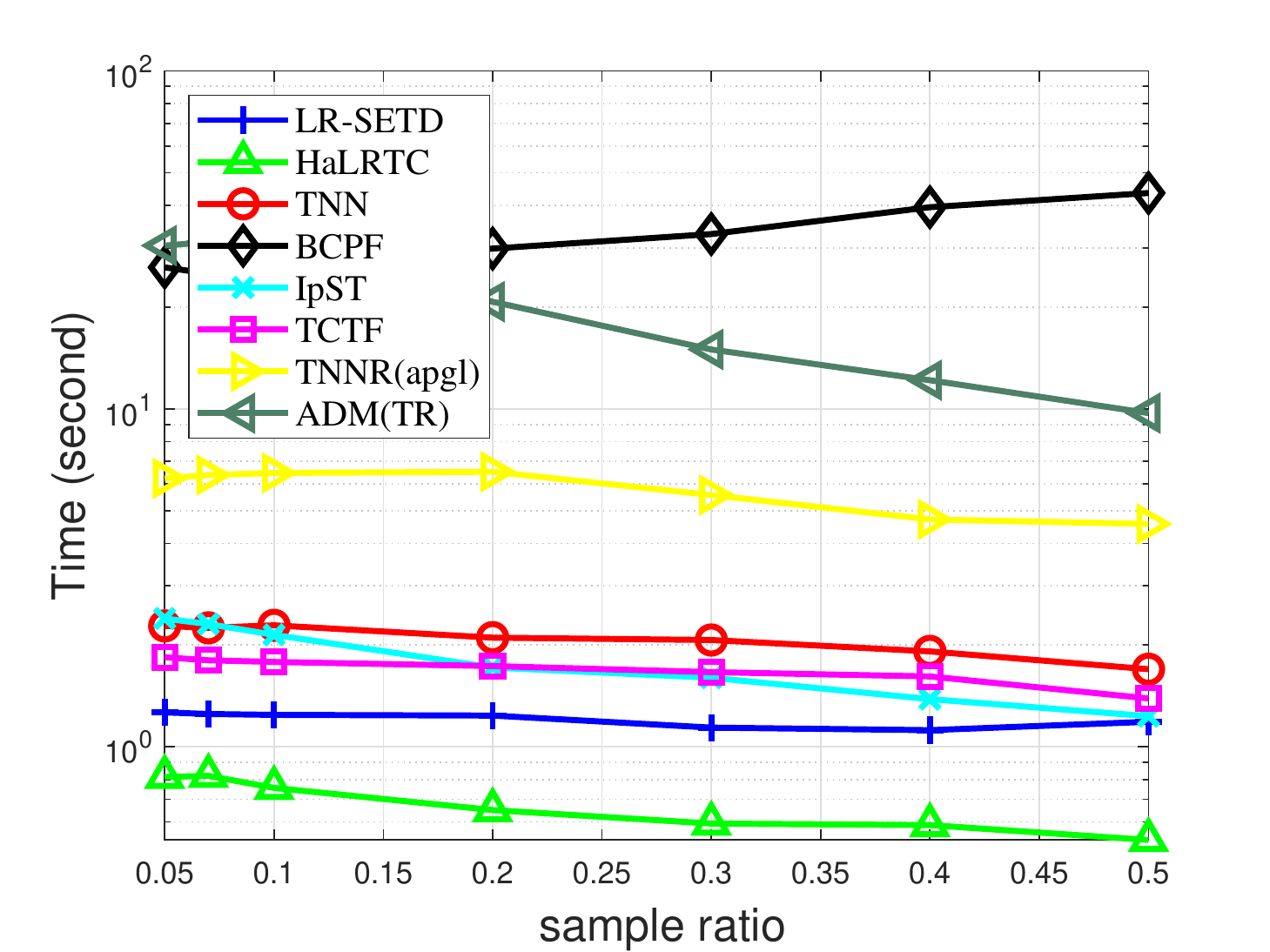}\\
	\caption{PSNR, Time(second) results under random missing of ORL Face.}
	\label{Face_rel_err}
\end{figure}

With the above numerical results, it can be seen from figures and tables that our approach LR-SETD works relatively stable, since it has the ability to handle different types of data including internet data, traffic data, color images, and face recognition data. All results also support the novelty of our model and numerical algorithm.

\section{Concluding remarks}\label{ConRemark}
In this paper, we proposed a low-rank and sparse enhanced Tucker decomposition model for tensor completion, where the core idea was that we exploited the potential low-rank properties of cofactor matrices and sparsity of the core tensor of Tucker decomposition. Comparing with the existing direct matricization methods, our model could maximally keep the inherent properties, e.g., periodicity and correlation, of the tensor data. Actually, a series of computational experiments on different types of real world data sets also demonstrate that our approach works well in terms of achieving ideal recovery accuracy. More promisingly, our approach performs much better than the other approaches when completing the tensor data with missing slices.

%
%\begin{table}[!t]
%% increase table row spacing, adjust to taste
%\renewcommand{\arraystretch}{1.3}
% if using array.sty, it might be a good idea to tweak the value of
% \extrarowheight as needed to properly center the text within the cells
%\caption{An Example of a Table}
%\label{table_example}
%\centering
%% Some packages, such as MDW tools, offer better commands for making tables
%% than the plain LaTeX2e tabular which is used here.
%\begin{tabular}{|c||c|}
%\hline
%One & Two\\
%\hline
%Three & Four\\
%\hline
%\end{tabular}
%\end{table}

% or


% \appendices

\section*{Acknowledgment}
The authors are grateful to Professor Kun Xie, Dr. Huibin Zhou and Dr. Yannan Chen for their kind help on the numerical experiments. Also, they would like to thank Dr. Yufan Li for sharing her code of \cite{SLH19}. Finally, many thanks go to the authors who shared their code and data on websites. C. Ling and H. He were supported in part by National Natural Science Foundation of China (Nos. 11971138 and 11771113) and Natural Science Foundation of Zhejiang Province (Nos. LY19A010019, LY20A010018, and LD19A010002).

\ifCLASSOPTIONcaptionsoff
  \newpage
\fi

%
%\bibliographystyle{IEEEtran}
%\bibliography{E:/Research/JabRef/TensorCD}


\end{document}